\documentclass{article}

\usepackage{tikz}
\usetikzlibrary{arrows.meta, positioning, calc, shapes.geometric, shapes.misc, shapes.callouts, decorations.pathreplacing, backgrounds, fit}

\definecolor{difforange}{HTML}{F5C095} % 浅橙色
\definecolor{diffblue}{HTML}{8DA0CB}   % 浅蓝色/蓝紫色
\definecolor{textorange}{HTML}{E89668} % 文字框橙色
\definecolor{textblue}{HTML}{7B8DBF}   % 文字框蓝色

\usepackage[algo2e,ruled,vlined,linesnumbered]{algorithm2e}
\usepackage{bm}
\usepackage{booktabs}   % 表格线
\usepackage{multirow}   % 表格合并
\usepackage{amssymb}    % 数学符号
\usepackage{amsmath}    % 数学支持
\usepackage{array}      % 表格控制
\usepackage{graphicx}   % 图片
\usepackage{subcaption} % 子图 (a), (b)...
\usepackage{pgfplots}   % 绘图核心包
\usepackage{caption}    % 用于 captionof
\pgfplotsset{compat=1.18} 
\usepackage{microtype}
\usepackage{graphicx}
\usepackage{stfloats}
\usepackage[colorlinks=true, allcolors=blue]{hyperref} % allcolors 可以改成你喜欢的颜色，比如 pink
\usepackage{url}
\usepackage{makecell}

\newenvironment{customlegend}[1][]{%
    \begingroup
    \csname pgfplots@init@cleared@structures\endcsname
    \pgfplotsset{#1}%
}{%
    \csname pgfplots@createlegend\endcsname
    \endgroup
}%
\def\addlegendimage{\csname pgfplots@addlegendimage\endcsname}
\pgfplotsset{compat=1.17}
\usepackage{tikz}
\usetikzlibrary{positioning, arrows.meta, backgrounds, fit, calc, patterns}
\definecolor{noise_col}{RGB}{189, 195, 199}  % 灰色代表噪声
\definecolor{data_col}{RGB}{65, 105, 225}    % 紫色代表语义/数据
\definecolor{arrow_col}{RGB}{44, 62, 80}     % 深色箭头
\definecolor{pred_col}{RGB}{230, 126, 34}

\usepackage{booktabs} % 用于三线表
\usepackage{float}    % 用于 [H] 强制定位
\usepackage{placeins} % 用于 \FloatBarrier
\usepackage[table]{xcolor}
\usepackage{booktabs} % for professional tables

\usepackage{hyperref}

\usepackage[accepted]{icml2025}

\usepackage{amsmath}
\usepackage{amssymb}
\usepackage{mathtools}
\usepackage{amsthm}

\usepackage[capitalize,noabbrev]{cleveref}

\theoremstyle{plain}

\theoremstyle{definition}

\theoremstyle{remark}

\usepackage[textsize=tiny]{todonotes}

\icmltitlerunning{FlowLM: Few-Step Language Modeling via Diffusion-to-Flow Adaptation}

\begin{document}

\twocolumn[
\icmltitle{FlowLM: Few-Step Language Modeling via Diffusion-to-Flow Adaptation}



\icmlsetsymbol{cor}{$\dagger$}

\begin{icmlauthorlist}
\icmlauthor{Runzhe Zhang}{SJTU}
\icmlauthor{Letian Chen}{SJTU}
\icmlauthor{Wenpeng Zhang}{comp}
\icmlauthor{Zhouhan Lin}{SJTU}
\icmlauthor{Peilin Zhao}{cor,SJTU}
\end{icmlauthorlist}

\icmlaffiliation{SJTU}{Shanghai Jiao Tong University}
\icmlaffiliation{comp}{ByteDance}

\icmlcorrespondingauthor{Peilin Zhao}{peilinzhao@sjtu.edu.cn}

\icmlkeywords{Machine Learning, ICML}

\vskip 0.3in
]



\printAffiliationsAndNotice{\textsuperscript{$\dagger$}Corresponding author.} 

% otherwise use the standard text.

\begin{abstract}
% This document provides a basic paper template and submission guidelines.
% Abstracts must be a single paragraph, ideally between 4--6 sentences long.
% Gross violations will trigger corrections at the camera-ready phase.
We present FlowLM, a flow matching language model transformed from pre-trained diffusion language models via efficient fine-tuning. By re-aligning the curved sampling trajectories of diffusion models into \textbf{straight-line} flows, FlowLM enables high quality few-step generation that rivals or even outperforms the quality of 2,000-step diffusion sampling with very few training epochs. Remarkably, finetuned FlowLM reaches performance saturation with only half as many training epochs as training from scratch, both approaches greatly outperforming the original diffusion model,
thereby validating our method. Furthermore, we validate a more effective training objective for flow matching: predicting clean data to consistently guide the sampling process towards the true data distribution. Empirical results demonstrate that our approach is highly effective for high-quality, few-step text generation.

\end{abstract}

\section{Introduction}
Currently, the field of Natural Language Processing is dominated by autoregressive models, such as GPT-4 \cite{achiam2023gpt} and LLaMA-3 \citep{dubey2024llama}. These models demonstrate remarkable capabilities in both understanding and generation tasks. However, AR models rely on a sequential, token-by-token generation mechanism. This characteristic inherently limits parallel computation and imposes a significant latency bottleneck during inference.

As a powerful alternative, Diffusion Language Models are rapidly emerging. 
% The core advantage of DLMs lies in their Non-Autoregressive generation paradigm. 
DLMs break the sequential constraint of AR models by generating the entire sequence or multiple tokens simultaneously through an iterative denoising process, theoretically offering higher inference throughput \cite{nie2025large, zhu2025llada}. Additionally, DLMs naturally leverage bidirectional context rather than the unidirectional attention mechanisms found in AR models. This capability facilitates finer-grained semantic understanding and controlled generation \cite{li2022diffusion, sahoo2024simple}.

Despite these advantages, DLMs face the challenge of the trade-off between sampling speed and quality. Standard diffusion processes typically require hundreds or even thousands of iterative denoising steps to generate high-quality text. Although parallel decoding strategies have been proposed to accelerate inference, when the number of sampling steps is reduced, there is a marked degradation in text coherence and quality \cite{gong2025scaling, wu2025fast}. Consequently, achieving significant reduction in sampling steps while maintaining generation quality remains a critical issue to be addressed in this field.

Diffusion Language Models (DLMs) can be broadly categorized into two main types: discrete and continuous. 
Discrete DLMs operate directly on categorical token spaces, aligning naturally with the symbolic nature of language \citep{gong2024scaling, xie2025dream}. 
Within this category, Masked Diffusion Language Models have emerged as a dominant paradigm. These models effectively bridge masked language modeling with the diffusion framework, demonstrating remarkable scalability in large-scale pre-training \citep{sahoo2024simple, lou2024discrete}.
While these models have shown promising performance, they suffer from prohibitively slow sampling.
% often requiring hundreds to thousands of steps due to the lack of effective acceleration techniques tailored for discrete diffusion.

In contrast, continuous DLMs \citep{gong2023diffuseqv2,tae2025tess} formulate the generation process within an embedding space. They are still inherently tied to stochastic differential equations that result in winding sampling trajectories. The core bottleneck is the mismatch between the curved paths learned during training and the desire for efficient few-step inference. Our approach, FlowLM, resolves this by re-aligning the curved trajectories of diffusion into straight-line flows.
% In contrast, this issue is less prominent in continuous DLMs, where continuous diffusion models and their corresponding acceleration methods predominate.
% Unlike discrete diffusion models, continuous Diffusion Language Models formulate the generation process within the embedding space compatible for acceleration, but they are inherently tied to stochastic differential equations that result in winding sampling trajectories. The core bottleneck is the mismatch between the complex, curved paths learned during training and the desire for large-step deterministic inference. Our approach, FlowLM, resolves this by re-aligning the curved trajectories of diffusion to the straight-line flows.

Our approach aims to fundamentally resolve the trade-off between generation quality and inference latency. 
% While auto-regressive models require $N$ function evaluations to generate a sequence of length $N$, and standard DLMs typically require hundreds of iterative refinement steps, whereas
FlowLM achieves high-quality generation with extremely few steps, offering a substantial speedup over diffusion language models.

Our contributions are summarized as follows:

\begin{itemize}
    \item We introduce a method that can finetune a continuous diffusion language model into a flow matching language model, achieving few-step sampling. This method greatly increases the sampling speed without sacrificing ability with very few training epochs, providing a reference for transforming diffusion language model into flow matching language model. 
    
    \item We also give further explanation for its validity, attributing it to average velocity, and demonstrate through experiments that predicting and sampling with instantaneous velocity does not work effectively in this context. 
    
    \item We further confirm the validity of training flow matching model by predicting the clean data distribution $z_0$ in the text generation field, as FlowLM has best performance. However, compared to \citet{li2025back}, we find training with x-pred and x-loss yields superior performance in this context.
\end{itemize}

\section{Preliminaries}

In this section, we briefly review the fundamentals of Diffusion Models and Flow Matching, and introduce the specific formulation used in DiffuSeq.

\subsection{Diffusion Models}
Diffusion models \citep{ho2020denoising} are generative models that learn to reverse a gradual noising process.
Given a data sample $\mathbf{z}_0 \sim q(\mathbf{z})$, the \textbf{forward process} adds Gaussian noise over $T$ steps according to a variance schedule $\beta_t$:
\begin{equation}
    q(\mathbf{z}_t | \mathbf{z}_{t-1}) = \mathcal{N}(\mathbf{z}_t; \sqrt{1-\beta_t}\mathbf{z}_{t-1}, \beta_t \mathbf{I}).
\end{equation}
Using the notation $\alpha_t = 1 - \beta_t$ and $\bar{\alpha}_t = \prod_{i=1}^t \alpha_i$, we can sample $\mathbf{z}_t$ directly from $\mathbf{z}_0$:
\begin{equation}
    \mathbf{z}_t = \sqrt{\bar{\alpha}_t}\mathbf{z}_0 + \sqrt{1 - \bar{\alpha}_t}\bm{\epsilon}, \quad \text{where } \bm{\epsilon} \sim \mathcal{N}(\mathbf{0}, \mathbf{I}).
\end{equation}
The \textbf{reverse process} aims to recover $\mathbf{z}_0$ from noise $\mathbf{z}_T \sim \mathcal{N}(\mathbf{0}, \mathbf{I})$. This is achieved by learning a denoising network $f_\theta(\mathbf{z}_t, t)$ to approximate the conditional distribution $p_\theta(\mathbf{z}_{t-1}|\mathbf{z}_t)$. The training objective is typically a simplified Mean Squared Error loss:
\begin{equation}
    \mathcal{L}_{\text{diff}} = \mathbb{E}\left[ \| \bm{\epsilon} - \bm{\epsilon}_\theta(\mathbf{z}_t, t) \|^2 \right] \quad 
    \text{or} \quad \mathbb{E}\left[ \| \mathbf{z}_0 - f_\theta(\mathbf{z}_t, t) \|^2 \right].
\end{equation}

\subsection{Flow Matching}
Flow Matching \citep{lipman2023flow} provides an alternative simulation-free framework to learn a vector field that defines a deterministic transport between distributions. FM constructs a straight-line trajectory between noise and data, enabling few-step generation.
A flow is defined by a time-dependent vector field $v_t(\mathbf{z})$, which generates a flow map via an Ordinary Differential Equation $\frac{d\mathbf{z}}{dt} = v_t(\mathbf{z}).$

The goal of FM is to regress this vector field $v_t(\mathbf{z})$ such that it pushes the prior distribution $p_0$ (noise) to the data distribution $p_1$ (data). A common and efficient instantiation is a straight-line path $
    \mathbf{z}_t = t \mathbf{z}_1 + (1-t) \mathbf{z}_0$,   
where $\mathbf{z}_1 \sim \mathcal{N}(\mathbf{0}, \mathbf{I})$ and $\mathbf{z}_0 \sim q(\mathbf{z})$.
% The target vector field for this linear path is simply $v_t(\mathbf{z}_t) = \mathbf{z}_1 - \mathbf{z}_0$.
The loss function is:
\begin{equation}
    \mathcal{L}_{\text{FM}} = \mathbb{E}_{t, \mathbf{z}_0, \mathbf{z}_1} \left[ \| v_\theta(\mathbf{z}_t, t) - (\mathbf{z}_1 - \mathbf{z}_0) \|^2 \right].
\end{equation}
This linear trajectory is crucial for fast sampling, as the ODE can be solved with fewer steps compared to the curved trajectories typical of diffusion models.

\subsection{DiffuSeq Formulation}
DiffuSeq \citep{gong2023diffuseq} extends continuous diffusion to sequence-to-sequence tasks by mapping discrete tokens into a continuous embedding space.
Let $\mathbf{w}^x$ be the source text sequence and $\mathbf{w}^y$ be the target text sequence. An embedding function $\text{EMB}(\cdot)$ maps them to continuous vectors. DiffuSeq constructs a joint latent variable $\mathbf{z}_0$ by concatenating the embeddings $
    \mathbf{z}_0 = [\text{EMB}(\mathbf{w}^x) ; \text{EMB}(\mathbf{w}^y)].$
    
Since the source text $\mathbf{w}^x$ is given as a condition, noise is only added to the target part $\mathbf{w}^y$. The forward process at step $t$ can be denoted as:
\begin{equation}
    \mathbf{z}_t = [\text{EMB}(\mathbf{w}^x) ; \mathbf{y}_t], \quad \text{where } \mathbf{y}_t \sim q(\mathbf{y}_t | \text{EMB}(\mathbf{w}^y)).
\end{equation}
The model is trained to recover the clean target embedding $\mathbf{y}_0$ from the partially noised input $\mathbf{z}_t$. The simplified training objective typically includes a reconstruction term and a regularization term:
\begin{equation}
    \mathcal{L} = \sum_{t=2}^T \|\mathbf{z}_0 - f_{\theta}(\mathbf{z}_t,t)\|^2 + \mathrm{CE}(decoder(EMB(\mathbf{w}^y)),\mathbf{w}^y),
\end{equation}
where loss CE is the cross entropy loss between the ground truth text and the result decoded from the embedded $\mathbf{w}^y$ to make sure the decoder head can accurately transform the data in continuous embedding space into text.

% --- 宏定义 ---
\newcommand{\drawtensor}[5]{
    \node[tensor, label={[labelnode]below:#3}] (#1) at #2 {};
    \begin{scope}
        \clip[rounded corners=2pt] 
            ($(#1.south west)+(0.05,0.05)$) rectangle ($(#1.north east)+(-0.05,-0.05)$);
        \fill[#5!80] 
            ($(#1.south west)$) rectangle ($(#1.east)+(0, {2.8*(#4-0.5)})$);
        \fill[noise_col!50] 
            ($(#1.west)+(0, {2.8*(#4-0.5)})$) rectangle ($(#1.north east)$);
        \pgfmathsetseed{100} 
        \foreach \i in {1,...,40} {
            \pgfmathsetmacro{\x}{rand*0.5}
            \pgfmathsetmacro{\y}{rand*1.3}
            \fill[arrow_col, opacity=0.3] ($(#1.center)+(\x,\y)$) circle (0.02);
        }
    \end{scope}
    \draw[thick, arrow_col, rounded corners=2pt] (#1.south west) rectangle (#1.north east);
}

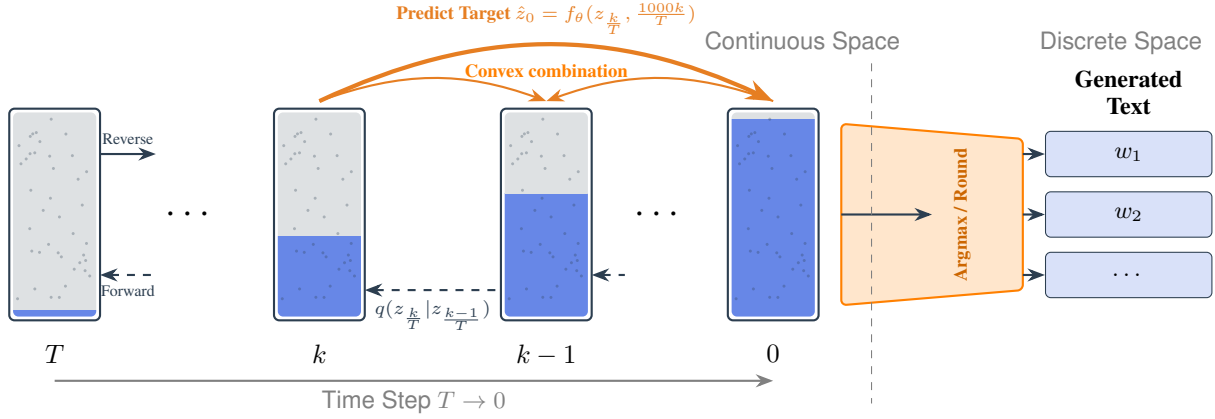
\begin{figure*}
\centering
\begin{tikzpicture}[
    font=\sffamily\small,
    >=Stealth,
    tensor/.style={
        minimum width=1.2cm, minimum height=2.8cm,
        inner sep=0pt, outer sep=0pt
    },
    token/.style={
        draw=arrow_col, fill=data_col!20, rounded corners=2pt,
        minimum width=2.2cm, minimum height=0.6cm,
        font=\footnotesize, align=center
    },
    labelnode/.style={
        text=black, font=\bfseries, below=0.2cm
    }
]

    % --- 1. 左侧扩散链 (Latent Space) ---
    \drawtensor{zt}{(-6,0)}{$T$}{0.05}{data_col}
    
    \node (dots1) at (-4.2, 0) {\Large $\cdots$};
    \draw[->, thick, arrow_col] ($(zt.east)+(0,0.8)$) -- node[above, font=\tiny] {Reverse} ($(dots1.west)+(0,0.8)$);
    \draw[<-, thick, arrow_col, dashed] ($(zt.east)+(0,-0.8)$) -- node[below, font=\tiny] {Forward} ($(dots1.west)+(0,-0.8)$);

    % z_t 和 z_{t-1}
    \drawtensor{ztt}{(-2.5,0)}{$k$}{0.4}{data_col}
    \drawtensor{zt1}{(0.5,0)}{$k-1$}{0.6}{data_col} %稍微拉开一点距离方便画箭头

    % --- 关键修改：你的特殊采样路径 ---
    
    % 1. 预测箭头 (从 z_t 飞向 z_0)
    % 使用 bend left 形成一个跨越的大弧线
    \draw[->, ultra thick, pred_col] 
        ($(ztt.north)+(0,0.1)$) 
        to[bend left=25] 
        node[midway, above, font=\bfseries\scriptsize, text=pred_col, align=center] {Predict Target $\hat{z}_0 = f_\theta(z_{\frac{k}{T}}, \frac{1000k}{T})$} 
        ($(3.5, 1.5)$); % 指向 z0 的上方
        
    % 2. 回归/插值箭头 (从 z_0 指回 z_{t-1})
    % 这里的逻辑是：基于预测的 z0，我们算出方向，然后迈出一步到达 z_{t-1}
    \draw[->, thick, pred_col] 
        ($(3.5, 1.5)$) % 从 z0 附近的落点开始
        to[bend right=20] 
        node[midway, below, font=\bfseries\scriptsize, text=pred_col] {}
        ($(zt1.north)+(0,0.1)$);
    % 在 ztt 和 zt1 之间的上方连一条弯曲的箭头
    \draw[->, thick, pred_col] 
        ($(ztt.north)+(0,0.1)$) 
        to[bend left=20] 
        node[midway, above, font=\bfseries\scriptsize, text=pred_col]{}  
        ($(zt1.north)+(0,0.1)$);
    \node[text=orange, font=\bfseries\scriptsize] at ($(zt1.north)+(0,0.5)$) {Convex combination};
    
    % 保留底部的 Forward 箭头 (q分布)
    \draw[<-, thick, arrow_col, dashed] 
        ($(ztt.east)+(0,-1.0)$) -- 
        node[below, font=\scriptsize] {$q(z_{\frac{k}{T}}|z_{\frac{k-1}{T}})$} 
        ($(zt1.west)+(0,-1.0)$);

    % --- 继续后续部分 ---
    \node (dots2) at (2.0, 0) {\Large $\cdots$};
    % 这里不需要画反向箭头了，因为逻辑已经在上面展示了，只画前向噪声过程
    \draw[<-, thick, arrow_col, dashed] ($(zt1.east)+(0,-0.8)$) -- ($(dots2.west)+(0,-0.8)$);

    % z_0 (纯数据)
    \drawtensor{z0}{(3.5,0)}{$0$}{0.95}{data_col}

    % --- 2. 右侧离散映射部分 ---
    
    \node[token] (w1) at (8.2, 0.8) {$w_1$};
    \node[token] (w2) at (8.2, 0.0) {$w_2$};
    \node[token] (w3) at (8.2, -0.8) {$\dots$};
    \node[above=0.1cm of w1, font=\footnotesize\bfseries, align=center] {Generated\\Text};

    % 梯形背景
    \path[fill=orange!20, draw=orange, thick, rounded corners=2pt] 
        ($(z0.east)+(0.3, 1.2)$) -- (6.8, 1.0) -- (6.8, -1.0) -- ($(z0.east)+(0.3, -1.2)$) -- cycle;
    
    \node[text=orange!80!black, rotate=90, font=\scriptsize\bfseries] at (6.0, 0) {Argmax / Round};

    \draw[->, thick, arrow_col] ($(z0.east)+(0.3,0)$) -- (5.6, 0); 
    \draw[->, thick, arrow_col] (6.8, 0.8) -- (w1.west);
    \draw[->, thick, arrow_col] (6.8, 0.0) -- (w2.west);
    \draw[->, thick, arrow_col] (6.8, -0.8) -- (w3.west);

    % --- 3. 底部注释 ---
    \draw[->, thick, gray] (-6, -2.2) -- node[below] {Time Step $T \to 0$} (3.5, -2.2);
    
    \draw[dashed, gray] (4.8, 2.0) -- (4.8, -2.5);
    \node[gray, anchor=south east] at (5.3, 2.0) {Continuous Space};
    \node[gray, anchor=south west] at (6.9, 2.0) {Discrete Space};

\end{tikzpicture}
\caption{Schematic of FlowLM}
\end{figure*}

\section{Method}

% \begin{figure}[H]
%     \centering
%     \includegraphics[width=1\linewidth]{trajectory_final.jpg}
%     \caption{Comparison in sampling Path.}
% \end{figure}
The core of our methodology lies in fine-tuning the diffusion model to transform its originally curved generative trajectory into a straight flow path.

\subsection{Learning the Straight-line Trajectory}
\begin{algorithm2e}[t]
\caption{FlowLM training}
\label{alg:diffuseq_v2_conditional}
\SetKwInOut{Input}{Input}
\SetKwInOut{Init}{Initialize}

\Input{Dataset $\mathcal{D}$ $(\mathbf{w}^x, \mathbf{w}^y)$ (Source, Target), total steps $T$}
\Init{initialized from diffusion LM}

\BlankLine
\While{not converged}{
    \tcp{1. Data Preparation}
    Sample batch $(\mathbf{w}^x, \mathbf{w}^y) \sim \mathcal{D}$\;
    
    $\mathbf{z}^x \leftarrow \text{EMB}(\mathbf{w}^x)$ \tcp*[r]{Condition}
    $\mathbf{z}^y_0 \leftarrow \text{EMB}(\mathbf{w}^y)$ \tcp*[r]{Target}
    Sample time step $t_{step} \sim \text{Uniform}(\{1, \dots, T\})$\;
    
    \tcp{2. Joint Noise on Target}
    Sample Gaussian noise $\boldsymbol{\epsilon} \sim \mathcal{N}(\mathbf{0}, \mathbf{I})$ 
    $t=\frac{t_{step}}{T}$\;
    
    $\mathbf{z}^y_{\text{t}} \leftarrow (1-t)*\mathbf{z}^y_0 + t*\boldsymbol{\epsilon}$\;
    
    $\mathbf{z}_{\text{in}} \leftarrow \text{Concat}(\mathbf{z}^x, \mathbf{z}^y_t)$\;
    
    $t_{in}=\frac{1000t}{T}$\; 
    
    $\mathbf{z}_{0,pred} \leftarrow \text{ExtractTarget}(f_\theta(\mathbf{z}_{\text{in}}, t_{in}))$\;
    \tcp{3. Loss Calculation}
$\mathcal{L}_{\text{total}} \leftarrow  \|\mathbf{z}_0 - \mathbf{z}_{0,pred}\|^2 + CE(decoder_{head}(\mathbf{z}_0^y),\mathbf{w}^y)+ Reg_{rate}*\frac{\|\mathbf{x}_{0-pred-diffu}-\mathbf{x}_{0-pred-flow} \|^2}{t^2}$;

    \tcp{4. Update}
}

\end{algorithm2e}
Our FlowLM fine-tuning initializes the Flow Matching model from the original diffusion language model, and forces the model to straighten its generative path by employing the linear interpolation
% In each training epoch, we randomly sample a time step $t$ and construct the noisy state $\mathbf{z}_t$ using the linear interpolation formula 
$\mathbf{z}_t = t \mathbf{z}_1 + (1-t) \mathbf{z}_0$,
where $\mathbf{z}_1$ represents the noise and $\mathbf{z}_0$ represents the clean data. We then feed $\mathbf{z}_t$ and $t$ into the model to predict $\mathbf{z}_0$.

The decision to have the model predict $\mathbf{z}_0$ is intended to preserve the original diffusion model's capability. 
According to \citet{li2025back}, training with x-pred and v-loss is the best in image generation field. However, we conduct a detailed comparison and find that using x-pred and x-loss is better here (x is equivalent to $\mathbf{z}_0$ in our paper). The v-loss applies a $\frac{1}{t^2}$ weight to the x-loss, focusing capacity on the near-clean stages of generation. Unlike high-resolution images that require high precision, the discrete nature of token decoding makes such late-stage refinement less beneficial as the model is inherently tolerant of the small latent deviations. A detailed comparison is provided in Table \ref{tab:compare with HKM}, Table \ref{tab:compare with HKM1} .

\subsection{High-speed Inference via Velocity Approximation}

\begin{algorithm2e}[t]
\caption{FlowLM Sampling}
\label{alg:inference}
\SetAlgoLined

\For{$k \leftarrow T$ \textbf{to} $1$}{
    $t \leftarrow k/T$, 
    $dt \leftarrow 1.0/T$ \\
    $t_{\text{input}} \leftarrow t \times 1000$ 
    
    $\mathbf{z}_{0,pred} \leftarrow \text{ExtractTarget}(f_\theta(\mathbf{z}_{\text{input}}, t_{\text{input}}))$ \\
     $\mathbf{v} = (\mathbf{z}_t - \mathbf{z}_{0,pred})/t$ \\
    $\mathbf{z}^y \leftarrow \mathbf{z}^y - \mathbf{v} \times dt$ \quad //From t=1 to t=0
}

$result \leftarrow \text{DecoderHead}(\mathbf{z}^y)$ \\
\end{algorithm2e}
% In curved diffusion paths, the instantaneous velocity only provides local guidance, necessitating hundreds of steps. In contrast, 
FlowLM samples with average velocity between $\mathbf{z}_t$ and $\mathbf{z}_0$, calculated by $\mathbf{v}_{average}= \frac{\mathbf{z}_t - \mathbf{z}_{0,pred}}{t}$, providing global guidance toward the data distribution. The updated $\mathbf{z}_{next}$ from $z_t$ is:
\begin{equation}
    \mathbf{z}_{next}=\mathbf{z}_{t}-\mathbf{v}_{average}*dt=(1-\frac{dt}{t})\mathbf{z}_{t}+\frac{dt}{t}\mathbf{z}_{0,pred}
\end{equation}
This implies that each updated $z_{\text{next}}$ is a convex combination of $z_t$ and predicted $z_0$ weighted by the time ratio.

If the path is perfectly straight and always satisfies $\mathbf{z}_t = t \mathbf{z}_1 + (1-t) \mathbf{z}_0$, the average velocity is equal to instantaneous velocity:
\begin{equation}
    \mathbf{v}_{instant} = \mathbf{z}_1-\mathbf{z}_0=\frac{\mathbf{z}_t-(1-t)\mathbf{z_0}}{t}-\mathbf{z}_0= \frac{\mathbf{z}_t - \mathbf{z}_{0}}{t}.
\end{equation}

However, note that our sampling formula $\mathbf{v}_{average}=\frac{\mathbf{z}_t - \mathbf{z}_{0,pred}}{t}$ cannot be treated as instantaneous velocity here because the equivalence holds only when the trajectory is perfectly straightened.  Using $\mathbf{v}_{average}$ is more stable in training and can effectively avoid the accumulation of errors in sampling. As the path can not be perfectly straight, the average velocity can provide more accurate guidance to clean data distribution. The effectiveness is demonstrated by our experiment results \ref{tab:main_results}. More detailed analysis of our method's advantages and comparison with other methods are shown in Appendix \ref{sec: More about FlowLM}.

\subsection{Training details }

\begin{table*}[t]
    \centering
    \caption{Comparison of loss distribution in different time step intervals between Diffusion and Flow Matching models.}
    \vspace{0.5em}
    % 增加行间距，使表格不拥挤
    \renewcommand{\arraystretch}{1.2}
    \begin{tabular}{lcccc}
        \toprule
        \textbf{Method} & $\mathcal{L}_{q_0}$ & $\mathcal{L}_{q_1}$ & $\mathcal{L}_{q_2}$ & $\mathcal{L}_{q_3}$ \\
        \midrule
        Diffusion & $1.48 \times 10^{-2}$ & $1.46 \times 10^{-2}$ & $1.49 \times 10^{-2}$ & $1.47 \times 10^{-2}$ \\
        Flow Matching & $9.04 \times 10^{-4}$ & $1.66 \times 10^{-3}$ & $2.70 \times 10^{-3}$ & $6.39 \times 10^{-3}$ \\
        \bottomrule
    \end{tabular}
    \label{tab:loss_comparison}
\end{table*}

\begin{table*}[t]
\centering
\caption{Comparison between FlowLM and Diffusion LM \textbf{(MBR=1)}. The best results of few-step generation model are \textbf{bold}.}
\vspace{0.5em}
\renewcommand{\arraystretch}{1.1} %稍微增加行高，让表格不那么拥挤
% 定义列格式：
% l: 左对齐
% c: 居中
% |: 竖线
\begin{tabular}{lll | ccc | cc }
\toprule
Tasks & Type &Methods & BLEU$\uparrow$ & R-L$\uparrow$ & BERTScore$\uparrow$ & dist-1$\uparrow$ & Training epoch\\
\midrule

\multirow{6}{*}{\shortstack[l]{Question \\Generation}} 
 & Multi-step & DiffuSeq(step=2000) & 0.1527 & 0.3474 & 0.5864 & 0.9113 & 34000\\
 \cmidrule(lr){2-8} % 线条从第2列（分类列）划到第8列
 & \multirow{4}{*}{Few-step} & DiffuSeq(DPM-solver,step=10) & 0.1425 & 0.3509 & 0.5730 & 0.8606 &34000\\
 & & FlowLM(FT, step=5) & \textbf{0.1596} & 0.3484 & \textbf{0.5898} & \textbf{0.9206}&6000  \\
 & & FlowLM(FT, step=3) & 0.1595 & 0.3489 & 0.5878 & 0.9169 &6000\\
 & & FlowLM(FT, step=1) & 0.1524 & \textbf{0.3550} & 0.5713 & 0.8411 &6000\\
\hline 
% ==================== Task 2: Paraphrase ====================
\multirow{5}{*}{\shortstack[l]{Paraphrase}} 
  & Multi-step & DiffuSeq(step=2000) & 0.1880 & 0.5306 & 0.7918 & 0.9736 &34000\\
  \cmidrule(lr){2-8}
 & \multirow{4}{*}{Few-step} & DiffuSeq(DPM-solver,step=10) & 0.1991 & \textbf{0.5580} & \textbf{0.7943} & 0.9570 &34000\\
 && FlowLM(FT, step=5) & 0.1942 & 0.5352 & 0.7830 & 0.9764&10000 \\
 && FlowLM(FT, step=3) & \textbf{0.2001} & 0.5390 & 0.7809 & \textbf{0.9766}&10000 \\
 && FlowLM(FT, step=1) & 0.1896 & 0.5404 & 0.7570 & 0.9443 &10000 \\
\hline 

% ==================== Task 3: Text Simplification ====================
\multirow{5}{*}{\shortstack[l]{Text\\Simpli-\\fication}} 
 & Multi-step & DiffuSeq(step=2000) & 0.2956 & 0.5315 & 0.7783 & 0.9258&34000 \\
 \cmidrule(lr){2-8}
 & \multirow{4}{*}{Few-step}& DiffuSeq(DPM-solver,step=10) & 0.2294 & 0.4676 & 0.6886 & 0.8779 &34000\\
 && FlowLM(FT, step=5) & \textbf{0.2601} & \textbf{0.4868} & \textbf{0.7316} & \textbf{0.9054} &8000\\
  && FlowLM(FT, step=3) & 0.2539 & 0.4827 & 0.7179 & 0.8834 &8000 \\
 && FlowLM(FT, step=1) & 0.2316 & 0.4513 & 0.6420 & 0.7622 &8000 \\
 % \cmidrule(lr){2-8}
 % & \multirow{3}{*}{Few-step}& \textsc{FlowLM2}(step=5) & 0.2294 & 0.4676 & 0.6886 & 0.8779 & 11.15 \\
 %  && \textsc{FlowLM2}(step=10) & 0.2539 & 0.4827 & 0.7179 & 0.8834 & 16.29 \\
 % && \textsc{FlowLM2}(step=20) & 0.2316 & 0.4513 & 0.6420 & 0.7622 & 13.40 \\
\hline 

\bottomrule
\end{tabular}

\label{tab:main_results}

\centering
\caption{Comparison between different accelerating methods \textbf{(MBR=1)}. The best results of few-step generation models are \textbf{bold}. Scr: Trained from scratch. FT: Finetuned.}
\renewcommand{\arraystretch}{1.1} %稍微增加行高，让表格不那么拥挤
% 定义列格式：
% l: 左对齐
% c: 居中
% |: 竖线
\vspace{0.5em}

\begin{tabular}{lll | ccc | c | cc }
\toprule
Tasks & Type &Methods & BLEU$\uparrow$ & R-L$\uparrow$ & BERT$\uparrow$ & dist$\uparrow$ & \makecell{Inference \\Time(s)} & \makecell{Training \\epoch}\\
\midrule

\multirow{9}{*}{\shortstack[l]{Question \\Generation}} 
 & Multi-step & DiffuSeq(step=2000) & 0.1527 & 0.3474 & 0.5864 & 0.9113 & 0.7264 & 34000\\
 \cmidrule(lr){2-9} % 线条从第2列（分类列）划到第8列

 & \multirow{8}{*}{Few-step} & FlowLM(Scr, step=1) & 0.1624 & \textbf{0.3650} & 0.5932& 0.8790 & 0.00014 & 34000\\
 &  & FlowLM(FT, step=1) & \textbf{0.1634} & 0.3549 & \textbf{0.5943} & \textbf{0.9006} & 0.00014 & 15000\\
 &  & FlowLM(FT, step=1) & 0.1524 & 0.3550 & 0.5713 & 0.8411 & 0.00014 & 6000\\
  & & FMseq(Scr,step=1) & 0.1617 & 0.3529 & 0.5938 & 0.8937 & 0.00054 & 34000 \\
 & & DLM-One(step=1) & 0.1512 & 0.3257 & 0.5683 & 0.9053 & - & -\\
 & & ReFlow(x-pred,step=1) & 0.1470 & 0.3545 & 0.5634 & 0.8201 & 0.00017 & 6000\\
 & &ReFlow(v-pred,step=5) & 0.0002 & 0.0009 & 0.2306 & 0.2336 & 0.00059 & 34000 \\
  & & Perflow(v-pred,step=5) & 0.0002 & 0.0011 & 0.2804 & 0.8476 & 0.00059 &34000 \\
\bottomrule
\end{tabular}

\label{tab:main_results}

\end{table*}

% \FloatBarrier 

To ensure a fair comparison with Diffusion-LM, we align our training configuration as closely as possible with that of the original DiffuSeq \citep{gong2023diffuseq}. FlowLM is initialized by provided checkpoints of DiffuSeq. We also provide comparison against the DiffuSeq-V2 model \citep{gong2023diffuseqv2} accelerated with the DPM-Solver to validate our method's efficiency.

We also strictly maintain consistency in the model inputs relative to the original implementation. Given that DiffuSeq originally rescales time inputs in the range $[0, 1000]$, we rescale our input time-steps to preserve its pre-trained capabilities. After sampling a time step $t$, we rescale it to the target range using $
    t_{\text{input}} = \frac{t}{T} \times 1000,$
before feeding it into the model. 
% \begin{equation}
%     \mathcal{L} = \sum_{t=2}^T \|\mathbf{z}_0 - f_{\theta}(\mathbf{z}_t,t)\|^2 + \mathrm{CE}(decoder(EMB(\mathbf{w}^y)),\mathbf{w}^y),
% \end{equation}

% \begin{equation}
%     \mathcal{L} = \mathcal{L}_{\text{MSE}}(\mathbf{z}_0, \hat{\mathbf{z}}_0) + \mathcal{L}_{\text{CE}}(\text{Text}, \text{Decoder}({\mathbf{z}}_0)).
% \end{equation}

However, we also introduce several modifications. 
In addition to the primary loss function used in Diffuseq, we add a regularization loss term, calculated by the mean square loss of $z_0$ predicted by original diffusion model and the flow matching model, referring \citet{fan2025online}. This regularization loss can prevent policy collapse and maintains generative diversity by constraining the deviation of the fine-tuned model from the pre-trained reference model. Details in Table \ref{tab: Ablation of regulation}.

Another small but important detail is that the performance of the student model improves significantly when the time step sampling strategy is changed from loss-aware to uniform. As shown in Table~\ref{tab:loss_comparison}, where $q_0$ denotes the first 25\% of time steps (proximal to clean data) and $q_3$ represents the final 25\% (proximal to noise), the loss distribution of the Diffusion model remains relatively stable. 
In contrast, the Flow Matching model exhibits substantial disparity across different intervals. 
Therefore, employing a loss-aware strategy would cause the Flow Matching model to focus excessively on the $q_3$ interval during training, inevitably leading to suboptimal performance. 

Furthermore, we reduce the number of discrete time steps during training. We will randomly sample t from a discrete set of $T=20$ steps instead of the original 2000 steps to align with our few-step sampling target. We conclude that sampling t from moderately larger T is optimal for few-step generation. Detailed comparison is listed in the appendix \ref{sec:appendix_training_steps}.

\section{Results}
To validate the efficacy of our proposed approach, we perform extensive evaluations on standard sequence-to-sequence benchmarks, same as those used by DiffuSeq \citep{gong2023diffuseq}. Our experimental results highlight the capability of our method to drastically enhance the sampling efficiency of continuous Diffusion Language Models without sacrificing model performance.

\subsection{Tasks and Evaluation Metrics}

Our evaluation encompasses three primary Seq2Seq tasks: Question Generation, Text Simplification, and Paraphrase Generation. We utilize established datasets for each domain:
\textbf{Question Generation (QG):} We employ the Quasar-T dataset \citep{dhingra2017quasar}, 117k training, 2k validation, 10k test; \textbf{Text Simplification (TS):} We use the Wiki-Auto dataset \citep{jiang2020neural}, 678k training, 2k validation, 5k testing; \textbf{Paraphrase Generation (PP):} We adopt widely used QQP sourced
from the community question answering forum Quora \footnote{\url{https://www.kaggle.com/c/quora-question-pairs}} , 145k training, 2k validation, 2k test.

We assess generation performance using four standard metrics covering both quality and diversity: \textbf{BLEU} \citep{papineni2002bleu} measures n-gram precision overlap between generated and reference sequences; \textbf{ROUGE-L} \citep{lin2004rouge} evaluates the longest common subsequence to capture recall-oriented similarity; \textbf{BERTScore} \citep{zhang2020bertscore} utilizes contextual embeddings to compute semantic similarity, addressing the limitations of rigid n-gram matching. \textbf{Dist-1} calculates the ratio of unique unigrams within generated sequences to evaluate lexical diversity at the sentence level.

% \textbf{Sequence-Level Diversity:} Following \citep{gong2022diffuseq}, we quantify diversity across multiple generated samples for the same input using \textbf{Self-BLEU} \citep{zhu2018texygen} (lower is better, indicating less overlap) and \textbf{Div-4} (higher is better, indicating a higher proportion of unique 4-grams).

% \begin{table*}[t]
% \centering
% \renewcommand{\arraystretch}{1.1} %稍微增加行高，让表格不那么拥挤
% % 定义列格式：
% % l: 左对齐
% % c: 居中
% % |: 竖线
% \begin{tabular}{ll | ccc | c | c}
% \toprule
% Task & Methods & BLEU$\uparrow$ & R-L$\uparrow$ & Score$\uparrow$ & dist-1$\uparrow$ & Len \\
% \midrule

% \multirow{5}{*}{\shortstack[l]{Question \\Generation}} 
% & \textsc{Flow}(step=10) & \textbf{0.1596} & 0.3484 & \textbf{0.5898} & \textbf{0.9206} & 9.08 \\
% & \textsc{Flow}(step=5) & \textbf{0.1596} & 0.3484 & \textbf{0.5898} & \textbf{0.9206} & 9.08 \\
%  & FlowLM(step=5) & \textbf{0.1596} & 0.3484 & \textbf{0.5898} & \textbf{0.9206} & 9.08 \\
%  & FlowLM(step=3) & 0.1595 & 0.3489 & 0.5878 & 0.9169 & 8.88 \\
%  & FlowLM(step=1) & 0.1524 & \textbf{0.3550} & 0.5713 & 0.8411 & 8.57 \\
% \hline \hline
% \end{tabular}
% \caption{Ablation experiment comparison}
% \label{tab:Ablation-experiment}

% \end{table*}

\begin{figure*}[t]
    \centering
    
    % =======================================================
    % 全局绘图样式设置
    % =======================================================
    \pgfplotsset{
        every axis/.style={
            width=\linewidth, % 宽度跟随子图宽度
            height=4.5cm,     % 【建议】高度稍微改小一点，防止比例过长
            grid=major,
            grid style={dashed, gray!30},
            xlabel={MBR (n)},
            xmin=1, xmax=10,
            xtick={1,3,5,7,9}, % 【建议】减少x轴刻度显示，防止重叠
            % xtick={1,2,3,4,5,6,7,8,9,10}, % 原来的刻度在窄图里会太挤
            tick label style={font=\tiny}, % 【修改】字体改小
            scaled y ticks=false,
            yticklabel style={
                /pgf/number format/.cd,
                fixed,
                fixed zerofill,
                precision=2, % 【建议】精度改小一点，防止Y轴数字太长挡住图
            },
            label style={font=\scriptsize}, % 【修改】标签字体改小
            title style={font=\scriptsize, at={(0.5,0.95)}}, % 稍微调整标题位置
        }
    }

    % =======================================================
    % 图例 (Legend)
    % =======================================================
    \begin{tikzpicture}
        \begin{customlegend}[
            legend columns=-1,
            legend style={
                draw=none,
                /tikz/every even column/.append style={column sep=0.3cm}
            },
            legend entries={DiffuSeq, 
            DiffuSeq (DPM step=10),
            FlowLM (step=5), FlowLM (step=3),
            FlowLM (step=1),
            }
        ]
        \addlegendimage{color=blue!70!black, mark=*, thick}
        \addlegendimage{color=yellow!90!black, mark=square*, thick, dashed}
        \addlegendimage{color=orange!90!black, mark=square*, thick, dashed}
        \addlegendimage{color=green!90!black, mark=square*, thick, dashed}
        \addlegendimage{color=red!90!black, mark=square*, thick, dashed}
        \end{customlegend}
    \end{tikzpicture}
    
    \vspace{-0.2cm} % 调整图例和图之间的距离

    % --- 子图 (a) BLEU ---
    % 【修改】宽度设为 0.24 (4个图并排)
    \begin{subfigure}{0.24\linewidth}
        \centering
        \begin{tikzpicture}
            \begin{axis}[title={BLEU}] % 【建议】把Ylabel改成Title或者放上面，省空间
            % DiffuSeq
            \addplot[color=blue!70!black, mark=*, thick] coordinates {
                (1, 0.15138) (2, 0.15316) (3, 0.15850) (4, 0.16103) (5, 0.16220) 
                (6, 0.16340) (7, 0.16490) (8, 0.16528) (9, 0.16542) (10, 0.16538)
            };
            % FlowLM 5
            \addplot[color=orange!90!black, mark=square*, thick, dashed] coordinates {
                (1, 0.15958) (2, 0.16076) (3, 0.16418) (4, 0.16557) (5, 0.16686) 
                (6, 0.16729) (7, 0.16773) (8, 0.16778) (9, 0.16821) (10, 0.16866)
            };
            % FlowLM 3
            \addplot[color=green!90!black, mark=square*, thick, dashed] coordinates {
                (1, 0.16003) (2, 0.16001) (3, 0.16309) (4, 0.16390) (5, 0.16490) 
                (6, 0.16537) (7, 0.16621) (8, 0.16638) (9, 0.16704) (10, 0.16707)
            };
            % FlowLM 1
            \addplot[color=red!90!black, mark=square*, thick, dashed] coordinates {
                (1, 0.15272) (2, 0.15263) (3, 0.15193) (4, 0.15353) (5, 0.15410) 
                (6, 0.15401) (7, 0.15403) (8, 0.15427) (9, 0.15425) (10, 0.15402)
            };
            % DPM
            \addplot[color=yellow!90!black, mark=square*, thick, dashed] coordinates {
                (1, 0.14339) (2, 0.14405) (3, 0.14611) (4, 0.14708) (5, 0.14795) 
                (6, 0.14846) (7, 0.14899) (8, 0.14914) (9, 0.14905) (10, 0.14869)
            };
            \end{axis}
        \end{tikzpicture}
        \caption{BLEU}
    \end{subfigure}%
    \hfill % 【重要】这里的hfill负责平均分配间距
    % --- 子图 (b) ROUGE-L ---
    \begin{subfigure}{0.24\linewidth}
        \centering
        \begin{tikzpicture}
            \begin{axis}[title={ROUGE-L}]
            \addplot[color=blue!70!black, mark=*, thick] coordinates {
                (1, 0.34774) (2, 0.34880) (3, 0.35750) (4, 0.36107) (5, 0.36211) 
                (6, 0.36277) (7, 0.36379) (8, 0.36440) (9, 0.36477) (10, 0.36592)
            };
            \addplot[color=orange!90!black, mark=square*, thick, dashed] coordinates {
                 (1, 0.34839) (2, 0.34998) (3, 0.35489) (4, 0.35777) (5, 0.35915) 
                (6, 0.36016) (7, 0.36079) (8, 0.36124) (9, 0.36233) (10, 0.36293)
            };
            \addplot[color=green!90!black, mark=square*, thick, dashed] coordinates {
                (1, 0.34985) (2, 0.35082) (3, 0.35486) (4, 0.35709) (5, 0.35772) 
                (6, 0.35890) (7, 0.36001) (8, 0.36101) (9, 0.36168) (10, 0.36204)
            };
            \addplot[color=red!90!black, mark=square*, thick, dashed] coordinates {
                 (1, 0.35488) (2, 0.35512) (3, 0.35507) (4, 0.35657) (5, 0.35746) 
                (6, 0.35674) (7, 0.35751) (8, 0.35771) (9, 0.35776) (10, 0.35750)
            };
            \addplot[color=yellow!90!black, mark=square*, thick, dashed] coordinates {
                 (1, 0.35126) (2, 0.35121) (3, 0.35523) (4, 0.35540) (5, 0.35709) 
                (6, 0.35756) (7, 0.35815) (8, 0.35842) (9, 0.35878) (10, 0.35860)
            };
            \end{axis}
        \end{tikzpicture}
        \caption{ROUGE-L}
    \end{subfigure}%
    \hfill
    % --- 子图 (c) BERTScore ---
    \begin{subfigure}{0.24\linewidth}
        \centering
        \begin{tikzpicture}
            \begin{axis}[title={BERTScore}]
            \addplot[color=blue!70!black, mark=*, thick] coordinates {
                (1, 0.58735) (2, 0.58802) (3, 0.59363) (4, 0.59725) (5, 0.59889) 
                (6, 0.60019) (7, 0.60111) (8, 0.60189) (9, 0.60292) (10, 0.60294)
            };
            \addplot[color=orange!90!black, mark=square*, thick, dashed] coordinates {
                (1, 0.58978) (2, 0.59108) (3, 0.59506) (4, 0.59662) (5, 0.59819) 
                (6, 0.59944) (7, 0.60027) (8, 0.60074) (9, 0.60147) (10, 0.60216)
            };
            \addplot[color=green!90!black, mark=square*, thick, dashed] coordinates {
                (1, 0.58797) (2, 0.58862) (3, 0.59159) (4, 0.59300) (5, 0.59430) 
                (6, 0.59521) (7, 0.59642) (8, 0.59658) (9, 0.59714) (10, 0.59807)
            };
            \addplot[color=red!90!black, mark=square*, thick, dashed] coordinates {
                (1, 0.57113) (2, 0.57065) (3, 0.57072) (4, 0.57200) (5, 0.57273) 
                (6, 0.57234) (7, 0.57245) (8, 0.57274) (9, 0.57299) (10, 0.57269)
            };
            \addplot[color=yellow!90!black, mark=square*, thick, dashed] coordinates {
                (1, 0.57401) (2, 0.57458) (3, 0.57606) (4, 0.57671) (5, 0.57734) 
                (6, 0.57820) (7, 0.57896) (8, 0.57895) (9, 0.57879) (10, 0.57887)
            };
            \end{axis}
        \end{tikzpicture}
        \caption{BERTScore}
    \end{subfigure}%
    \hfill
    % --- 子图 (d) Dist-1 ---
    \begin{subfigure}{0.24\linewidth}
        \centering
        \begin{tikzpicture}
            \begin{axis}[title={Dist-1}]
            \addplot[color=blue!70!black, mark=*, thick] coordinates {
                (1, 0.91257) (2, 0.90441) (3, 0.91311) (4, 0.91240) (5, 0.91158) 
                (6, 0.91018) (7, 0.90853) (8, 0.90709) (9, 0.90680) (10, 0.90630)
            };
            \addplot[color=orange!90!black, mark=square*, thick, dashed] coordinates {
                (1, 0.9206) (2, 0.9172) (3, 0.9196) (4, 0.9186) (5, 0.9173) 
                (6, 0.9171) (7, 0.9159) (8, 0.9150) (9, 0.9149) (10, 0.9147)
            };
            \addplot[color=green!90!black, mark=square*, thick, dashed] coordinates {
                (1, 0.91660) (2, 0.91533) (3, 0.91580) (4, 0.91460) (5, 0.91381) 
                (6, 0.91235) (7, 0.91219) (8, 0.91137) (9, 0.91109) (10, 0.91085)
            };
            \addplot[color=red!90!black, mark=square*, thick, dashed] coordinates {
                (1, 0.84301) (2, 0.84548) (3, 0.84186) (4, 0.84050) (5, 0.84047) 
                (6, 0.83916) (7, 0.83844) (8, 0.83854) (9, 0.83854) (10, 0.83869)
            };
            \addplot[color=yellow!90!black, mark=square*, thick, dashed] coordinates {
                (1, 0.86358) (2, 0.86083) (3, 0.86167) (4, 0.86104) (5, 0.86189) 
                (6, 0.86037) (7, 0.86080) (8, 0.86094) (9, 0.86049) (10, 0.86019)
            };
            \end{axis}
        \end{tikzpicture}
        \caption{Dist-1}
    \end{subfigure}

    \caption{Evaluation metrics on Question Generation using MBR decoding across 1 to 10 candidates. Comparison between FlowLM (6000 epochs, step=5, 3,1) and DiffuSeq (34000 epochs, step=2000, DPM step=10).}
    \label{fig:mbr_analysis}
    \centering
    \captionof{table}{Average time needed for inference per sample (seconds), exclude data loading and decoding.}
    \vspace{0.5em}
    \begin{tabular}{lcccc} 
        \toprule
        Model & DiffuSeq (step=2000)  & FlowLM (step=1) & FlowLM (step=3) & FlowLM (step=5)\\
        Time (s) & 0.7264 & 0.00014 & 0.00038 & 0.000572\\
        \midrule
        Model & DiffuSeq (DPM-step=10) & FMSeq (step=1) & FMSeq (step=3) & FMSeq (step=5) \\
        Time (s) & 0.0012 & 0.00054 & 0.00159 & 0.002101 \\ 
        \bottomrule
    \end{tabular}
    \label{tab:sampling-time}
\end{figure*}

\begin{figure*}[t] 
    \centering
    
    % 解决双括号 ((a)) 的关键设置
    \captionsetup[subfigure]{labelformat=simple}
    \renewcommand\thesubfigure{(\alph{subfigure})}
    
    % ==================== 第一部分：表格 ====================
    % 注意：去掉末尾括号外的句号
    \captionof{table}{Performance comparison on Question Generation task. We compare FlowLM (trained from scratch, same step as Diffusion-LM) and FlowLM (initialized from Diffusion-LM) with Diffuseq. Scr: Trained from scratch. FT: Finetuned.}
    \label{tab:combined_results}
    
    \renewcommand{\arraystretch}{1.1}
    \setlength{\tabcolsep}{6pt}
    \footnotesize 
    
    \begin{tabular}{lll | ccc | c|c}
        \toprule
        Tasks & Type & Methods & BLEU$\uparrow$ & R-L$\uparrow$ & BERTScore$\uparrow$ & dist-1$\uparrow$ & Training epochs\\
        \midrule
        \multirow{5}{*}{\shortstack[l]{Question \\Generation}} 
         & Multi-step & DiffuSeq (step=2000) & 0.1527 & 0.3474 & 0.5864 & 0.9113 & 34000\\
         \cmidrule(lr){2-8} 
         & \multirow{7}{*}{Few-step}   & FlowLM(FT, step=5) & 0.1596 & 0.3484 & 0.5898 & 0.9206 &6000\\
         & & FlowLM(FT, step=1) & 0.1524 & 0.3550 & 0.5713 & 0.8411 &6000\\
         & &FlowLM(Scr, step=5) & 0.0000 & 0.0000 & 0.0001 & 0.0002 &6000 \\
         & & FlowLM(Scr, step=1) & 0.0000 & 0.0000 & 0.0085 & 0.0245 &6000\\

         & & FlowLM(FT, step=5) & 0.1648 & 0.3510 & 0.5897 & 0.9222 &15000\\
           & & FlowLM(FT, step=1) & 0.1634 & 0.3549 & 0.5939 & 0.9006 &15000\\   
           & & FlowLM(Scr step=5) & 0.1689 & 0.3568 & 0.5982 & 0.9141 &34000 \\
           % & & \textsc{FlowLM1}(scratch, step=3) & 0.1685 & 0.3585 & 0.5981 & 0.9119\\
           & & FlowLM(Scr, step=1) & 0.1624 & 0.3650 & 0.5932 & 0.8790 &34000\\
          
        \bottomrule
    \end{tabular}
    \setcaptiontype{figure}

    % ==================== 第二部分：图片 ====================
    % 注意：修复了重复的 label
    \begin{subfigure}{0.45\linewidth}
        \centering
        \includegraphics[width=\linewidth]{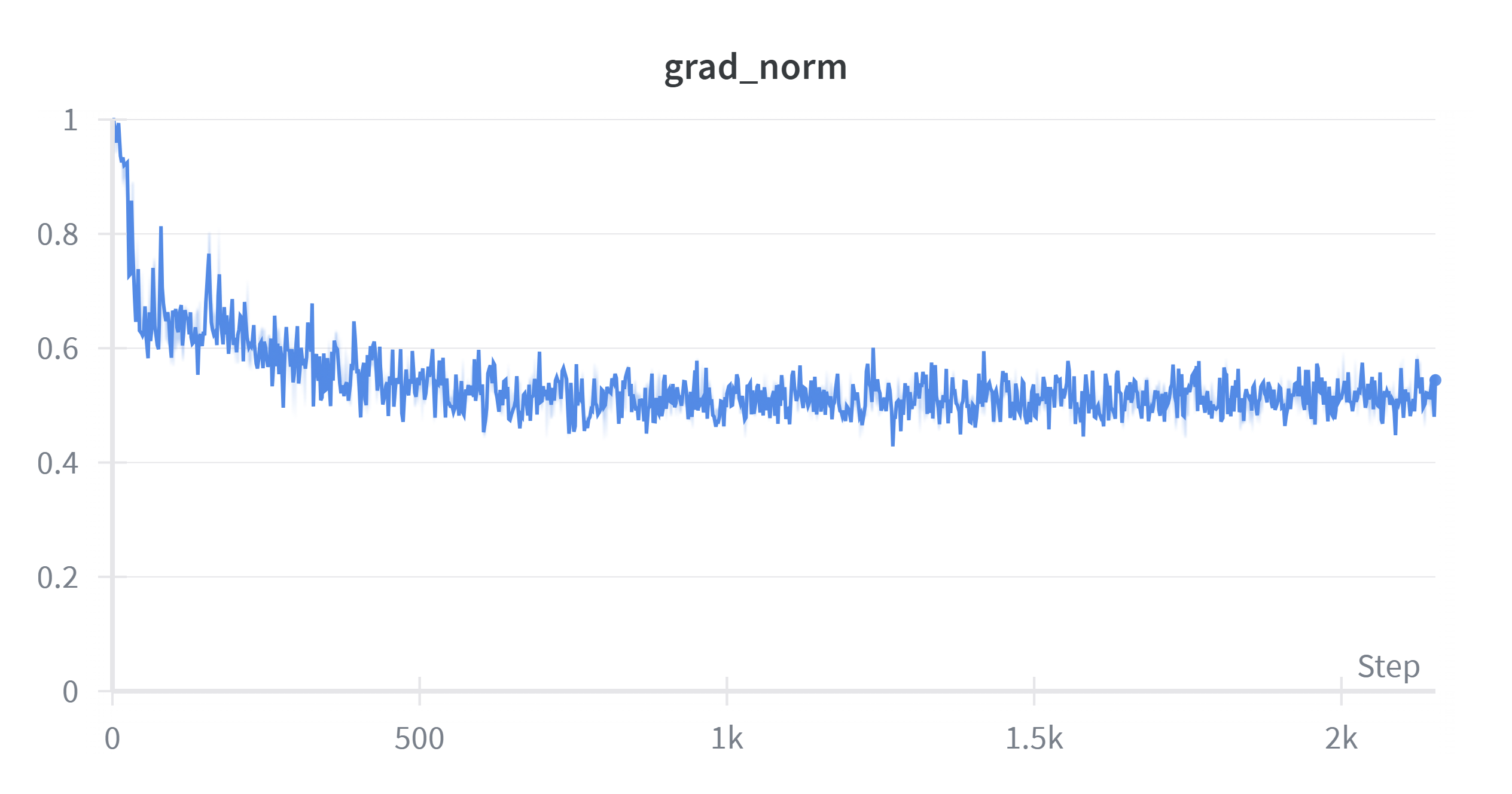}
        \caption{Gradient of x-pred FlowLM-Scr}
        \label{fig:grad_lm_scratch}
    \end{subfigure}
    \hfill
    \begin{subfigure}{0.45\linewidth}
    \centering
    \includegraphics[width=\linewidth]{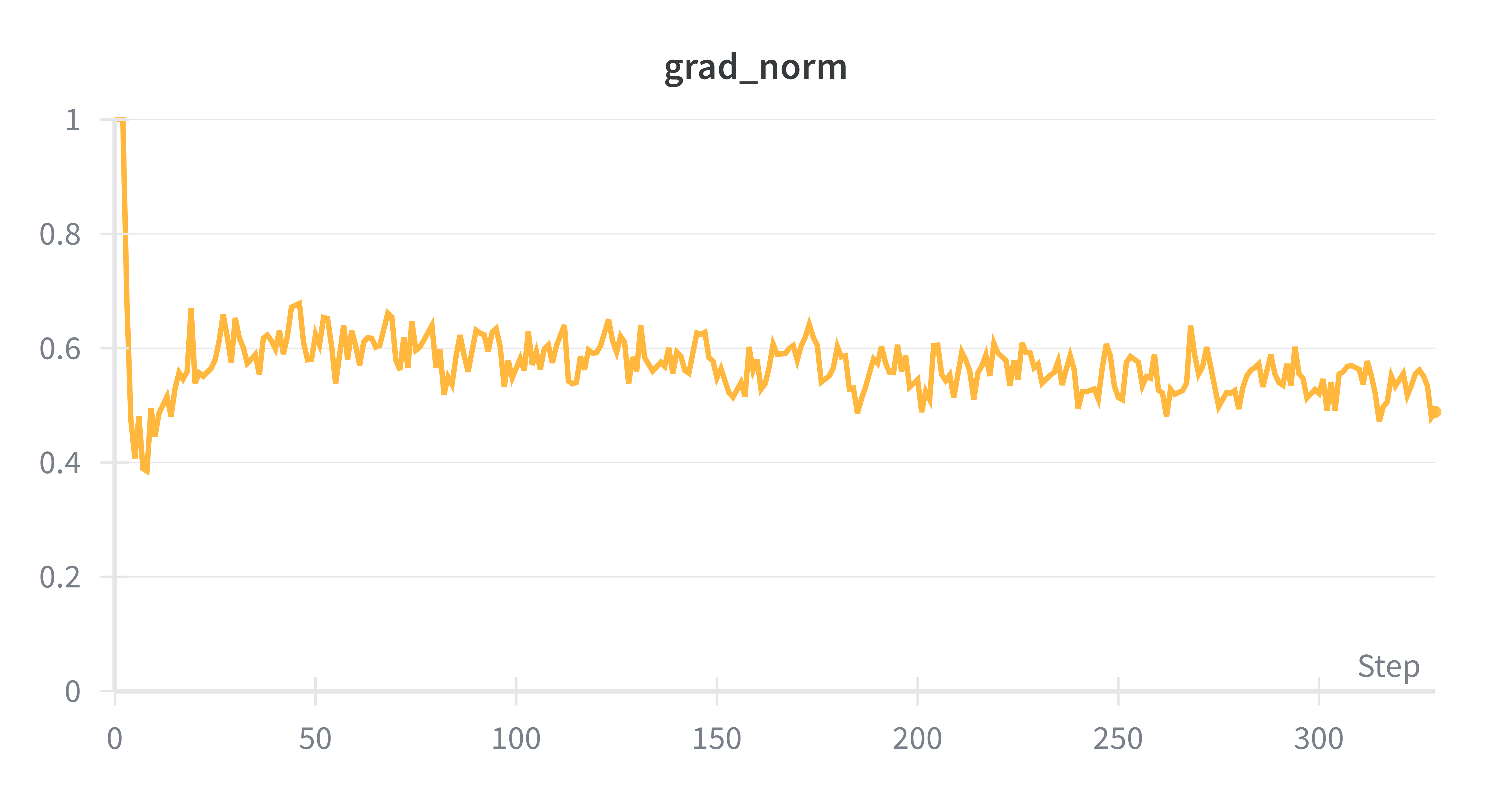}
    \caption{Gradient of x-pred FlowLM-FT}
    \label{fig:grad_lm_finetune}
    \end{subfigure}
    
    \begin{subfigure}{0.45\linewidth}
        \centering
        \includegraphics[width=\linewidth]{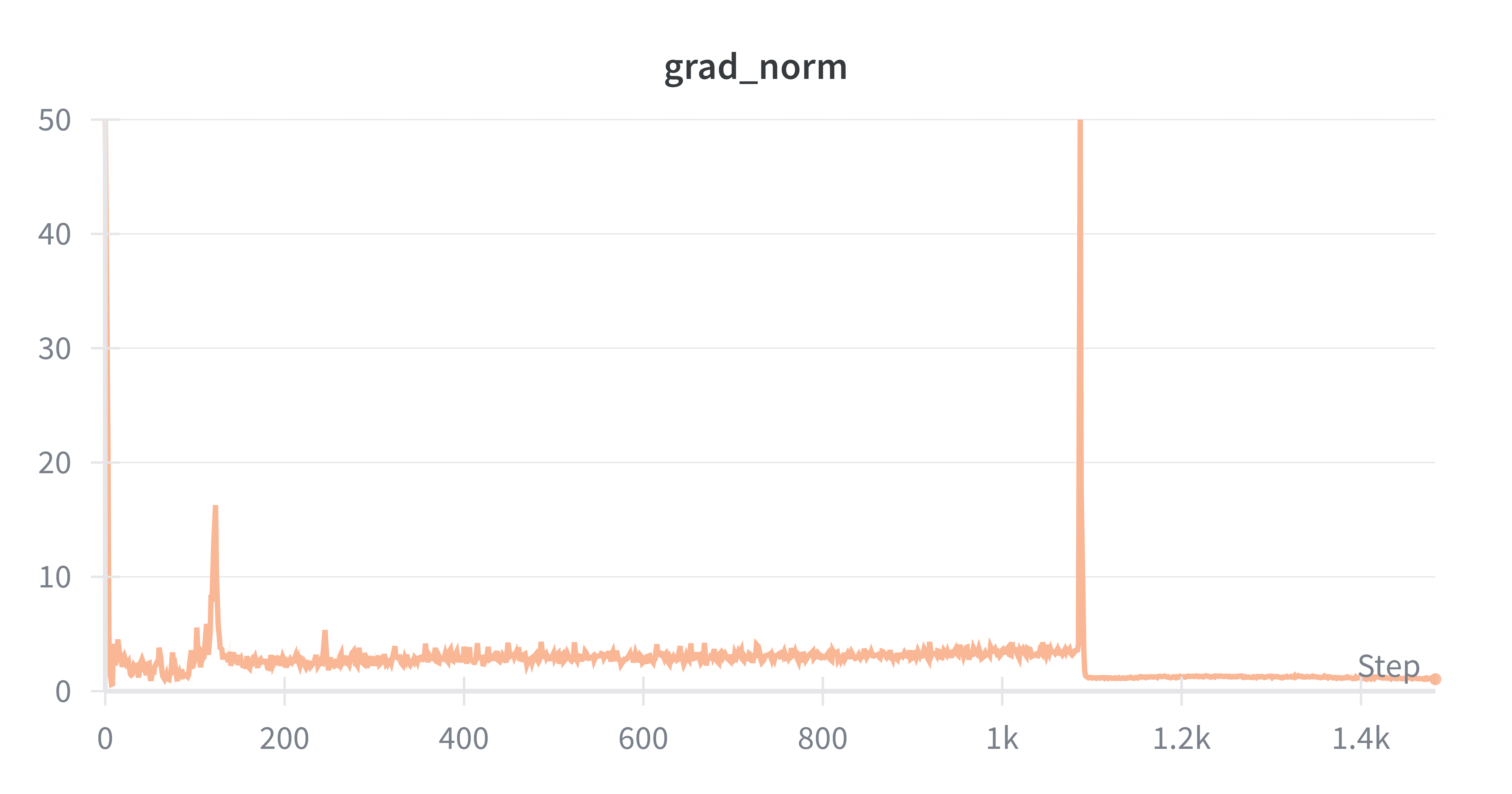}
        \caption{Gradient of v-pred FlowLM-Scr}
        \label{fig:grad_vpred_scratch}
    \end{subfigure}
    \hfill
    \begin{subfigure}{0.45\linewidth}
        \centering
        \includegraphics[width=\linewidth]{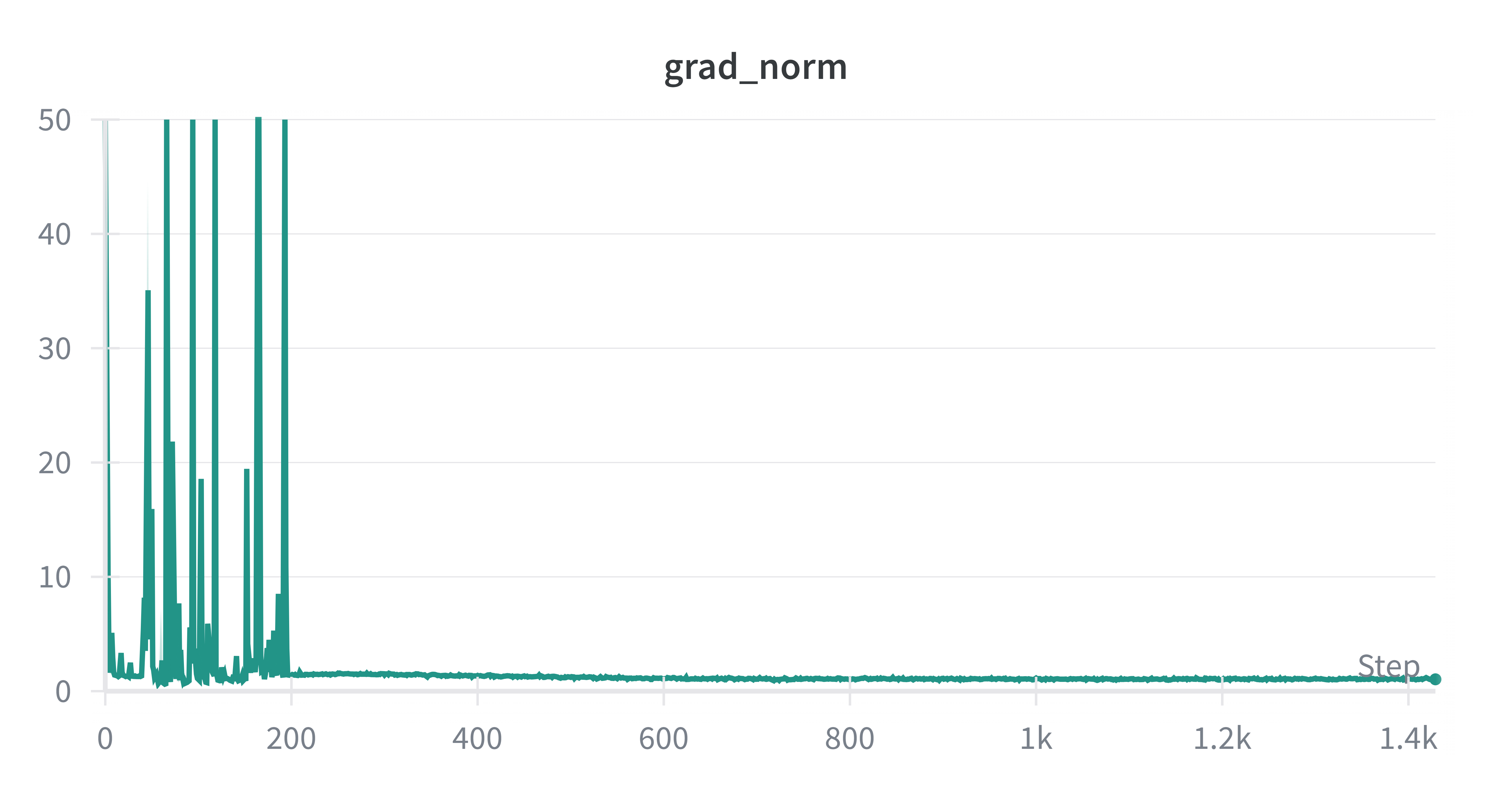}
        \caption{Gradient of v-pred FlowLM-FT}
        \label{fig:grad_vpred_finetune}
    \end{subfigure}
    
    % 注意：这里改成了 \caption (去掉了末尾多余的 s)
    \caption{Comparison of gradient during the training process of different methods (Note that maximum values are different). Scr: Trained from scratch. FT: Finetuned}
    \label{fig:Ablation study}
\end{figure*}

\subsection{Experiment results}

We establish DiffuSeq-V1 and the DPM-Solver \citep{lu2022dpm}accelerated versions of DiffuSeq-V2 as our comparative baselines. As listed in Table~\ref{tab:main_results}, across several tasks, our fine-tuned flow matching model achieves performance comparable to or even surpassing both the original 2000-step generation and the 10-step DPM-Solver with significantly fewer sampling steps. 

As shown in Appendix \ref{sec: More about FlowLM}, FlowLM achieves a nearly perfect straight trajectory. We recorded the inference time of different models across varying step counts. Experiments were conducted on the task of Question Generation with a fixed batch size of 500 on a single NVIDIA H100 GPU. As listed in Table \ref{tab:sampling-time}, our few-step generation approach demonstrates a massive advantage in overall inference efficiency.

FlowLM also alleviates the semantic drift sometimes observed in DiffuSeq, maintaining robustness even under extreme few-step constraints, as listed in Appendix \ref{sec:examples}.

We also compare FlowLM with other methods that accelerate sampling, including DLM-One \citep{chen2025dlm}, Rectified Flow\cite{liu2023flow}, PerfLow\cite{yan2024perflow}, and FMSeq \citep{liu2024enable} on the Question generation task. FlowLM outperforms these methods, demonstrating its superiority compared with other methods. Furthermore, we observe that models predicting velocity perform poorly.

To compare the quality of the text if multiple candidates are generated, we employ Minimum Bayes Risk (MBR) decoding \citep{koehn2004statistical}, generating candidate sets and filtering out low-probability outliers. Specifically, we generate candidate sets of sizes 1–10 and select the output that maximizes the expected BLEU utility relative to all members in the set.

Figure~\ref{fig:mbr_analysis} analyzes the impact of the MBR candidate size $N$ on generation performance. FlowLM (step=5) exhibits superior single-sample efficiency, outperforming the DiffuSeq baseline in BLEU at $N=1$. Furthermore, as shown in Figure~\ref{fig:mbr_analysis}(d), it maintains consistently higher lexical diversity. FlowLM (step=3) is very close to FlowLM (step=5), this suggests that FlowLM (step=3) offers the best balance between both sampling budget and performance. In contrast, the overall performance of DPM-solver is worse than that of FlowLM step=3 or 5.

Notably, the performance of FlowLM (step=1) remains nearly constant regardless of the candidate size. This phenomenon can be attributed to the reduced randomness in one-step generation. 
In contrast, DiffuSeq shows steeper performance gains as $N$ increases, eventually surpassing FlowLM in recall-oriented metrics (ROUGE-L, BERTScore). This phenomenon suggests that DiffuSeq relies on larger candidate pools to mitigate the high randomness introduced by its 2000-step sampling process.

However, we also observe certain anomalies. Specifically, in the Text Simplification task, both the DPM-Solver and FlowLM(step=1,3,5) exhibit suboptimal performance when MBR=1, lagging behind the original 2000-step baseline. We hypothesize that this is because the task presents greater challenges given the complexity of dataset and difficulty of simplification. It may cause a few-step generating model with such small parameter number (0.1B) struggle to match the performance of baseline diffusion model. However, as shown in Appendix \ref{fig:combined_results_ts}, when the candidate size increases, the gap narrows significantly, and our model performs much better than DPM-Solver(step=10). 
% % As the model condenses the trajectory into a single jump, it tends to produce highly deterministic outputs with relatively low diversity, thereby rendering the MBR re-ranking strategy ineffective. 

% Furthermore, we have tried to train on \citep{pile} to test long text sequence generation. It doesn't work on 0.1B model, due to limited resources, we doesn't try on larger model.

In conclusion, FlowLM demonstrates promising potential as a highly efficient few-step generation method. 
\subsection{Ablation experiment}
We conducted ablation studies to validate the effectiveness of our proposed method from two perspectives: generation performance and training stability. Specifically, we compared FlowLM against traditional flow matching models that use the standard \textbf{v-prediction} objective (both trained from scratch and initialized from the Diffusion-LM).

\textbf{Superior Performance of Both Finetuned and Trained From Scratch.}
As presented in Table~\ref{tab:main_results}, the finetuned FlowLM can reach the performance of original diffusion model with very few training epochs, when the model trained from scratch suffers from significant degradation. Finetuned model can also reach performance saturation with less than half of training epochs compared with training from scratch, both greatly outperforming original diffusion model.

\textbf{Training Stability Analysis.}
% In addition to performance, 
Figure~\ref{fig:Ablation study} demonstrates the superior convergence properties of our method compared to the traditional v-prediction target of flow matching. As shown in Fig.~\ref{fig:Ablation study}(a, b), our method maintains remarkably stable gradient norms around 0.6 without requiring aggressive clipping. This indicates that preserving the $z_0$-prediction objective aligns well with the model's architecture, allowing for seamless and stable optimization.
In contrast, the standard v-prediction objective exhibits severe instabilities. When trained from scratch (Fig.~\ref{fig:Ablation study}(c)), the model suffers from catastrophic divergence in later stages, suggesting high sensitivity to random initialization. When finetuned (Fig.~\ref{fig:Ablation study}(d)), it faces severe  early-stage oscillations. This is attributed to objective misalignment: forcing weights optimized for data output ($z_0$) to suddenly predict velocity ($v_t$) disrupts effective capability inheritance.

Furthermore, we provide the ablation study results on the influence of training epochs on the finetuned FlowLM model ability, which show a stable improvement. Training for more epochs yields even better results. Details are listed in figure \ref{fig:qg_ablation_original_size} and \ref{fig:para_ablation_original_size}.
We also provide detailed ablation experiments on our training hyperparameter and new regularization loss term, 
% we conclude that sampling t from moderately larger T is the best if your target is few-step generation. And rescale the time input to the same as original diffusion model is the best. 
details listed in Table \ref{tab: Ablation of rescale}, Table \ref{tab: Ablation of regulation}.

In conclusion, if finetuned from Diffusion-LM, FlowLM avoids both the early-stage shock of objective switching and the optimization difficulty associated with velocity prediction, and finetuned model requires much fewer training epochs. Given sufficient training time, both FlowLM trained from scratch and the finetuned model greatly outperform diffusion model, demonstrating the validity of this new training objective for flow matching.

\section{Related Works}

To push inference efficiency to the extreme (i.e., one-step generation), recent works have explored distilling pre-trained diffusion or flow models. \textbf{DLM-One} \citep{chen2025dlm} adapts Score Distillation techniques \citep{zhou2024score} to Diffusion Language Models. The student is optimized by minimizing a Model-based Explicit Score Matching loss, and further stabilizes this process with adversarial regularization \citep{zhou2025adversarial}, achieving significant speedups in text generation tasks. As this distillation should be trained through two stages and it contains  adversarial generative loss, it will cause a harder and more unstable training process compared to our method.

\textbf{Improved Mean Flows} \citep{geng2025improved} tackles one-step generation by  reformulating the flow matching objective and expressing the instantaneous velocity $\mathbf{v}$ explicitly in terms of the average velocity $\mathbf{u}$. This reformulation creates a standard regression problem that stabilizes training without requiring distillation from a separate teacher: $
    \mathbf{v}(\mathbf{z}_t) = \mathbf{u}(\mathbf{z}_t) + (t-r) \frac{d}{dt}\mathbf{u}(\mathbf{z}_t).$

FlowLM is similar to IMF as both of them sample with average velocity. However, FlowLM is easier to train as it does not require learning the average velocity between two time steps, but only needs to predict $z_0$. As IMF cannot be directly used to speed up diffusion language model, we did not conduct extra experiments to compare between them.
% though we believe our method can also gain great result if trained from scratch. We will do this in following works.

Other relevant approaches exist to accelerate sampling. \textbf{Reflow} \cite{liu2023flow} introduces an iterative approach to straighten diffusion trajectories, theoretically enabling one-step generation. \textbf{Perflow} \cite{yan2024perflow} enhances this by applying piecewise rectification across time windows, serving as a plug-and-play accelerator. \citet{hu2024flow} predicts and samples with instantaneous velocity directly, which is empirically shown to be ineffective. \textbf{FMSeq} \citep{liu2024enable} is the most similar method as it also predicts \textsc{$z_0$} and sample with estimated velocity. However, they didn't reveal the huge difference between training with x-pred and v-pred. Furthermore, they treat the estimated velocity as instantaneous velocity but not average velocity, which fails to capture the underlying mechanism of this sampling formula.  Our experiments using the Improved Mean Flow method, where $\mathbf{v}_{average}= \frac{\mathbf{z}_t - \mathbf{z}_{0,pred}}{t}$ was treated as the instantaneous velocity in training, yielded poor results. Moreover,  our attempt at Reflow model and Perflow model using instantaneous velocity prediction also failed. There are also differences in training and sampling, as listed in table \ref{tab: comparison}.

Our ablation study reveals that maintaining the $\mathbf{z}_0$-prediction objective is crucial for the success of FlowLM, 
% whereas v-prediction leads to training instability and poor ability. 
this finding also resonates with \citet{li2025back}. 
They argue that predicting the clean data $\mathbf{z}_0$ is fundamentally distinct from predicting noise or velocity, as $\mathbf{z}_0$ lies on a low-dimensional data manifold while $\mathbf{v}$ is inherently high-dimensional and off-manifold. 
Consequently, $\mathbf{z}_0$-prediction is a more tractable objective for neural networks, particularly when fine-tuning or operating in high-dimensional spaces. 
Our experiment results in the text domain provide further evidence for this perspective, but we find using x-loss instead of v-loss is better in our tasks, detailed comparison provided in Table \ref{tab:compare with HKM}.

\section{Limitations and Future Work}

While FlowLM successfully bridges the gap between diffusion capability and flow matching efficiency, several limitations remain. 

First, 
% as a fine-tuning framework, the upper bound of FlowLM's generation quality is inherently constrained by the pre-trained Diffusion-LM. And 
% as we observed a performance drop in specific tasks like Text Simplification, this suggests that maybe 
for some tasks requiring highly fine-grained edits, few-step generation might still struggle to capture all subtle dependencies with such small parameter (0.1B for FlowLM), necessitating further research to solve this issue or further test on larger models.

Secondly, this method is currently applicable only in continuous diffusion language model, but cannot be directly applied to discrete diffusion language models.  
% Secondly, similar to other distillation techniques, compressing the sampling trajectory may lead to a slight reduction in generation diversity, a trade-off that future work could address through more advanced regularization techniques.

In the future we will try to solve the limitations mentioned above, focusing on achieving few-step generation in discrete diffusion models.

\section{Conclusion}

In this paper, we presented FlowLM, a novel and efficient framework that transforms diffusion language models into flow language models via fine-tuning. 
By directly predicting clean data distribution and sampling with estimated average velocity, FlowLM re-aligns the generative trajectory to a straight flow path. Our approach can effectively increase the sampling speed without sacrificing sampling quality with few training epochs, and alleviates the semantic drift sometimes observed in DiffuSeq , maintaining robustness. Furthermore, finetuned model reaches performance saturation with less than half of training epochs compared with training from scratch, both greatly outperforming DiffuSeq. We attribute these to the global guidance provided by average velocity.

We show in Table \ref{tab:compare with HKM} that for text generation, training with x-pred and x-loss is better than x-pred and v-loss. Furthermore, we conclude that v-loss excels in continuous domains requiring high-precision refinement. And x-loss is superior for discrete-output tasks, as the final rounding step provides tolerance for minor latent deviations, rendering the additional focus on clean data within the loss term unnecessary. 

\section{Impact statement}

This paper presents work whose goal is to advance the field of Machine Learning by improving the efficiency and quality of diffusion-based language generation with the technique of flow matching finetuning. There are many potential societal consequences of our work, none which we feel must be
specifically highlighted here


\bibliography{example_paper}
\bibliographystyle{icml2025}

%%%%%%%%%%%%%%%%%%%%%%%%%%%%%%%%%%%%%%%%%%%%%%%%%%%%%%%%%%%%%%%%%%%%%%%%%%%%%%%
%%%%%%%%%%%%%%%%%%%%%%%%%%%%%%%%%%%%%%%%%%%%%%%%%%%%%%%%%%%%%%%%%%%%%%%%%%%%%%%
% APPENDIX
%%%%%%%%%%%%%%%%%%%%%%%%%%%%%%%%%%%%%%%%%%%%%%%%%%%%%%%%%%%%%%%%%%%%%%%%%%%%%%%
%%%%%%%%%%%%%%%%%%%%%%%%%%%%%%%%%%%%%%%%%%%%%%%%%%%%%%%%%%%%%%%%%%%%%%%%%%%%%%%
\newpage
\appendix
\onecolumn

% You can have as much text here as you want. The main body must be at most $8$ pages long.
% For the final version, one more page can be added.
% If you want, you can use an appendix like this one.  

% The $\mathtt{\backslash onecolumn}$ command above can be kept in place if you prefer a one-column appendix, or can be removed if you prefer a two-column appendix.  Apart from this possible change, the style (font size, spacing, margins, page numbering, etc.) should be kept the same as the main body.
%%%%%%%%%%%%%%%%%%%%%%%%%%%%%%%%%%%%%%%%%%%%%%%%%%%%%%%%%%%%%%%%%%%%%%%%%%%%%%%
%%%%%%%%%%%%%%%%%%%%%%%%%%%%%%%%%%%%%%%%%%%%%%%%%%%%%%%%%%%%%%%%%%%%%%%%%%%%%%%

\section{More about FlowLM}
\label{sec: More about FlowLM}
\begin{figure}[H]
    \centering
    \includegraphics[width=0.7\linewidth]{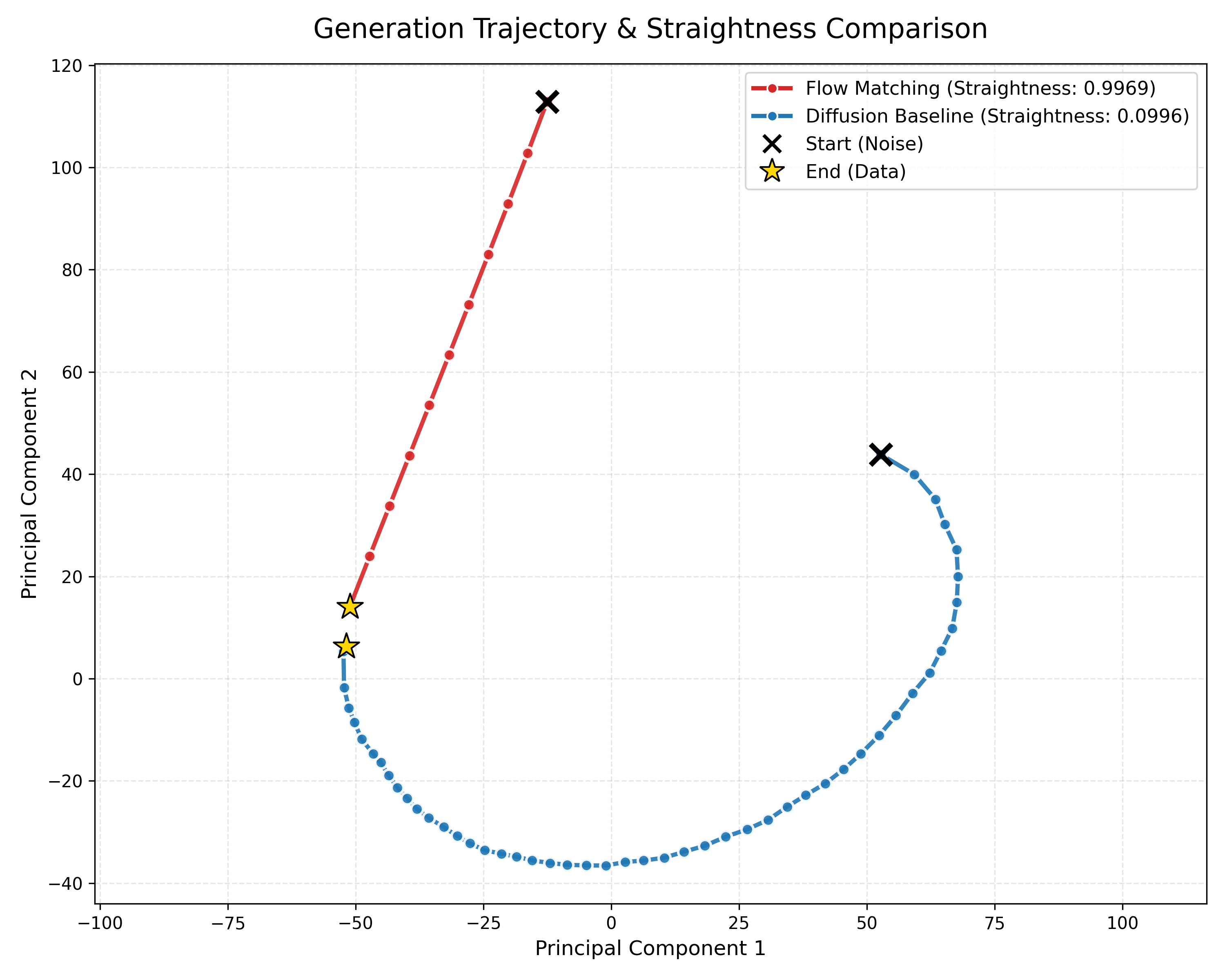}
    \caption{Visualization of generation trajectories in 2D PCA space.}
    \label{fig:trajectory}
\end{figure}
While the baseline diffusion model follows a curved path (blue, straightness=$0.0996$), our method achieves a nearly perfect linear trajectory (red, straightness=$\mathbf{0.9969}$). 
This straightened path minimizes truncation error during ODE solving, enabling efficient few-step generation.

\begin{figure}[H]
    \centering
    \includegraphics[width=0.8\linewidth]{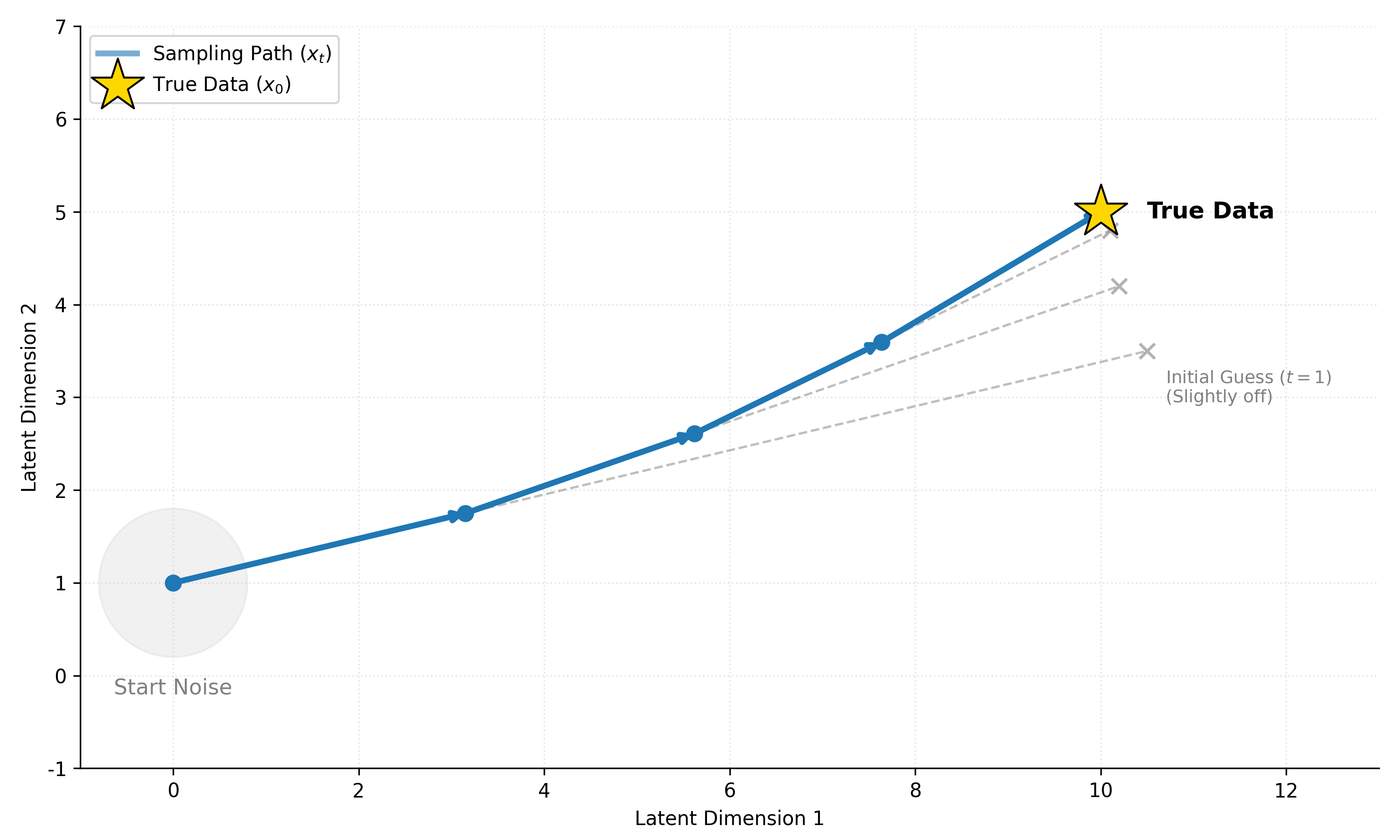}
    \caption{Schematic diagram of sampling process}
    \label{fig:enter-label}
\end{figure}

\begin{table}[H] % arXiv 建议用 htbp，尽量不用 H
    \centering
    \caption{Average time needed for sampling per sample (seconds), exclude data loading.}
    \begin{tabular}{lcccc} 
        \toprule
        Model & DiffuSeq (step=2000)  & FlowLM (step=1) & FlowLM (step=3) &FlowLM (step=5)\\
        Time (s) & 0.7264 & 0.00014 & 0.00038 & 0.000572\\
        \midrule
        Model & DiffuSeq (DPM-step=10) & FMSeq (step=1) & FMSeq (step=3) & FMSeq (step=5) \\
        Time (s) & 0.0012 & 0.00054 & 0.00159 & 0.002101 \\ % 补了一个占位数据
        \bottomrule
    \end{tabular}
\end{table}

\begin{table}[H]
    \centering
    \caption{Detailed comparison of DiffuSeq, FlowLM, and FMSeq (Transposed)}
    \label{tab:comparison_transposed}
    \begin{tabular}{lcccccccc} % 7列：1列模型 + 6列对比项
        \toprule
        \textbf{Model} & \makecell{\textbf{Time}\\ \textbf{sample}} & \makecell{\textbf{Pred}\\ \textbf{objective}}  & \makecell{\textbf{Num}\\ \textbf{timesteps}} &\makecell{\textbf{Regularization}\\ \textbf{loss}} & \makecell{\textbf{Time}\\ \textbf{difference}} &\makecell{\textbf{Self}\\ \textbf{conditioning}}& \textbf{Speed} \\ 
        \midrule
        DiffuSeq & loss-aware   & x-pred &  2000& No &No & No  & Slow \\
        FlowLM  & uniform  & x-pred &  20  &Yes & No  & No & Faster \\
        FMSeq    & loss-aware & x-pred & 2000 &No &Yes & Yes & Fast  \\
        \bottomrule
    \end{tabular}
    \label{tab: comparison}
\end{table}
Here we provide a detailed comparison between DiffuSeq\citep{gong2023diffuseq}, our method FlowLM, and FMseq \citep{liu2024enable} as they are the most similar. Num-timesteps here refers to the number of time steps we sample from during training. Time difference means FMSeq use $min(1,t+0.5)$ as time-step input in the process of sampling.

The results in paper demonstrate the effectiveness of the $z_0$-prediction strategy. Formally, the velocity at time $t$ is defined as:
\begin{equation}
    v(z_t, t) = \frac{z_t - \hat{z}_0(z_t, t)}{t}
\end{equation}
where $\hat{z}_0(z_t, t)$ is the model's prediction of the clean data given the current noisy state $z_t$ and t. Unlike standard Flow Matching which integrates local velocity tangents, this "shooting" method provides several advantages:

\begin{itemize}
    \item \textbf{Error Compensation:}
    Though the path of our few-step sampling has been very straight, some deviation may inevitably exist. As shown in the trajectory visualization, initial predictions $\hat{x}_0$ at high noise levels ($t \approx 1$) may contain bias. However, because the model re-estimates the endpoint at every iteration, the sampling trajectory is not bound to early errors. Each step effectively estimates velocity vector toward the most current estimate of the data manifold, resulting in a self-correcting curved path.
    
    \item \textbf{Global Anchoring:} By defining velocity relative to a global target $\hat{x}_0$ rather than a local derivative, the integration becomes more robust to large step sizes $\Delta t$. This explains the superior performance of our model in few-step regimes (e.g., $N=3$ or $N=5$), where discretization errors typically cause standard ODE solvers to drift significantly from the true data distribution.
    
    \item \textbf{Dynamic Contracting Flow:} The $1/t$ scaling factor ensures that as the sampling nears completion ($t \to 0$), the "pull" toward the predicted clean data becomes increasingly dominant. This ensures that the final samples $z_0$ are sharp and well-aligned with the learned vocabulary embeddings, reducing the likelihood of generating out-of-distribution or "blurry" latent representations.
     \item \textbf{Stability in training:} This has been shown in ablation experiment results.
     \item \textbf{Faster sampling speed:}
     As shown in Table \ref{tab:sampling-time}.
\end{itemize}

Through this mechanism, the model achieves a balance between generation speed and sample quality, successfully mapping Gaussian noise to complex text structures with minimal integration drift.

\newpage
\FloatBarrier

\begin{table*}[t]
\footnotesize
\centering
\caption{Comparison between x-pred x-loss and x-pred, v-loss in FlowLM\textbf{(mbr=1)} using \textbf{uniform time sampling}. The best results of few-step generation model are \textbf{bold}.}
\renewcommand{\arraystretch}{1.1} %稍微增加行高，让表格不那么拥挤
% 定义列格式：
% l: 左对齐
% c: 居中
% |: 竖线
\begin{tabular}{lll | ccc | cc }
\toprule
Tasks & Type &Methods & BLEU$\uparrow$ & R-L$\uparrow$ & BERTScore$\uparrow$ & dist-1$\uparrow$ &Training epoch\\
\midrule

\multirow{6}{*}{\shortstack[l]{Question \\Generation}} 
 &\multirow{6}{*}{Few-step} & FlowLM(x-pred,v-loss, step=5) & 0.1557 & 0.3468 & 0.5845 & 0.9168 & 6000  \\
 & & FlowLM(x-pred,v-loss, step=3) & 0.1559 & 0.3480 & 0.5822 & 0.9149 &6000\\
 & & FlowLM(x-pred,v-loss, step=1) & 0.1473 & 0.3530 & 0.5616 & 0.8189 &6000\\
 
 & & FlowLM(x-pred,x-loss, step=5) & \textbf{0.1596} & 0.3484 & \textbf{0.5898} & \textbf{0.9206}&6000  \\
 & & FlowLM(x-pred,x-loss, step=3) & 0.1595 & 0.3489 & 0.5878 & 0.9169 &6000\\
 & & FlowLM(x-pred,x-loss, step=1) & 0.1524 & \textbf{0.3550} & 0.5713 & 0.8411 &6000\\
\hline 
% ==================== Task 2: Paraphrase ====================
\multirow{6}{*}{\shortstack[l]{Paraphrase}} 
  &\multirow{6}{*}{Few-step}& FlowLM(x-pred,v-loss, step=5) & 0.1390 & 0.4553 & 0.7161 & 0.9752 & 10000 \\
 && FlowLM(x-pred,v-loss, step=3) & 0.1477 & 0.4593 & 0.7087 & 0.9738 & 10000 \\
 && FlowLM(x-pred,v-loss, step=1) & 0.1525 & 0.4914 & 0.6968 & 0.9426 & 10000 \\
 && FlowLM(x-pred,x-loss, step=5) & 0.1942 & 0.5352 & \textbf{0.7830} & 0.9764 & 10000 \\
 && FlowLM(x-pred,x-loss, step=3) & \textbf{0.2001} & 0.5390 & 0.7809 & \textbf{0.9766} & 10000 \\
 && FlowLM(x-pred,x-loss, step=1) & 0.1896 & \textbf{0.5404} & 0.7570 & 0.9443 & 10000 \\
\hline 

% ==================== Task 3: Text Simplification ====================
\multirow{6}{*}{\shortstack[l]{Text\\Simpli-\\fication}} 
 &\multirow{6}{*}{Few-step}& FlowLM(x-pred,v-loss, step=5) & 0.2523 & 0.4820 & 0.7268 & 0.8958 & 8000\\
  && FlowLM(x-pred,v-loss, step=3) & 0.2451 & 0.4726 & 0.7022 & 0.8704 &8000 \\
 && FlowLM(x-pred,v-loss, step=1) & 0.2230 & 0.4443 & 0.6362 & 0.7662 &8000 \\
 && FlowLM(x-pred,x-loss, step=5) & \textbf{0.2601} & \textbf{0.4868} & \textbf{0.7316} & \textbf{0.9054} &8000\\
  && FlowLM(x-pred,x-loss, step=3) & 0.2539 & 0.4827 & 0.7179 & 0.8834 &8000 \\
 && FlowLM(x-pred,x-loss, step=1) & 0.2316 & 0.4513 & 0.6420 & 0.7622 &8000 \\
 % \cmidrule(lr){2-8}
 % & \multirow{3}{*}{Few-step}& \textsc{FlowLM2}(step=5) & 0.2294 & 0.4676 & 0.6886 & 0.8779 & 11.15 \\
 %  && \textsc{FlowLM2}(step=10) & 0.2539 & 0.4827 & 0.7179 & 0.8834 & 16.29 \\
 % && \textsc{FlowLM2}(step=20) & 0.2316 & 0.4513 & 0.6420 & 0.7622 & 13.40 \\
\hline 

\bottomrule
\end{tabular}

\label{tab:compare with HKM}

\caption{Comparison between x-pred x-loss and x-pred, v-loss in FlowLM\textbf{(mbr=1)} using \textbf{logit-normal time sampling}. The best results of few-step generation model are \textbf{bold}.}
\begin{tabular}{lll | ccc | cc }
\toprule
Tasks & Type &Methods & BLEU$\uparrow$ & R-L$\uparrow$ & BERTScore$\uparrow$ & dist-1$\uparrow$ &Training epoch\\
\midrule

\multirow{6}{*}{\shortstack[l]{Question \\Generation}} 

 & \multirow{6}{*}{Few-step} & FlowLM(x-pred,v-loss, step=5) & 0.1414 & 0.3326 & 0.5712 & 0.9159 & 6000  \\
 & & FlowLM(x-pred,v-loss, step=3) & 0.1400 & 0.3326 & 0.5676 & 0.9126 &6000\\
 & & FlowLM(x-pred,v-loss, step=1) & 0.1292 & 0.3376 & 0.5428 & 0.7998 &6000\\
 & & FlowLM(x-pred,x-loss, step=5) & \textbf{0.1445} & 0.3379 & \textbf{0.5789} & \textbf{0.9180} & 6000  \\
 & & FlowLM(x-pred,x-loss, step=3) & 0.1441 & 0.3390 & 0.5761 & 0.9139 &6000\\
 & & FlowLM(x-pred,x-loss, step=1) & 0.1338 & \textbf{0.3433} & 0.5502 & 0.7998 & 6000\\
 
\hline 
% ==================== Task 2: Paraphrase ====================
\multirow{6}{*}{\shortstack[l]{Paraphrase}} 
  &\multirow{6}{*}{Few-step}& FlowLM(x-pred,v-loss, step=5) & 0.1500 & 0.4732 & 0.7323 & 0.9701 & 10000 \\
 && FlowLM(x-pred,v-loss, step=3) & 0.1523 & 0.4701 & 0.7219 & \textbf{0.9708} & 10000 \\
 && FlowLM(x-pred,v-loss, step=1) & 0.1624 & 0.5067 & 0.7177 & 0.9364 & 10000 \\
 && FlowLM(x-pred,x-loss, step=5) & 0.1523 & 0.4760 & \textbf{0.7360} & 0.9640 & 10000 \\
 && FlowLM(x-pred,x-loss, step=3) & 0.1563 & 0.4767 & 0.7338 & 0.9639 & 10000 \\
 && FlowLM(x-pred,x-loss, step=1) & \textbf{0.1726} & \textbf{0.5144} & 0.7250 & 0.9187 & 10000 \\
\hline 

% ==================== Task 3: Text Simplification ====================
\multirow{6}{*}{\shortstack[l]{Text\\Simpli-\\fication}} 

 &\multirow{6}{*}{Few-step}& FlowLM(x-pred,v-loss, step=5) & 0.2499 & 0.4806 & 0.7284 & 0.9097 &8000\\
  && FlowLM(x-pred,v-loss, step=3) & 0.2429 & 0.4741 & 0.7107 & 0.8941 &8000 \\
 && FlowLM(x-pred,v-loss, step=1) & 0.2101 & 0.4351 & 0.6296 & 0.7809 &8000 \\
 && FlowLM(x-pred,x-loss, step=5) & \textbf{0.2502} & \textbf{0.4813} & \textbf{0.7294} & 0\textbf{.9108} & 8000\\
  && FlowLM(x-pred,x-loss, step=3) & 0.2445 & 0.4767 & 0.7159 & 0.8963 &8000 \\
 && FlowLM(x-pred,x-loss, step=1) & 0.2144 & 0.4398 & 0.6321 & 0.7908 &8000 \\
 % \cmidrule(lr){2-8}
 % & \multirow{3}{*}{Few-step}& \textsc{FlowLM2}(step=5) & 0.2294 & 0.4676 & 0.6886 & 0.8779 & 11.15 \\
 %  && \textsc{FlowLM2}(step=10) & 0.2539 & 0.4827 & 0.7179 & 0.8834 & 16.29 \\
 % && \textsc{FlowLM2}(step=20) & 0.2316 & 0.4513 & 0.6420 & 0.7622 & 13.40 \\
\hline 

\bottomrule
\end{tabular}

\label{tab:compare with HKM1}

\vspace{10pt} % 在表格和文字之间加一点间距

% --- 将原本在外部的文字移入内部 ---
\begin{minipage}{\textwidth} % 使用 minipage 保证文字宽度
\small
We compare x-loss and v-loss respectively using uniform time sampling (used in FlowLM) and sigmoid $t = 1 - \text{sigmoid}(\mu + \sigma \epsilon), \epsilon \sim \mathcal{N}(0, 1)$ (used in \citep{li2025back}).  However, we introduce a minor modification as we treat $x_1$ as noise in FlowLM. Our v-loss function is modified as $\frac{\|\mathbf{z}_0 - \mathbf{z}_{0,pred}\|^2}{t^2}$, calculated by:
$\|\mathbf{v}_{\text{estimated}} - \mathbf{v}_{\text{average}}\|^2 = \left\| \frac{\mathbf{z}_t - \mathbf{z}_{0,pred}}{t} - \frac{\mathbf{z}_t - \mathbf{z}_0}{t} \right\|^2 = \frac{\|\mathbf{z}_0 - \mathbf{z}_{0,pred}\|^2}{t^2}$.

We observe that FlowLM trained with x-pred and x-loss has better performance, no matter what time sampling method is applied during training. This indicates that in FlowLM, using x-pred, x-loss is better, which is different from what \citet{li2025back} conclude in image generation field. This disparity arises from the distinct nature of the data and its associated processing pipeline. Extrapolating this to other fields, we argue that while v-loss excels in continuous domains requiring high-precision refinement, x-loss is superior for discrete-output tasks. In the latter case, the model’s inherent tolerance for minor latent deviations—owing to the final quantization step—renders the high-precision focus of v-loss unnecessary.

Additionally, we observe that model trained with logit-normal time sampling has worse performance, which indicates that the logit-normal time-step sampling is not suitable for FlowLM. 

% This is caused by difference of task and later processing method. Promote it to other fields, we think x-pred and v-loss is suitable for tasks requiring high precision in the denoising process of flow matching, while for those need later rounding and thus doesn't require high precision, x-pred and x-loss is better.
\end{minipage}
\end{table*}

% We compare between x-loss and v-loss relatively using uniform time sampling(used in FlowLM) and sigmoid $
%     t =1- \text{sigmoid}(\mu + \sigma \epsilon), \quad \epsilon \sim \mathcal{N}(0, 1)$ (same as \citep{li2025back}). But we make small change as we treat $x_1$ as noise in FlowLM. Our v-loss function is also modified by $\frac{\|\mathbf{z}_0 - \mathbf{z}_{0,pred}\|^2}{t^2}$ ,calculated by
% $\|\mathbf{v_{estimated}} - \mathbf{v_{average}}\|=\|\frac{\mathbf{z}_t - \mathbf{z}_{0,pred}}{t}-\frac{\mathbf{z}_t - \mathbf{z}_0}{t}\|^2=\frac{\|\mathbf{z}_0 - \mathbf{z}_{0,pred}\|^2}{t^2} $.
\FloatBarrier
\newpage
\section{Experimental Details}

% =======================================================
% FIGURE: Hyperparameters + MBR Analysis Curves (QG Data)
% =======================================================
\begin{figure*}[ht]
    \centering
    \captionof{table}{Experimental hyperparameter settings for Question Generation task.}
    % --- 1. 参数表 (针对 QG 任务的设置) ---
    \begin{minipage}{0.85\linewidth}
        \centering
        \scriptsize 
        \renewcommand{\arraystretch}{1.0}
        \begin{tabular}{lc|lc}
            \toprule
            \textbf{Parameter} & \textbf{Value} & \textbf{Parameter} & \textbf{Value} \\ \midrule
            \rowcolor[gray]{0.9} \multicolumn{4}{l}{\textit{Architecture \& Diffusion Configuration}} \\
            Dataset & Question Generation & \textbf{Train-num-Steps} ($T$) & 20 \\
            Vocabulary Size & 30,522 (BERT) &\\
            Max Seq Length & 128 & Predict Objective & $x_{start}$ \\
            Hidden Dim & 128 & EMA & 0.9999 \\ \midrule
            \rowcolor[gray]{0.9} \multicolumn{4}{l}{\textit{Training Hyperparameters}} \\
            Batch Size & 2048 & Learning Rate & $1 \times 10^{-4}$ \\
            Training epochs & 6,000 & Warmup Steps & 500 \\
            Precision & BF16 & Dropout / Weight Decay & 0.1 / 0.0 \\
            \bottomrule
        \end{tabular}
        \label{tab:qg_hyperparams}
    \end{minipage}

    % --- 2. 绘图配置 ---
    \pgfplotsset{
        every axis/.style={
            width=\linewidth,
            height=5cm,
            grid=major,
            grid style={dashed, gray!30},
            xlabel={MBR (n)},
            xmin=1, xmax=10,
            xtick={1,2,3,4,5,6,7,8,9,10},
            tick label style={font=\tiny},
            scaled y ticks=false,
            yticklabel style={/pgf/number format/.cd, fixed, fixed zerofill, precision=3},
            label style={font=\scriptsize},
        }
    }

    % 图例
    \begin{tikzpicture}
        \begin{customlegend}[
            legend columns=-1,
            legend style={draw=none, column sep=0.2cm, font=\tiny},
            legend entries={DiffuSeq(2000), DiffuSeq(DPM10), FlowLM(step=5), FlowLM(step=3), FlowLM(step=1)}
        ]
        \addlegendimage{color=blue!70!black, mark=*, thick}
        \addlegendimage{color=black, mark=triangle*, thick, dashed}
        \addlegendimage{color=orange!90!black, mark=square*, thick}
        \addlegendimage{color=green!70!black, mark=diamond*, thick}
        \addlegendimage{color=red!80!black, mark=x, thick}
        \end{customlegend}
    \end{tikzpicture}

    % 四个子图 (使用图片后半段 QG 真实数据)
    \begin{subfigure}{0.48\linewidth}
        \centering
        \begin{tikzpicture}
            \begin{axis}[ylabel={BLEU}]
            \addplot[color=blue!70!black, mark=*, thick] coordinates {(1,0.1514)(2,0.1532)(3,0.1585)(4,0.1610)(5,0.1622)(6,0.1634)(7,0.1649)(8,0.1653)(9,0.1654)(10,0.1654)};
            \addplot[color=black, mark=triangle*, thick, dashed] coordinates {(1,0.1434)(2,0.1441)(3,0.1461)(4,0.1471)(5,0.1480)(6,0.1485)(7,0.1490)(8,0.1491)(9,0.1491)(10,0.1487)};
            \addplot[color=orange!90!black, mark=square*, thick] coordinates {(1,0.1596)(2,0.1608)(3,0.1642)(4,0.1656)(5,0.1669)(6,0.1673)(7,0.1677)(8,0.1678)(9,0.1682)(10,0.1687)};
            \addplot[color=green!70!black, mark=diamond*, thick] coordinates {(1,0.1600)(2,0.1600)(3,0.1631)(4,0.1639)(5,0.1649)(6,0.1654)(7,0.1662)(8,0.1664)(9,0.1670)(10,0.1671)};
            \addplot[color=red!80!black, mark=x, thick] coordinates {(1,0.1527)(2,0.1526)(3,0.1519)(4,0.1535)(5,0.1541)(6,0.1540)(7,0.1540)(8,0.1543)(9,0.1543)(10,0.1540)};
            \end{axis}
        \end{tikzpicture}
        \caption{BLEU}
    \end{subfigure}
    \hfill
    \begin{subfigure}{0.48\linewidth}
        \centering
        \begin{tikzpicture}
            \begin{axis}[ylabel={ROUGE-L}]
            \addplot[color=blue!70!black, mark=*, thick] coordinates {(1,0.3477)(2,0.3488)(3,0.3575)(4,0.3611)(5,0.3621)(6, 0.3628) (7, 0.3638) (8, 0.3644) (9, 0.3648) (10, 0.3659)};
            \addplot[color=black, mark=triangle*, thick, dashed] coordinates {(1,0.3513)(2,0.3512)(3,0.3552)(4,0.3554)(5,0.3571)(6,0.3576)(7,0.3582)(8,0.3584)(9,0.3588)(10,0.3586)};
            \addplot[color=orange!90!black, mark=square*, thick] coordinates {(1,0.3484)(2,0.3500)(3,0.3549)(4,0.3578)(5,0.3592)(6,0.3602)(7,0.3608)(8,0.3612)(9,0.3623)(10,0.3629)};
            \addplot[color=green!70!black, mark=diamond*, thick] coordinates {(1,0.3499)(2,0.3508)(3,0.3549)(4,0.3571)(5,0.3577)(6,0.3589)(7,0.3600)(8,0.3610)(9,0.3617)(10,0.3620)};
            \addplot[color=red!80!black, mark=x, thick] coordinates {(1,0.3549)(2,0.3551)(3,0.3551)(4,0.3566)(5,0.3575)(6,0.3567)(7,0.3575)(8,0.3577)(9,0.3578)(10,0.3575)};
            \end{axis}
        \end{tikzpicture}
        \caption{ROUGE-L}
    \end{subfigure}

    \begin{subfigure}{0.48\linewidth}
        \centering
        \begin{tikzpicture}
            \begin{axis}[ylabel={BERTScore}]
            \addplot[color=blue!70!black, mark=*, thick] coordinates {(1,0.5874)(2,0.5880)(3,0.5936)(4,0.5973)(5,0.5989)(6,0.6002)(7,0.6011)(8,0.6019)(9,0.6029)(10,0.6029)};
            \addplot[color=black, mark=triangle*, thick, dashed] coordinates {(1,0.5740)(2,0.5746)(3,0.5761)(4,0.5767)(5,0.5773)(6,0.5782)(7,0.5790)(8,0.5790)(9,0.5788)(10,0.5789)};
            \addplot[color=orange!90!black, mark=square*, thick] coordinates {(1,0.5898)(2,0.5911)(3,0.5951)(4,0.5966)(5,0.5982)(6,0.5994)(7,0.6003)(8,0.6007)(9,0.6015)(10,0.6022)};
            \addplot[color=green!70!black, mark=diamond*, thick] coordinates {(1,0.5880)(2,0.5886)(3,0.5916)(4,0.5930)(5,0.5943)(6,0.5952)(7,0.5964)(8,0.5966)(9,0.5971)(10,0.5981)};
            \addplot[color=red!80!black, mark=x, thick] coordinates {(1,0.5711)(2,0.5707)(3,0.5707)(4,0.5720)(5,0.5727)(6,0.5723)(7,0.5725)(8,0.5727)(9,0.5730)(10,0.5727)};
            \end{axis}
        \end{tikzpicture}
        \caption{BERTScore}
    \end{subfigure}
    \hfill
    \begin{subfigure}{0.48\linewidth}
        \centering
        \begin{tikzpicture}
            \begin{axis}[ylabel={Dist-1}]
            \addplot[color=blue!70!black, mark=*, thick] coordinates {(1,0.9126)(2,0.9044)(3,0.9131)(4,0.9124)(5,0.9116)(6,0.9102)(7,0.9085)(8,0.9071)(9,0.9068)(10,0.9063)};
            \addplot[color=black, mark=triangle*, thick, dashed] coordinates {(1,0.8636)(2,0.8608)(3,0.8617)(4,0.8610)(5,0.8619)(6,0.8604)(7,0.8608)(8,0.8609)(9,0.8605)(10,0.8602)};
            \addplot[color=orange!90!black, mark=square*, thick] coordinates {(1,0.9206)(2,0.9172)(3,0.9196)(4,0.9186)(5,0.9173)(6,0.9171)(7,0.9159)(8,0.9150)(9,0.9149)(10,0.9147)};
            \addplot[color=green!70!black, mark=diamond*, thick] coordinates {(1,0.9166)(2,0.9153)(3,0.9158)(4,0.9146)(5,0.9138)(6,0.9124)(7,0.9122)(8,0.9114)(9,0.9111)(10,0.9109)};
            \addplot[color=red!80!black, mark=x, thick] coordinates {(1,0.8430)(2,0.8455)(3,0.8419)(4,0.8405)(5,0.8405)(6,0.8392)(7,0.8384)(8,0.8385)(9,0.8385)(10,0.8387)};
            \end{axis}
        \end{tikzpicture}
        \caption{Dist-1}
    \end{subfigure}
    \captionof{figure}{Question Generation results analysis across MBR candidate sizes (1--10).}
\end{figure*}
\FloatBarrier
% =======================================================
% TABLE: Detailed Results for Question Generation (All 10 MBR)
% =======================================================
\begin{table*}[p] 
\centering
\caption{Detailed performance metrics for Question Generation across all 10 MBR candidate levels.}
\footnotesize
\begin{tabular}{cclcccc}
\toprule
MBR ($n$) & Category & Model & BLEU & ROUGE-L & BERTScore & Dist-1 \\ 
\midrule
% --- n=1 ---
\multirow{5}{*}{1} 
 & Multi-step & Diffuseq(2000) & 0.1514 & 0.3477 & 0.5874 & 0.9126 \\
 \cmidrule(lr){2-7}
 & \multirow{4}{*}{Few-step} & Diffuseq(DPM, 10) & 0.1434 & 0.3513 & 0.5740 & 0.8636 \\
 & & FlowLM(step=5) & 0.1596 & 0.3484 & \textbf{0.5898} & \textbf{0.9206} \\
 & & FlowLM(step=3) & \textbf{0.1600} & 0.3499 & 0.5880 & 0.9166 \\
 & & FlowLM(step=1) & 0.1527 & \textbf{0.3549} & 0.5711 & 0.8430 \\
\midrule
% --- n=2 ---
\multirow{5}{*}{2} 
 & Multi-step & Diffuseq(2000) & 0.1532 & 0.3488 & 0.5880 & 0.9044 \\
 \cmidrule(lr){2-7}
 & \multirow{4}{*}{Few-step} & Diffuseq(DPM, 10) & 0.1441 & 0.3512 & 0.5746 & 0.8608 \\
 & & FlowLM(step=5) & \textbf{0.1608} & 0.3500 & \textbf{0.5911} & \textbf{0.9172} \\
 & & FlowLM(step=3) & 0.1600 & 0.3508 & 0.5886 & 0.9153 \\
 & & FlowLM(step=1) & 0.1526 & \textbf{0.3551} & 0.5707 & 0.8455 \\
\midrule
% --- n=3 ---
\multirow{5}{*}{3} 
 & Multi-step & Diffuseq(2000) & 0.1585 & 0.3575 & 0.5936 & 0.9131 \\
 \cmidrule(lr){2-7}
 & \multirow{4}{*}{Few-step} & Diffuseq(DPM, 10) & 0.1461 & \textbf{0.3552} & 0.5761 & 0.8617 \\
 & & FlowLM(step=5) & \textbf{0.1642} & 0.3549 & \textbf{0.5951} & \textbf{0.9196} \\
 & & FlowLM(step=3) & 0.1631 & 0.3549 & 0.5916 & 0.9158 \\
 & & FlowLM(step=1) & 0.1519 & 0.3551 & 0.5707 & 0.8419 \\
\midrule
% --- n=4 ---
\multirow{5}{*}{4} 
 & Multi-step & Diffuseq(2000) & 0.1610 & 0.3611 & 0.5973 & 0.9124 \\
 \cmidrule(lr){2-7}
 & \multirow{4}{*}{Few-step} & Diffuseq(DPM, 10) & 0.1471 & 0.3554 & 0.5767 & 0.8610 \\
 & & FlowLM(step=5) & \textbf{0.1656} & \textbf{0.3578} & \textbf{0.5966} & \textbf{0.9186} \\
 & & FlowLM(step=3) & 0.1639 & 0.3571 & 0.5930 & 0.9146 \\
 & & FlowLM(step=1) & 0.1535 & 0.3566 & 0.5720 & 0.8405 \\
\midrule
% --- n=5 ---
\multirow{5}{*}{5} 
 & Multi-step & Diffuseq(2000) & 0.1622 & 0.3621 &0.5989 & 0.9116 \\
 \cmidrule(lr){2-7}
 & \multirow{4}{*}{Few-step} & Diffuseq(DPM, 10) & 0.1480 & 0.3571 & 0.5773 & 0.8619 \\
 & & FlowLM(step=5) & \textbf{0.1669} & \textbf{0.3592} & \textbf{0.5982} & \textbf{0.9173} \\
 & & FlowLM(step=3) & 0.1649 & 0.3577 & 0.5943 & 0.9138 \\
 & & FlowLM(step=1) & 0.1541 & 0.3575 & 0.5727 & 0.8405 \\
\midrule
% --- n=6 ---
\multirow{5}{*}{6} 
 & Multi-step & Diffuseq(2000) & 0.1634 & 0.3628 & 0.6002 & 0.9102 \\
 \cmidrule(lr){2-7}
 & \multirow{4}{*}{Few-step} & Diffuseq(DPM, 10) & 0.1485 & 0.3576 & 0.5782 & 0.8604 \\
 & & FlowLM(step=5) & \textbf{0.1673} & \textbf{0.3602} & \textbf{0.5994} & \textbf{0.9171} \\
 & & FlowLM(step=3) & 0.1654 & 0.3589 & 0.5952 & 0.9124 \\
 & & FlowLM(step=1) & 0.1540 & 0.3567 & 0.5723 & 0.8392 \\
\midrule
% --- n=7 ---
\multirow{5}{*}{7} 
 & Multi-step & Diffuseq(2000) & 0.1649 & 0.3638 & 0.6011 & 0.9085 \\
 \cmidrule(lr){2-7}
 & \multirow{4}{*}{Few-step} & Diffuseq(DPM, 10) & 0.1490 & 0.3582 & 0.5790 & 0.8608 \\
 & & FlowLM(step=5) & \textbf{0.1677} & \textbf{0.3608} & \textbf{0.6003} & \textbf{0.9159} \\
 & & FlowLM(step=3) & 0.1662 & 0.3600 & 0.5964 & 0.9122 \\
 & & FlowLM(step=1) & 0.1540 & 0.3575 & 0.5725 & 0.8384 \\
\midrule
% --- n=8 ---
\multirow{5}{*}{8} 
 & Multi-step & Diffuseq(2000) & 0.1653 & 0.3644 & 0.6019 & 0.9071 \\
 \cmidrule(lr){2-7}
 & \multirow{4}{*}{Few-step} & Diffuseq(DPM, 10) & 0.1491 & 0.3584 & 0.5790 & 0.8609 \\
 & & FlowLM(step=5) & \textbf{0.1678} & \textbf{0.3612} & \textbf{0.6007} & \textbf{0.9150} \\
 & & FlowLM(step=3) & 0.1664 & 0.3610 & 0.5966 & 0.9114 \\
 & & FlowLM(step=1) & 0.1543 & 0.3577 & 0.5727 & 0.8385 \\
\midrule
% --- n=9 ---
\multirow{5}{*}{9} 
 & Multi-step & Diffuseq(2000) & 0.1654 & 0.3648 & 0.6029 & 0.9068 \\
 \cmidrule(lr){2-7}
 & \multirow{4}{*}{Few-step} & Diffuseq(DPM, 10) & 0.1491 & 0.3588 & 0.5788 & 0.8605 \\
 & & FlowLM(step=5) & \textbf{0.1682} & \textbf{0.3623} & \textbf{0.6015} & \textbf{0.9149} \\
 & & FlowLM(step=3) & 0.1670 & 0.3617 & 0.5971 & 0.9111 \\
 & & FlowLM(step=1) & 0.1543 & 0.3578 & 0.5730 & 0.8385 \\
\midrule
% --- n=10 ---
\multirow{5}{*}{10} 
 & Multi-step & Diffuseq(2000) & 0.1654 & 0.3659 & 0.6029 & 0.9063 \\
 \cmidrule(lr){2-7}
 & \multirow{4}{*}{Few-step} & Diffuseq(DPM, 10) & 0.1487 & 0.3586 & 0.5789 & 0.8602 \\
 & & FlowLM(step=5) & \textbf{0.1687} & \textbf{0.3629} & \textbf{0.6022} & \textbf{0.9147} \\
 & & FlowLM(step=3) & 0.1671 & 0.3620 & 0.5981 & 0.9109 \\
 & & FlowLM(step=1) & 0.1540 & 0.3575 & 0.5727 & 0.8387 \\
\bottomrule
\end{tabular}
\end{table*}

% 使用 figure* 环境包裹“表+图”，确保它们作为一个整体跨两栏显示且不被拆分
\begin{figure*}[ht]
    \centering
    \captionof{table}{Experimental hyperparameter settings for Paraphrase task.}
    % --- 1. 参数表 (放上面) ---
    \begin{minipage}{0.8\linewidth} % 限制表格宽度，让它居中更美观
        \centering
        \scriptsize % 使用 scriptsize 节省空间
        \renewcommand{\arraystretch}{1} % 稍微压缩行高
        \begin{tabular}{lc|lc}
            \toprule
            \textbf{Parameter} & \textbf{Value} & \textbf{Parameter} & \textbf{Value} \\ \midrule
            \rowcolor[gray]{0.9} \multicolumn{4}{l}{\textit{Architecture \& Diffusion}} \\
            Dataset & Paraphrase & \textbf{Train-num-Steps} ($T$) & 20 \\
            Vocabulary Size & 30,522 (BERT) & \\
            Max Seq Length & 128 & Predict Objective & $x_{start}$ \\
            Hidden Dim & 128 &  EMA & 0.9999 \\ \midrule
            \rowcolor[gray]{0.9} \multicolumn{4}{l}{\textit{Training Hyperparameters}} \\
            Batch Size & 2048 & Learning Rate & $1 \times 10^{-4}$ \\
            Training epochs & 10,000 & Warmup Steps & 500 \\
            Precision & BF16 & Dropout / Decay & 0.1 / 0.0 \\
            \bottomrule
        \end{tabular}
        \label{tab:hyperparams}
    \end{minipage}

    \vspace{0.6cm} % 表和图之间的间距

    % --- 2. 绘图部分 (放下面) ---
    \pgfplotsset{
        every axis/.style={
            width=\linewidth,
            height=5cm, % 进一步压缩绘图高度 (从 5cm 降到 4.2cm)
            grid=major,
            grid style={dashed, gray!30},
            xlabel={MBR (n)},
            xmin=1, xmax=10,
            xtick={1,2,3,4,5,6,7,8,9,10},
            tick label style={font=\tiny}, % 坐标轴数字调到最小
            scaled y ticks=false,
            yticklabel style={/pgf/number format/.cd, fixed, fixed zerofill, precision=3},
            label style={font=\scriptsize},
        }
    }

    % 图例 (Legend) - 紧凑排列
    \begin{tikzpicture}
        \begin{customlegend}[
            legend columns=-1,
            legend style={draw=none, column sep=0.2cm, font=\tiny},
            legend entries={DiffuSeq(2000), DiffuSeq(DPM10), FlowLM(step=5), FlowLM(step=3), FlowLM(step=1)}
        ]
        \addlegendimage{color=blue!70!black, mark=*, thick}
        \addlegendimage{color=black, mark=triangle*, thick, dashed}
        \addlegendimage{color=orange!90!black, mark=square*, thick}
        \addlegendimage{color=green!70!black, mark=diamond*, thick}
        \addlegendimage{color=red!80!black, mark=x, thick}
        \end{customlegend}
    \end{tikzpicture}
    
    \vspace{0.1cm}

    % 2x2 子图布局
    \begin{subfigure}{0.48\linewidth}
        \centering
        \begin{tikzpicture}
            \begin{axis}[ylabel={BLEU}]
            \addplot[color=blue!70!black, mark=*, thick] coordinates {(1,0.1868)(2,0.1856)(3,0.2087)(4,0.2168)(5,0.2229)(6,0.2269)(7,0.2296)(8,0.2330)(9,0.2348)(10,0.2377)};
            \addplot[color=black, mark=triangle*, thick, dashed] coordinates {(1,0.1952)(2,0.2016)(3,0.2091)(4,0.2130)(5,0.2145)(6,0.2164)(7,0.2179)(8,0.2186)(9,0.2191)(10,0.2204)};
            \addplot[color=orange!90!black, mark=square*, thick] coordinates {(1,0.1916)(2,0.1946)(3,0.2114)(4,0.2170)(5,0.2204)(6,0.2256)(7,0.2285)(8,0.2306)(9,0.2307)(10,0.2319)};
            \addplot[color=green!70!black, mark=diamond*, thick] coordinates {(1,0.1987)(2,0.1992)(3,0.2114)(4,0.2162)(5,0.2188)(6,0.2222)(7,0.2271)(8,0.2279)(9,0.2262)(10,0.2278)};
            \addplot[color=red!80!black, mark=x, thick] coordinates {(1,0.1910)(2,0.1890)(3,0.1914)(4,0.1915)(5,0.1908)(6,0.1907)(7,0.1906)(8,0.1906)(9,0.1914)(10,0.1919)};
            \end{axis}
        \end{tikzpicture}
        \caption{BLEU}
    \end{subfigure}
    \hfill
    \begin{subfigure}{0.48\linewidth}
        \centering
        \begin{tikzpicture}
            \begin{axis}[ylabel={ROUGE-L}]
            \addplot[color=blue!70!black, mark=*, thick] coordinates {(1,0.5316)(2,0.5268)(3,0.5561)(4,0.5661)(5,0.5721)(6,0.5772)(7,0.5791)(8,0.5829)(9,0.5843)(10,0.5870)};
            \addplot[color=black, mark=triangle*, thick, dashed] coordinates {(1,0.5583)(2,0.5559)(3,0.5632)(4,0.5697)(5,0.5713)(6,0.5713)(7,0.5736)(8,0.5744)(9,0.5749)(10,0.5761)};
            \addplot[color=orange!90!black, mark=square*, thick] coordinates {(1,0.5289)(2,0.5318)(3,0.5515)(4,0.5579)(5,0.5633)(6,0.5685)(7,0.5717)(8,0.5748)(9,0.5763)(10,0.5784)};
            \addplot[color=green!70!black, mark=diamond*, thick] coordinates {(1,0.5357)(2,0.5381)(3,0.5523)(4,0.5575)(5,0.5612)(6,0.5642)(7,0.5690)(8,0.5707)(9,0.5707)(10,0.5715)};
            \addplot[color=red!80!black, mark=x, thick] coordinates {(1,0.5394)(2,0.5385)(3,0.5407)(4,0.5414)(5,0.5420)(6,0.5412)(7,0.5410)(8,0.5410)(9,0.5423)(10,0.5432)};
            \end{axis}
        \end{tikzpicture}
        \caption{ROUGE-L}
    \end{subfigure}

    \vspace{0.2cm} % 行间距

    \begin{subfigure}{0.48\linewidth}
        \centering
        \begin{tikzpicture}
            \begin{axis}[ylabel={BERTScore}]
            \addplot[color=blue!70!black, mark=*, thick] coordinates {(1,0.7920)(2,0.7877)(3,0.8065)(4,0.8173)(5,0.8217)(6,0.8262)(7,0.8283)(8,0.8304)(9,0.8321)(10,0.8333)};
            \addplot[color=black, mark=triangle*, thick, dashed] coordinates {(1,0.7932)(2,0.7916)(3,0.7982)(4,0.8041)(5,0.8055)(6,0.8061)(7,0.8077)(8,0.8085)(9,0.8094)(10,0.8105)};
            \addplot[color=orange!90!black, mark=square*, thick] coordinates {(1,0.7827)(2,0.7826)(3,0.7972)(4,0.8042)(5,0.8079)(6,0.8117)(7,0.8146)(8,0.8170)(9,0.8177)(10,0.8188)};
            \addplot[color=green!70!black, mark=diamond*, thick] coordinates {(1,0.7784)(2,0.7797)(3,0.7909)(4,0.7964)(5,0.7995)(6,0.8035)(7,0.8070)(8,0.8090)(9,0.8102)(10,0.8103)};
            \addplot[color=red!80!black, mark=x, thick] coordinates {(1,0.7560)(2,0.7554)(3,0.7561)(4,0.7575)(5,0.7585)(6,0.7587)(7,0.7585)(8,0.7585)(9,0.7597)(10,0.7601)};
            \end{axis}
        \end{tikzpicture}
        \caption{BERTScore}
    \end{subfigure}
    \hfill
    \begin{subfigure}{0.48\linewidth}
        \centering
        \begin{tikzpicture}
            \begin{axis}[ylabel={Dist-1}]
            \addplot[color=blue!70!black, mark=*, thick] coordinates {(1,0.9737)(2,0.9700)(3,0.9755)(4,0.9783)(5,0.9787)(6,0.9797)(7,0.9797)(8,0.9812)(9,0.9817)(10,0.9813)};
            \addplot[color=black, mark=triangle*, thick, dashed] coordinates {(1,0.9566)(2,0.9606)(3,0.9615)(4,0.9628)(5,0.9635)(6,0.9635)(7,0.9640)(8,0.9644)(9,0.9654)(10,0.9661)};
            \addplot[color=orange!90!black, mark=square*, thick] coordinates {(1,0.9785)(2,0.9775)(3,0.9787)(4,0.9794)(5,0.9795)(6,0.9793)(7,0.9805)(8,0.9809)(9,0.9805)(10,0.9805)};
            \addplot[color=green!70!black, mark=diamond*, thick] coordinates {(1,0.9757)(2,0.9764)(3,0.9772)(4,0.9779)(5,0.9779)(6,0.9782)(7,0.9783)(8,0.9783)(9,0.9778)(10,0.9784)};
            \addplot[color=red!80!black, mark=x, thick] coordinates {(1,0.9446)(2,0.9481)(3,0.9452)(4,0.9460)(5,0.9462)(6,0.9460)(7,0.9459)(8,0.9459)(9,0.9461)(10,0.9463)};
            \end{axis}
        \end{tikzpicture}
        \caption{Dist-1}
    \end{subfigure}

    \captionof{figure}{Paraphrase results analysis across MBR candidate sizes (1--10).}
    \label{fig:combined_results}
\end{figure*}

\begin{table*}[t]
\centering
\caption{Detailed performance metrics for Paraphrase across all 10 MBR candidate levels.(Best in Fast is \textbf{bold}).}
\footnotesize % 建议使用较小字号以适应 50 行数据
\begin{tabular}{cclcccc}
\toprule
MBR ($n$) & Category & Model & BLEU & ROUGE-L & BERTScore & Dist-1 \\ 
\midrule
% --- n=1 ---
\multirow{5}{*}{1} 
 & Multi-step & Diffuseq(2000) & 0.1868 & 0.5316 & 0.7920 & 0.9737 \\
 \cmidrule(lr){2-7}
 & \multirow{4}{*}{Few-step} & Diffuseq(DPM, 10) & 0.1952 & \textbf{0.5583} & \textbf{0.7932} & 0.9566 \\
 & & FlowLM(step=5) & 0.1916 & 0.5289 & 0.7827 & \textbf{0.9785} \\
 & & FlowLM(step=3) & \textbf{0.1987} & 0.5357 & 0.7784 & 0.9757 \\
 & & FlowLM(step=1) & 0.1910 & 0.5394 & 0.7560 & 0.9446 \\
\midrule
% --- n=2 ---
\multirow{5}{*}{2} 
 & Multi-step & Diffuseq(2000) & 0.1856 & 0.5268 & 0.7877 & 0.9700 \\
 \cmidrule(lr){2-7}
 & \multirow{4}{*}{Few-step} & Diffuseq(DPM, 10) & \textbf{0.2016} & \textbf{0.5559} & \textbf{0.7916} & 0.9606 \\
 & & FlowLM(step=5) & 0.1946 & 0.5318 & 0.7826 & \textbf{0.9775} \\
 & & FlowLM(step=3) & 0.1992 & 0.5381 & 0.7797 & 0.9764 \\
 & & FlowLM(step=1) & 0.1890 & 0.5385 & 0.7554 & 0.9481 \\
\midrule
% --- n=3 ---
\multirow{5}{*}{3} 
 & Multi-step & Diffuseq(2000) & 0.2087 & 0.5561 & 0.8065 & 0.9755 \\
 \cmidrule(lr){2-7}
 & \multirow{4}{*}{Few-step} & Diffuseq(DPM, 10) & 0.2091 & \textbf{0.5632} & \textbf{0.7982} & 0.9615 \\
 & & FlowLM(step=5) & 0.2114 & 0.5515 & 0.7972 & \textbf{0.9787} \\
 & & FlowLM(step=3) & \textbf{0.2114} & 0.5523 & 0.7909 & 0.9772 \\
 & & FlowLM(step=1) & 0.1914 & 0.5407 & 0.7561 & 0.9452 \\
\midrule
% --- n=4 ---
\multirow{5}{*}{4} 
 & Multi-step & Diffuseq(2000) & 0.2168 & 0.5661 & 0.8173 & 0.9783 \\
 \cmidrule(lr){2-7}
 & \multirow{4}{*}{Few-step} & Diffuseq(DPM, 10) & 0.2130 & \textbf{0.5697} & 0.8041 & 0.9628 \\
 & & FlowLM(step=5) & \textbf{0.2170} & 0.5579 & \textbf{0.8042} & \textbf{0.9794} \\
 & & FlowLM(step=3) & 0.2162 & 0.5575 & 0.7964 & 0.9779 \\
 & & FlowLM(step=1) & 0.1915 & 0.5414 & 0.7575 & 0.9460 \\
\midrule
% --- n=5 ---
\multirow{5}{*}{5} 
 & Multi-step & Diffuseq(2000) & 0.2229 & 0.5721 & 0.8217 & 0.9787 \\
 \cmidrule(lr){2-7}
 & \multirow{4}{*}{Few-step} & Diffuseq(DPM, 10) & 0.2145 & \textbf{0.5713} & 0.8055 & 0.9635 \\
 & & FlowLM(step=5) & \textbf{0.2204} & 0.5633 & \textbf{0.8079} & \textbf{0.9795} \\
 & & FlowLM(step=3) & 0.2188 & 0.5612 & 0.7995 & 0.9779 \\
 & & FlowLM(step=1) & 0.1908 & 0.5420 & 0.7585 & 0.9462 \\
\midrule
% --- n=6 ---
\multirow{5}{*}{6} 
 & Multi-step & Diffuseq(2000) & 0.2269 & 0.5772 & 0.8262 & 0.9797 \\
 \cmidrule(lr){2-7}
 & \multirow{4}{*}{Few-step} & Diffuseq(DPM, 10) & 0.2164 & \textbf{0.5713} & 0.8061 & 0.9635 \\
 & & FlowLM(step=5) & \textbf{0.2256} & 0.5685 & \textbf{0.8117} & \textbf{0.9793} \\
 & & FlowLM(step=3) & 0.2222 & 0.5642 & 0.8035 & 0.9782 \\
 & & FlowLM(step=1) & 0.1907 & 0.5412 & 0.7587 & 0.9460 \\
\midrule
% --- n=7 ---
\multirow{5}{*}{7} 
 & Multi-step & Diffuseq(2000) & 0.2296 & 0.5791 & 0.8283 & 0.9797 \\
 \cmidrule(lr){2-7}
 & \multirow{4}{*}{Few-step} & Diffuseq(DPM, 10) & 0.2179 & \textbf{0.5736} & 0.8077 & 0.9640 \\
 & & FlowLM(step=5) & \textbf{0.2285} & 0.5717 & \textbf{0.8146} & \textbf{0.9805} \\
 & & FlowLM(step=3) & 0.2270 & 0.5690 & 0.8070 & 0.9783 \\
 & & FlowLM(step=1) & 0.1906 & 0.5410 & 0.7585 & 0.9459 \\
\midrule
% --- n=8 ---
\multirow{5}{*}{8} 
 & Multi-step & Diffuseq(2000) & 0.2330 & 0.5829 & 0.8304 & 0.9812 \\
 \cmidrule(lr){2-7}
 & \multirow{4}{*}{Few-step} & Diffuseq(DPM, 10) & 0.2186 & 0.5744 & 0.8085 & 0.9644 \\
 & & FlowLM(step=5) & \textbf{0.2306} & \textbf{0.5748} & \textbf{0.8170} & \textbf{0.9809} \\
 & & FlowLM(step=3) & 0.2279 & 0.5707 & 0.8090 & 0.9783 \\
 & & FlowLM(step=1) & 0.1906 & 0.5410 & 0.7585 & 0.9459 \\
\midrule
% --- n=9 ---
\multirow{5}{*}{9} 
 & Multi-step & Diffuseq(2000) & 0.2348 & 0.5843 & 0.8321 & 0.9817 \\
 \cmidrule(lr){2-7}
 & \multirow{4}{*}{Few-step} & Diffuseq(DPM, 10) & 0.2191 & 0.5749 & 0.8094 & 0.9654 \\
 & & FlowLM(step=5) & \textbf{0.2307} & \textbf{0.5763} & \textbf{0.8177} & 0.9805 \\
 & & FlowLM(step=3) & 0.2262 & 0.5707 & 0.8102 & 0.9778 \\
 & & FlowLM(step=1) & 0.1914 & 0.5423 & 0.7597 & 0.9461 \\
\midrule
% --- n=10 ---
\multirow{5}{*}{10} 
 & Multi-step & Diffuseq(2000) & 0.2377 &0.5870 & 0.8333 & 0.9813 \\
 \cmidrule(lr){2-7}
 & \multirow{4}{*}{Few-step} & Diffuseq(DPM, 10) & 0.2204 & 0.5761 & 0.8105 & 0.9661 \\
 & & FlowLM(step=5) & \textbf{0.2319} & \textbf{0.5784} & \textbf{0.8188} & \textbf{0.9805} \\
 & & FlowLM(step=3) & 0.2278 & 0.5715 & 0.8103 & 0.9784 \\
 & & FlowLM(step=1) & 0.1919 & 0.5432 & 0.7601 & 0.9463 \\
\bottomrule
\end{tabular}
\end{table*}
\FloatBarrier

% \begin{figure*}[ht]
%     \centering
    
%     % --- 1. 参数表 (放上面) ---
%     \begin{minipage}{0.8\linewidth} % 限制表格宽度，让它居中更美观
%         \centering
%         \scriptsize % 使用 scriptsize 节省空间
%         \renewcommand{\arraystretch}{1} % 稍微压缩行高
%         \begin{tabular}{lc|lc}
%             \toprule
%             \textbf{Parameter} & \textbf{Value} & \textbf{Parameter} & \textbf{Value} \\ \midrule
%             \rowcolor[gray]{0.9} \multicolumn{4}{l}{\textit{Architecture \& Diffusion}} \\
%             Dataset & Paraphrase & \textbf{Train-num-Steps} ($T$) & 20 \\
%             Vocabulary Size & 30,522 (BERT) & \\
%             Max Seq Length & 128 & Predict Objective & $x_{start}$ \\
%             Hidden Dim & 128 &  EMA & 0.9999 \\ \midrule
%             \rowcolor[gray]{0.9} \multicolumn{4}{l}{\textit{Training Hyperparameters}} \\
%             Batch Size & 2048 & Learning Rate & $1 \times 10^{-5}$ \\
%             Training Steps & 85,00 & Warmup Steps & 500 \\
%             Precision & BF16 & Dropout / Decay & 0.1 / 0.0 \\
%             \bottomrule
%         \end{tabular}
%         \captionof{table}{Hyperparameter settings for FlowLM1 experiments shown in table \ref{tab:main_results}}
%         \label{tab:hyperparams}
%     \end{minipage}

% \end{figure*}

% 使用 figure* 环境包裹“表+图”，确保它们作为一个整体跨两栏显示且不被拆分
\begin{figure*}[ht]
    \centering
    \captionof{table}{Hyperparameter settings for DiffuSeq and FlowLM experiments on Text Simplification.}
    % --- 1. 参数表 (放上面) ---
    \begin{minipage}{0.8\linewidth} % 限制表格宽度，让它居中更美观
        \centering
        \scriptsize % 使用 scriptsize 节省空间
        \renewcommand{\arraystretch}{1} % 稍微压缩行高
        \begin{tabular}{lc|lc}
            \toprule
            \textbf{Parameter} & \textbf{Value} & \textbf{Parameter} & \textbf{Value} \\ \midrule
            \rowcolor[gray]{0.9} \multicolumn{4}{l}{\textit{Architecture \& Diffusion}} \\
            Dataset & Paraphrase & \textbf{Train-num-Steps} ($T$) & 20 \\
            Vocabulary Size & 30,522 (BERT) & \\
            Max Seq Length & 128 & Predict Objective & $x_{start}$ \\
            Hidden Dim & 128 &  EMA & 0.9999 \\ \midrule
            \rowcolor[gray]{0.9} \multicolumn{4}{l}{\textit{Training Hyperparameters}} \\
            Batch Size & 2048 & Learning Rate & $1 \times 10^{-5}$ \\
            Training epochs & 8000 & Warmup Steps & 500 \\
            Precision & BF16 & Dropout / Decay & 0.1 / 0.0 \\
            \bottomrule
        \end{tabular}
        
        \label{tab:hyperparams}
    \end{minipage}

    \vspace{0.6cm} % 表和图之间的间距

    % --- 2. 绘图部分 (放下面) ---
    \pgfplotsset{
        every axis/.style={
            width=\linewidth,
            height=5cm, % 进一步压缩绘图高度
            grid=major,
            grid style={dashed, gray!30},
            xlabel={MBR (n)},
            xmin=1, xmax=10,
            xtick={1,2,3,4,5,6,7,8,9,10},
            tick label style={font=\tiny}, % 坐标轴数字调到最小
            scaled y ticks=false,
            yticklabel style={/pgf/number format/.cd, fixed, fixed zerofill, precision=2},
            label style={font=\scriptsize},
        }
    }

    % 图例 (Legend) - 紧凑排列
    \begin{tikzpicture}
        \begin{customlegend}[
            legend columns=-1,
            legend style={draw=none, column sep=0.2cm, font=\tiny},
            legend entries={DiffuSeq(2000), DiffuSeq(DPM10), FlowLM(step=5), FlowLM(step=3), FlowLM(step=1)}
        ]
        \addlegendimage{color=blue!70!black, mark=*, thick}
        \addlegendimage{color=black, mark=triangle*, thick, dashed}
        \addlegendimage{color=orange!90!black, mark=square*, thick}
        \addlegendimage{color=green!70!black, mark=diamond*, thick}
        \addlegendimage{color=red!80!black, mark=x, thick}
        \end{customlegend}
    \end{tikzpicture}
    
    \vspace{0.1cm}

    % 2x2 子图布局
    \begin{subfigure}{0.48\linewidth}
        \centering
        \begin{tikzpicture}
            \begin{axis}[ylabel={BLEU}]
            % DiffuSeq(2000) - Slow
            \addplot[color=blue!70!black, mark=*, thick] coordinates {(1,0.2971)(2,0.3078)(3,0.3327)(4,0.3455)(5,0.3504)(6,0.3536)(7,0.3572)(8,0.3583)(9,0.3631)(10,0.3644)};
            % DiffuSeq(DPM, 10) - Fast
            \addplot[color=black, mark=triangle*, thick, dashed] coordinates {(1,0.2318)(2,0.2287)(3,0.2316)(4,0.2329)(5,0.2338)(6,0.2339)(7,0.2346)(8,0.2340)(9,0.2348)(10,0.2345)};
            % FlowLM(step=5) - Fast
            \addplot[color=orange!90!black, mark=square*, thick] coordinates {(1,0.2527)(2,0.2548)(3,0.2932)(4,0.3100)(5,0.3204)(6,0.3278)(7,0.3335)(8,0.3371)(9,0.3404)(10,0.3430)};
            % FlowLM(step=3) - Fast
            \addplot[color=green!70!black, mark=diamond*, thick] coordinates {(1,0.2484)(2,0.2519)(3,0.2753)(4,0.2883)(5,0.2984)(6,0.3042)(7,0.3091)(8,0.3125)(9,0.3145)(10,0.3178)};
            % FlowLM(step=1) - Fast
            \addplot[color=red!80!black, mark=x, thick] coordinates {(1,0.2274)(2,0.2270)(3,0.2279)(4,0.2286)(5,0.2289)(6,0.2304)(7,0.2292)(8,0.2298)(9,0.2297)(10,0.2303)};
            \end{axis}
        \end{tikzpicture}
        \caption{BLEU}
    \end{subfigure}
    \hfill
    \begin{subfigure}{0.48\linewidth}
        \centering
        \begin{tikzpicture}
            \begin{axis}[ylabel={ROUGE-L}]
            % DiffuSeq(2000)
            \addplot[color=blue!70!black, mark=*, thick] coordinates {(1,0.5330)(2,0.5431)(3,0.5614)(4,0.5718)(5,0.5756)(6,0.5771)(7,0.5799)(8,0.5814)(9,0.5859)(10,0.5867)};
            % DiffuSeq(DPM, 10)
            \addplot[color=black, mark=triangle*, thick, dashed] coordinates {(1,0.4674)(2,0.4663)(3,0.4687)(4,0.4698)(5,0.4704)(6,0.4705)(7,0.4705)(8,0.4706)(9,0.4708)(10,0.4709)};
            % FlowLM(step=5)
            \addplot[color=orange!90!black, mark=square*, thick] coordinates {(1,0.4850)(2,0.4857)(3,0.5210)(4,0.5352)(5,0.5458)(6,0.5516)(7,0.5573)(8,0.5609)(9,0.5639)(10,0.5660)};
            % FlowLM(step=3)
            \addplot[color=green!70!black, mark=diamond*, thick] coordinates {(1,0.4798)(2,0.4819)(3,0.5042)(4,0.5158)(5,0.5242)(6,0.5295)(7,0.5344)(8,0.5374)(9,0.5396)(10,0.5427)};
            % FlowLM(step=1)
            \addplot[color=red!80!black, mark=x, thick] coordinates {(1,0.4440)(2,0.4438)(3,0.4445)(4,0.4452)(5,0.4458)(6,0.4466)(7,0.4459)(8,0.4463)(9,0.4461)(10,0.4463)};
            \end{axis}
        \end{tikzpicture}
        \caption{ROUGE-L}
    \end{subfigure}

    \vspace{0.2cm} % 行间距

    \begin{subfigure}{0.48\linewidth}
        \centering
        \begin{tikzpicture}
            \begin{axis}[ylabel={BERTScore}]
            % DiffuSeq(2000)
            \addplot[color=blue!70!black, mark=*, thick] coordinates {(1,0.7787)(2,0.7833)(3,0.7951)(4,0.8022)(5,0.8057)(6,0.8070)(7,0.8090)(8,0.8103)(9,0.8125)(10,0.8136)};
            % DiffuSeq(DPM, 10)
            \addplot[color=black, mark=triangle*, thick, dashed] coordinates {(1,0.6896)(2,0.6882)(3,0.6904)(4,0.6920)(5,0.6923)(6,0.6919)(7,0.6925)(8,0.6923)(9,0.6926)(10,0.6926)};
            % FlowLM(step=5)
            \addplot[color=orange!90!black, mark=square*, thick] coordinates {(1,0.7293)(2,0.7280)(3,0.7545)(4,0.7654)(5,0.7729)(6,0.7780)(7,0.7822)(8,0.7855)(9,0.7877)(10,0.7896)};
            % FlowLM(step=3)
            \addplot[color=green!70!black, mark=diamond*, thick] coordinates {(1,0.7122)(2,0.7126)(3,0.7304)(4,0.7409)(5,0.7478)(6,0.7520)(7,0.7564)(8,0.7591)(9,0.7615)(10,0.7635)};
            % FlowLM(step=1)
            \addplot[color=red!80!black, mark=x, thick] coordinates {(1,0.6332)(2,0.6329)(3,0.6347)(4,0.6352)(5,0.6360)(6,0.6364)(7,0.6357)(8,0.6360)(9,0.6361)(10,0.6362)};
            \end{axis}
        \end{tikzpicture}
        \caption{BERTScore}
    \end{subfigure}
    \hfill
    \begin{subfigure}{0.48\linewidth}
        \centering
        \begin{tikzpicture}
            \begin{axis}[ylabel={Dist-1}]
            % DiffuSeq(2000)
            \addplot[color=blue!70!black, mark=*, thick] coordinates {(1,0.9263)(2,0.9135)(3,0.9233)(4,0.9239)(5,0.9262)(6,0.9259)(7,0.9261)(8,0.9261)(9,0.9257)(10,0.9254)};
            % DiffuSeq(DPM, 10)
            \addplot[color=black, mark=triangle*, thick, dashed] coordinates {(1,0.8795)(2,0.8828)(3,0.8802)(4,0.8804)(5,0.8798)(6,0.8807)(7,0.8801)(8,0.8804)(9,0.8805)(10,0.8802)};
            % FlowLM(step=5)
            \addplot[color=orange!90!black, mark=square*, thick] coordinates {(1,0.9022)(2,0.8962)(3,0.9057)(4,0.9063)(5,0.9081)(6,0.9085)(7,0.9099)(8,0.9109)(9,0.9118)(10,0.9127)};
            % FlowLM(step=3)
            \addplot[color=green!70!black, mark=diamond*, thick] coordinates {(1,0.8766)(2,0.8700)(3,0.8813)(4,0.8855)(5,0.8869)(6,0.8886)(7,0.8908)(8,0.8918)(9,0.8929)(10,0.8933)};
            % FlowLM(step=1)
            \addplot[color=red!80!black, mark=x, thick] coordinates {(1,0.7493)(2,0.7522)(3,0.7496)(4,0.7497)(5,0.7493)(6,0.7498)(7,0.7490)(8,0.7493)(9,0.7493)(10,0.7500)};
            \end{axis}
        \end{tikzpicture}
        \caption{Dist-1}
    \end{subfigure}

    \captionof{figure}{Text Simplification results analysis across MBR candidate sizes (1--10).}
    \label{fig:combined_results_ts}
\end{figure*}

\begin{table*}[t]
\centering
\caption{Detailed performance metrics for Text Simplification across all 10 MBR candidate levels. (Best in Fast is \textbf{bold}).}
\footnotesize 
\begin{tabular}{cclcccc}
\toprule
MBR ($n$) & Category & Model & BLEU & ROUGE-L & BERTScore & Dist-1 \\ 
\midrule
% --- n=1 ---
\multirow{5}{*}{1} 
 & Multi-step & Diffuseq(2000) & 0.2971 & 0.5330 & 0.7787 & 0.9263 \\
 \cmidrule(lr){2-7}
 & \multirow{4}{*}{Few-step} & Diffuseq(DPM, 10) & 0.2318 & 0.4674 & 0.6896 & 0.8795 \\
 & & FlowLM(step=5) & \textbf{0.2527} & \textbf{0.4850} & \textbf{0.7293} & \textbf{0.9022} \\
 & & FlowLM(step=3) & 0.2484 & 0.4798 & 0.7122 & 0.8766 \\
 & & FlowLM(step=1) & 0.2274 & 0.4440 & 0.6332 & 0.7493 \\
\midrule
% --- n=2 ---
\multirow{5}{*}{2} 
 & Multi-step & Diffuseq(2000) & 0.3078 & 0.5431 & 0.7833 & 0.9135 \\
 \cmidrule(lr){2-7}
 & \multirow{4}{*}{Few-step} & Diffuseq(DPM, 10) & 0.2287 & 0.4663 & 0.6882 & 0.8828 \\
 & & FlowLM(step=5) & \textbf{0.2548} & \textbf{0.4857} & \textbf{0.7280} & \textbf{0.8962} \\
 & & FlowLM(step=3) & 0.2519 & 0.4819 & 0.7126 & 0.8700 \\
 & & FlowLM(step=1) & 0.2270 & 0.4438 & 0.6329 & 0.7522 \\
\midrule
% --- n=3 ---
\multirow{5}{*}{3} 
 & Multi-step & Diffuseq(2000) & 0.3327 & 0.5614 & 0.7951 & 0.9233 \\
 \cmidrule(lr){2-7}
 & \multirow{4}{*}{Few-step} & Diffuseq(DPM, 10) & 0.2316 & 0.4687 & 0.6904 & 0.8802 \\
 & & FlowLM(step=5) & \textbf{0.2932} & \textbf{0.5210} & \textbf{0.7545} & \textbf{0.9057} \\
 & & FlowLM(step=3) & 0.2753 & 0.5042 & 0.7304 & 0.8813 \\
 & & FlowLM(step=1) & 0.2279 & 0.4445 & 0.6347 & 0.7496 \\
\midrule
% --- n=4 ---
\multirow{5}{*}{4} 
 & Multi-step & Diffuseq(2000) & 0.3455 & 0.5718 & 0.8022 & 0.9239 \\
 \cmidrule(lr){2-7}
 & \multirow{4}{*}{Few-step} & Diffuseq(DPM, 10) & 0.2329 & 0.4698 & 0.6920 & 0.8804 \\
 & & FlowLM(step=5) & \textbf{0.3100} & \textbf{0.5352} & \textbf{0.7654} & \textbf{0.9063} \\
 & & FlowLM(step=3) & 0.2883 & 0.5158 & 0.7409 & 0.8855 \\
 & & FlowLM(step=1) & 0.2286 & 0.4452 & 0.6352 & 0.7497 \\
\midrule
% --- n=5 ---
\multirow{5}{*}{5} 
 & Multi-step & Diffuseq(2000) & 0.3504 & 0.5756 & 0.8057 & 0.9262 \\
 \cmidrule(lr){2-7}
 & \multirow{4}{*}{Few-step} & Diffuseq(DPM, 10) & 0.2338 & 0.4704 & 0.6923 & 0.8798 \\
 & & FlowLM(step=5) & \textbf{0.3204} & \textbf{0.5458} & \textbf{0.7729} & \textbf{0.9081} \\
 & & FlowLM(step=3) & 0.2984 & 0.5242 & 0.7478 & 0.8869 \\
 & & FlowLM(step=1) & 0.2289 & 0.4458 & 0.6360 & 0.7493 \\
\midrule
% --- n=6 ---
\multirow{5}{*}{6} 
 & Multi-step & Diffuseq(2000) & 0.3536 & 0.5771 & 0.8070 & 0.9259 \\
 \cmidrule(lr){2-7}
 & \multirow{4}{*}{Few-step} & Diffuseq(DPM, 10) & 0.2339 & 0.4705 & 0.6919 & 0.8807 \\
 & & FlowLM(step=5) & \textbf{0.3278} & \textbf{0.5516} & \textbf{0.7780} & \textbf{0.9085} \\
 & & FlowLM(step=3) & 0.3042 & 0.5295 & 0.7520 & 0.8886 \\
 & & FlowLM(step=1) & 0.2304 & 0.4466 & 0.6364 & 0.7498 \\
\midrule
% --- n=7 ---
\multirow{5}{*}{7} 
 & Multi-step & Diffuseq(2000) & 0.3572 & 0.5799 & 0.8090 & 0.9261 \\
 \cmidrule(lr){2-7}
 & \multirow{4}{*}{Few-step} & Diffuseq(DPM, 10) & 0.2346 & 0.4705 & 0.6925 & 0.8801 \\
 & & FlowLM(step=5) & \textbf{0.3335} & \textbf{0.5573} & \textbf{0.7822} & \textbf{0.9099} \\
 & & FlowLM(step=3) & 0.3091 & 0.5344 & 0.7564 & 0.8908 \\
 & & FlowLM(step=1) & 0.2292 & 0.4459 & 0.6357 & 0.7490 \\
\midrule
% --- n=8 ---
\multirow{5}{*}{8} 
 & Multi-step & Diffuseq(2000) & 0.3583 & 0.5814 & 0.8103 & 0.9261 \\
 \cmidrule(lr){2-7}
 & \multirow{4}{*}{Few-step} & Diffuseq(DPM, 10) & 0.2340 & 0.4706 & 0.6923 & 0.8804 \\
 & & FlowLM(step=5) & \textbf{0.3371} & \textbf{0.5609} & \textbf{0.7855} & \textbf{0.9109} \\
 & & FlowLM(step=3) & 0.3125 & 0.5374 & 0.7591 & 0.8918 \\
 & & FlowLM(step=1) & 0.2298 & 0.4463 & 0.6360 & 0.7493 \\
\midrule
% --- n=9 ---
\multirow{5}{*}{9} 
 & Multi-step & Diffuseq(2000) & 0.3631 & 0.5859 & 0.8125 & 0.9257 \\
 \cmidrule(lr){2-7}
 & \multirow{4}{*}{Few-step} & Diffuseq(DPM, 10) & 0.2348 & 0.4708 & 0.6926 & 0.8805 \\
 & & FlowLM(step=5) & \textbf{0.3404} & \textbf{0.5639} & \textbf{0.7877} & \textbf{0.9118} \\
 & & FlowLM(step=3) & 0.3145 & 0.5396 & 0.7615 & 0.8929 \\
 & & FlowLM(step=1) & 0.2297 & 0.4461 & 0.6361 & 0.7493 \\
\midrule
% --- n=10 ---
\multirow{5}{*}{10} 
 & Multi-step & Diffuseq(2000) & 0.3644 & 0.5867 & 0.8136 & 0.9254 \\
 \cmidrule(lr){2-7}
 & \multirow{4}{*}{Few-step} & Diffuseq(DPM, 10) & 0.2345 & 0.4709 & 0.6926 & 0.8802 \\
 & & FlowLM(step=5) & \textbf{0.3430} & \textbf{0.5660} & \textbf{0.7896} & \textbf{0.9127} \\
 & & FlowLM(step=3) & 0.3178 & 0.5427 & 0.7635 & 0.8933 \\
 & & FlowLM(step=1) & 0.2303 & 0.4463 & 0.6362 & 0.7500 \\
\bottomrule
\end{tabular}
\label{table: TS}
\end{table*}
\FloatBarrier

\begin{figure*}[ht]
    \centering
    % --- 1. 图例 (放在顶部) ---
    \begin{tikzpicture}
        \begin{customlegend}[
            legend columns=-1,
            legend style={draw=none, column sep=0.5cm, font=\scriptsize},
            legend entries={FlowLM(step=1), FlowLM(step=3), FlowLM(step=5)}
        ]
        \addlegendimage{color=red!80!black, mark=x, thick}
        \addlegendimage{color=green!70!black, mark=diamond*, thick}
        \addlegendimage{color=orange!90!black, mark=square*, thick}
        \end{customlegend}
    \end{tikzpicture}

    % --- 绘图全局配置 (恢复到较大的高度 5cm) ---
    \pgfplotsset{
        every axis/.style={
            width=\linewidth,
            height=5.5cm,
            grid=major,
            grid style={dashed, gray!30},
            xlabel={Training epochs (Relative)},
            xmin=1000, xmax=6000,
            xtick={1000, 2000, 3000, 4000, 5000, 6000},
            xticklabels={1k, 2k, 3k, 4k, 5k, 6k},
            tick label style={font=\tiny},
            label style={font=\scriptsize},
            scaled y ticks=false,
            yticklabel style={/pgf/number format/.cd, fixed, fixed zerofill, precision=3},
        }
    }

    % --- 四个子图 (2x2 布局) ---
    \begin{subfigure}{0.48\linewidth}
        \centering
        \begin{tikzpicture}
            \begin{axis}[ylabel={BLEU}]
            \addplot[color=red!80!black, mark=x, thick] coordinates {(1000,0.1390)(2000,0.1415)(3000,0.1441)(4000,0.1468)(5000,0.1504)(6000,0.1522)};
            \addplot[color=green!70!black, mark=diamond*, thick] coordinates {(1000,0.1479)(2000,0.1510)(3000,0.1540)(4000,0.1555)(5000,0.1578)(6000,0.1592)};
            \addplot[color=orange!90!black, mark=square*, thick] coordinates {(1000,0.1479)(2000,0.1499)(3000,0.1530)(4000,0.1560)(5000,0.1590)(6000,0.1603)};
            \end{axis}
        \end{tikzpicture}
        \caption{BLEU}
    \end{subfigure}
    \hfill
    \begin{subfigure}{0.48\linewidth}
        \centering
        \begin{tikzpicture}
            \begin{axis}[ylabel={ROUGE-L}]
            \addplot[color=red!80!black, mark=x, thick] coordinates {(1000,0.3476)(2000,0.3491)(3000,0.3513)(4000,0.3527)(5000,0.3535)(6000,0.3552)};
            \addplot[color=green!70!black, mark=diamond*, thick] coordinates {(1000,0.3404)(2000,0.3433)(3000,0.3455)(4000,0.3473)(5000,0.3489)(6000,0.3495)};
            \addplot[color=orange!90!black, mark=square*, thick] coordinates {(1000,0.3387)(2000,0.3396)(3000,0.3428)(4000,0.3460)(5000,0.3486)(6000,0.3490)};
            \end{axis}
        \end{tikzpicture}
        \caption{ROUGE-L}
    \end{subfigure}

    \vspace{2mm}

    \begin{subfigure}{0.48\linewidth}
        \centering
        \begin{tikzpicture}
            \begin{axis}[ylabel={BERTScore}]
            \addplot[color=red!80!black, mark=x, thick] coordinates {(1000,0.5530)(2000,0.5566)(3000,0.5597)(4000,0.5638)(5000,0.5673)(6000,0.5702)};
            \addplot[color=green!70!black, mark=diamond*, thick] coordinates {(1000,0.5767)(2000,0.5796)(3000,0.5817)(4000,0.5830)(5000,0.5860)(6000,0.5879)};
            \addplot[color=orange!90!black, mark=square*, thick] coordinates {(1000,0.5783)(2000,0.5805)(3000,0.5841)(4000,0.5855)(5000,0.5884)(6000,0.5901)};
            \end{axis}
        \end{tikzpicture}
        \caption{BERTScore}
    \end{subfigure}
    \hfill
    \begin{subfigure}{0.48\linewidth}
        \centering
        \begin{tikzpicture}
            \begin{axis}[ylabel={Dist-1}]
            \addplot[color=red!80!black, mark=x, thick] coordinates {(1000,0.8022)(2000,0.8085)(3000,0.8160)(4000,0.8250)(5000,0.8347)(6000,0.8416)};
            \addplot[color=green!70!black, mark=diamond*, thick] coordinates {(1000,0.9120)(2000,0.9136)(3000,0.9151)(4000,0.9155)(5000,0.9165)(6000,0.9185)};
            \addplot[color=orange!90!black, mark=square*, thick] coordinates {(1000,0.9149)(2000,0.9169)(3000,0.9176)(4000,0.9194)(5000,0.9205)(6000,0.9224)};
            \end{axis}
        \end{tikzpicture}
        \caption{Dist-1}
    \end{subfigure}
    \caption{Ablation analysis on Training epochs (mapped to 1k--6k) for the Question Generation task. We compare FlowLM performance under 1, 3, and 5 sampling steps.}
    \label{fig:qg_ablation_original_size}
\end{figure*}
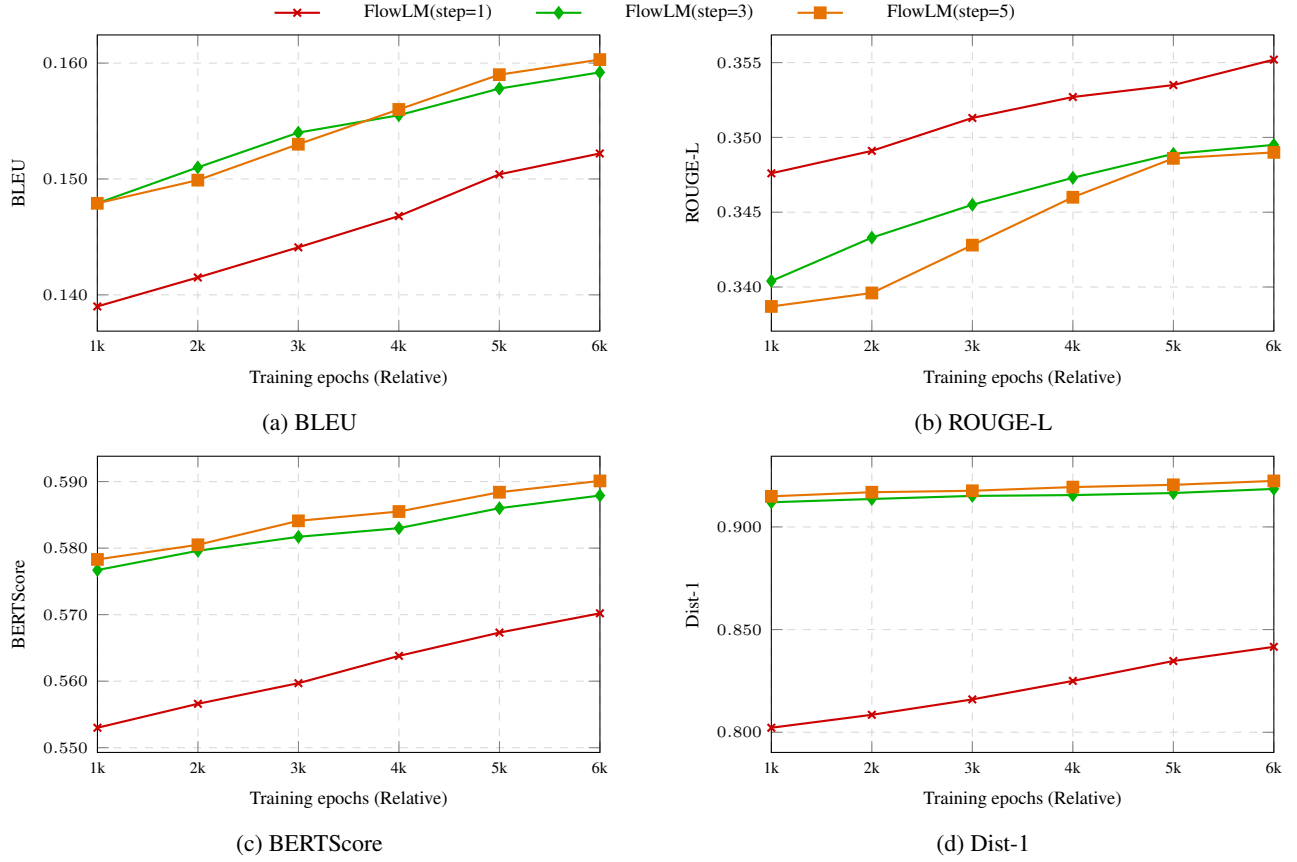

\begin{figure*}[ht]
    \centering
    % --- 1. 图例 ---
    \begin{tikzpicture}
        \begin{customlegend}[
            legend columns=-1,
            legend style={draw=none, column sep=0.5cm, font=\scriptsize},
            legend entries={FlowLM(step=1), FlowLM(step=3), FlowLM(step=5)}
        ]
        \addlegendimage{color=red!80!black, mark=x, thick}
        \addlegendimage{color=green!70!black, mark=diamond*, thick}
        \addlegendimage{color=orange!90!black, mark=square*, thick}
        \end{customlegend}
    \end{tikzpicture}

    % --- 绘图全局配置 ---
    \pgfplotsset{
        every axis/.style={
            width=\linewidth,
            height=5.5cm,
            grid=major,
            grid style={dashed, gray!30},
            xlabel={Training epochs (Relative)},
            xmin=1000, xmax=10000,
            xtick={1000, 2000, 3000, 4000, 5000, 6000, 7000, 8000, 9000, 10000},
            xticklabels={1k,, 3k,, 5k,, 7k,, 9k,}, % 隔开显示避免重叠
            tick label style={font=\tiny},
            label style={font=\scriptsize},
            scaled y ticks=false,
            yticklabel style={/pgf/number format/.cd, fixed, fixed zerofill, precision=3},
        }
    }

    % --- 四个子图 (2x2 布局) ---
    \begin{subfigure}{0.48\linewidth}
        \centering
        \begin{tikzpicture}
            \begin{axis}[ylabel={BLEU}]
            \addplot[color=red!80!black, mark=x, thick] coordinates {(1000,0.1628)(2000,0.1666)(3000,0.1666)(4000,0.1683)(5000,0.1714)(6000,0.1763)(7000,0.1797)(8000,0.1838)(9000,0.1868)(10000,0.1896)};
            \addplot[color=green!70!black, mark=diamond*, thick] coordinates {(1000,0.1722)(2000,0.1697)(3000,0.1779)(4000,0.1836)(5000,0.1852)(6000,0.1878)(7000,0.1900)(8000,0.1920)(9000,0.1970)(10000,0.1978)};
            \addplot[color=orange!90!black, mark=square*, thick] coordinates {(1000,0.1524)(2000,0.1587)(3000,0.1626)(4000,0.1665)(5000,0.1790)(6000,0.1805)(7000,0.1805)(8000,0.1851)(9000,0.1943)(10000,0.1920)};
            \end{axis}
        \end{tikzpicture}
        \caption{BLEU}
    \end{subfigure}
    \hfill
    \begin{subfigure}{0.48\linewidth}
        \centering
        \begin{tikzpicture}
            \begin{axis}[ylabel={ROUGE-L}]
            \addplot[color=red!80!black, mark=x, thick] coordinates {(1000,0.5070)(2000,0.5098)(3000,0.5138)(4000,0.5168)(5000,0.5189)(6000,0.5232)(7000,0.5270)(8000,0.5322)(9000,0.5352)(10000,0.5379)};
            \addplot[color=green!70!black, mark=diamond*, thick] coordinates {(1000,0.5017)(2000,0.5011)(3000,0.5071)(4000,0.5139)(5000,0.5137)(6000,0.5220)(7000,0.5247)(8000,0.5284)(9000,0.5333)(10000,0.5354)};
            \addplot[color=orange!90!black, mark=square*, thick] coordinates {(1000,0.4829)(2000,0.4875)(3000,0.4980)(4000,0.4996)(5000,0.5078)(6000,0.5155)(7000,0.5155)(8000,0.5242)(9000,0.5338)(10000,0.5331)};
            \end{axis}
        \end{tikzpicture}
        \caption{ROUGE-L}
    \end{subfigure}

    \vspace{2mm}

    \begin{subfigure}{0.48\linewidth}
        \centering
        \begin{tikzpicture}
            \begin{axis}[ylabel={BERTScore}]
            \addplot[color=red!80!black, mark=x, thick] coordinates {(1000,0.7166)(2000,0.7185)(3000,0.7232)(4000,0.7241)(5000,0.7288)(6000,0.7347)(7000,0.7402)(8000,0.7449)(9000,0.7511)(10000,0.7557)};
            \addplot[color=green!70!black, mark=diamond*, thick] coordinates {(1000,0.7470)(2000,0.7469)(3000,0.7498)(4000,0.7567)(5000,0.7592)(6000,0.7618)(7000,0.7680)(8000,0.7718)(9000,0.7771)(10000,0.7782)};
            \addplot[color=orange!90!black, mark=square*, thick] coordinates {(1000,0.7429)(2000,0.7481)(3000,0.7553)(4000,0.7581)(5000,0.7629)(6000,0.7683)(7000,0.7683)(8000,0.7783)(9000,0.7835)(10000,0.7841)};
            \end{axis}
        \end{tikzpicture}
        \caption{BERTScore}
    \end{subfigure}
    \hfill
    \begin{subfigure}{0.48\linewidth}
        \centering
        \begin{tikzpicture}
            \begin{axis}[ylabel={Dist-1}]
            \addplot[color=red!80!black, mark=x, thick] coordinates {(1000,0.9374)(2000,0.9399)(3000,0.9433)(4000,0.9445)(5000,0.9465)(6000,0.9487)(7000,0.9475)(8000,0.9483)(9000,0.9479)(10000,0.9439)};
            \addplot[color=green!70!black, mark=diamond*, thick] coordinates {(1000,0.9699)(2000,0.9719)(3000,0.9722)(4000,0.9750)(5000,0.9732)(6000,0.9720)(7000,0.9738)(8000,0.9743)(9000,0.9744)(10000,0.9767)};
            \addplot[color=orange!90!black, mark=square*, thick] coordinates {(1000,0.9721)(2000,0.9727)(3000,0.9736)(4000,0.9737)(5000,0.9753)(6000,0.9751)(7000,0.9751)(8000,0.9757)(9000,0.9761)(10000,0.9766)};
            \end{axis}
        \end{tikzpicture}
        \caption{Dist-1}
    \end{subfigure}
    \caption{Ablation analysis on Training epochs (mapped to 1k--10k) for the Paraphrase task. Results demonstrate consistent quality gains as the relative training budget increases.}
    \label{fig:para_ablation_original_size}
\end{figure*}
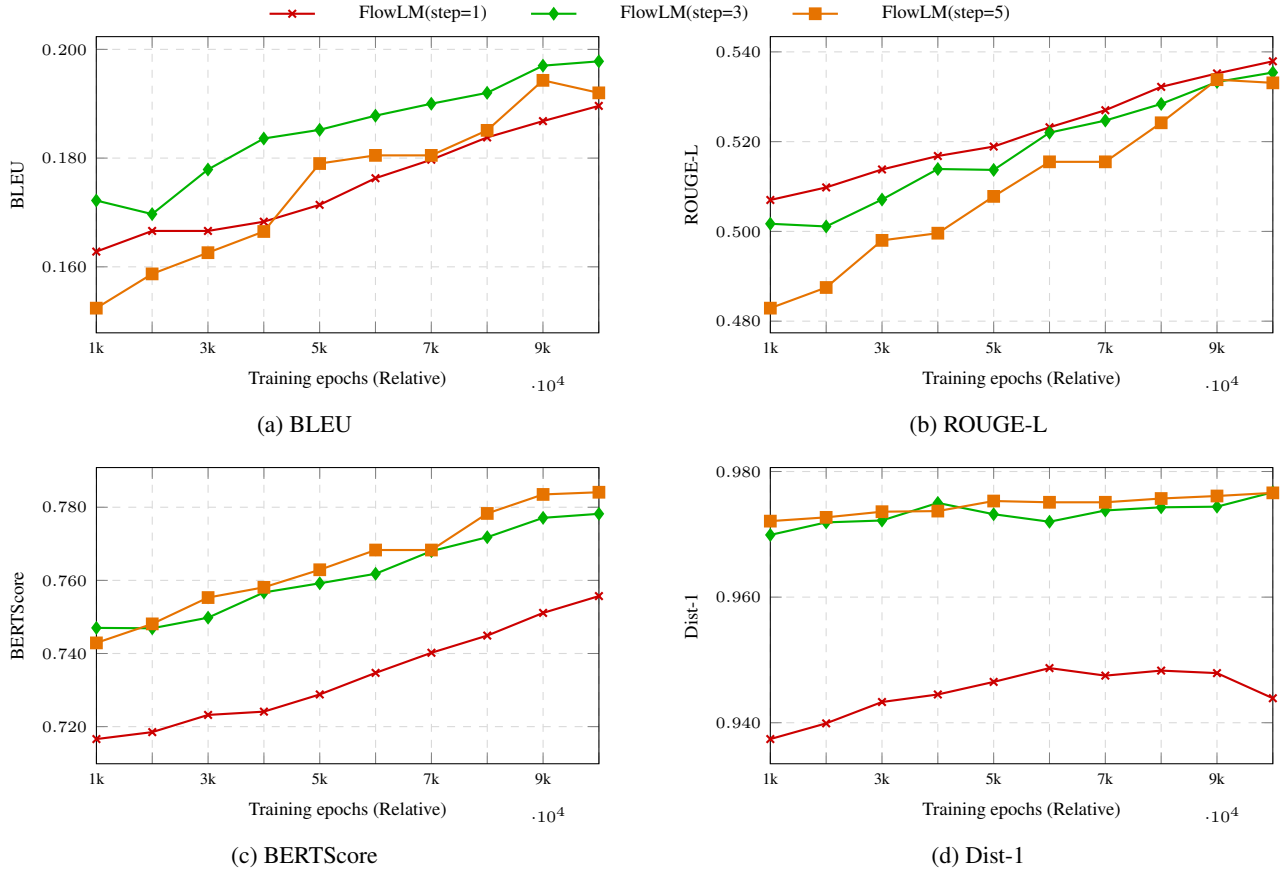
\FloatBarrier

\FloatBarrier
\newpage
% \section{More comparison on different training strategies}
% \label{sec:appendix_training_steps}

% \subsection{The number of possible training time steps}
% In standard diffusion models, the training process typically involves sampling from a dense discretization of the time horizon, often up to $T=2000$ steps. While this fine-grained discretization allows the model to approximate the continuous ODE/SDE accurately, it creates a misalignment when the goal is rapid, few-step inference. Our \textbf{FlowLM} model is explicitly designed to operate within a few-step regime (e.g., $N \in \{1, 3, 5\}$). 

% We hypothesize that training on a massive number of steps (e.g., 2000) forces the model to learn local vector fields for intermediate states that are skipped during fast inference, potentially leading to inefficient allocation of model capacity. Conversely, reducing the training time steps to match the inference scale might improve focus, though significantly reducing $T$ carries the risk of overfitting or failing to capture the complex data distribution. To verify this, we perform an ablation study on the Paraphrase task, comparing our optimized FlowLM (trained with $T=20$) against a baseline Flow Matching model trained with $T=2000$ (denoted as fm2k).

% =======================================================
% FIGURE: Hyperparameters + MBR Curves (包含所有 Step)
% =======================================================
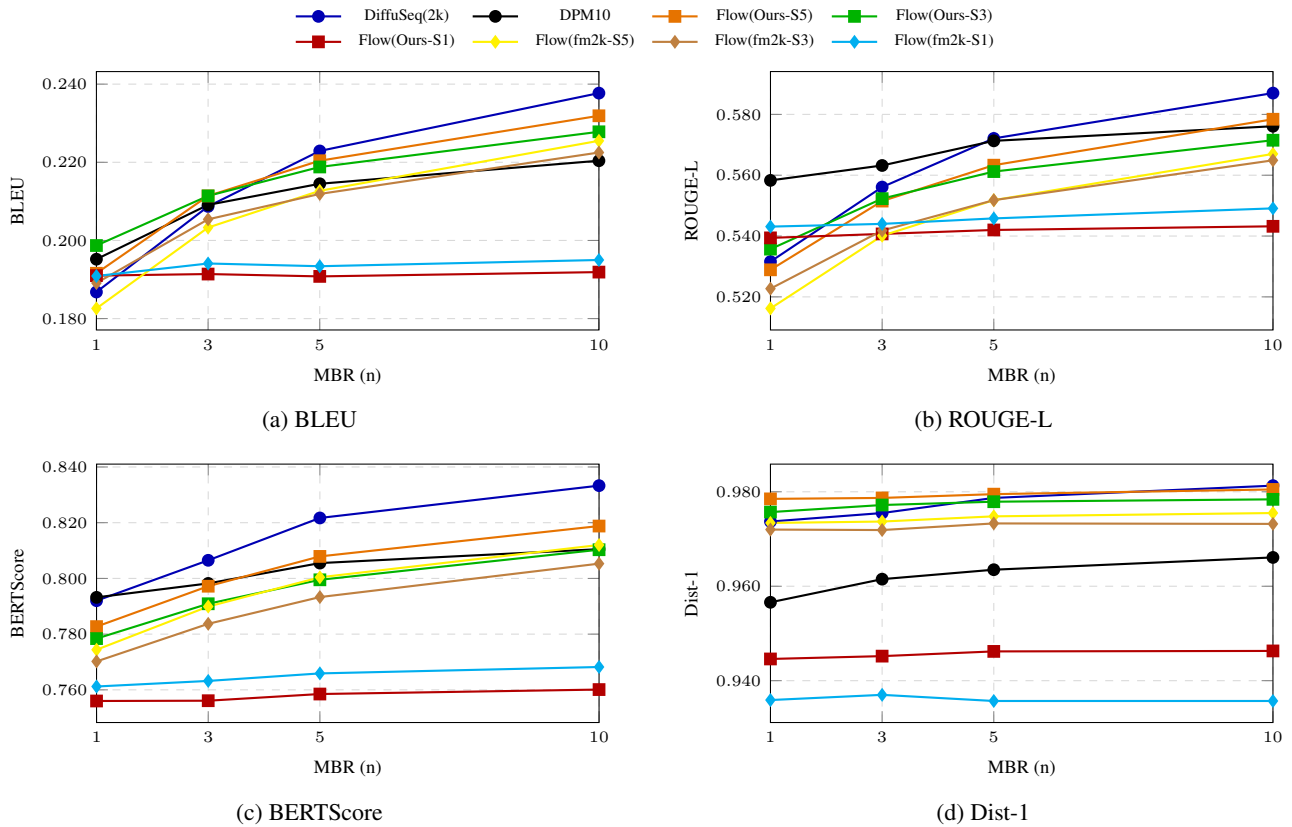
\begin{figure*}[ht]
    \centering
    \begin{minipage}{\textwidth}
        \section{More comparison on different training strategies}
\label{sec:appendix_training_steps}

In standard diffusion models, the training process typically involves sampling from a dense discretization of the time horizon, often up to $T=2000$ steps. While this fine-grained discretization allows the model to approximate the continuous ODE/SDE accurately, it creates a misalignment when the goal is rapid, few-step inference. Our \textbf{FlowLM} model is explicitly designed to operate within a few-step regime (e.g., $N \in \{1, 3, 5\}$). 

We hypothesize that training on a massive number of steps (e.g., 2000) forces the model to learn local vector fields for intermediate states that are skipped during fast inference, potentially leading to inefficient allocation of model capacity. Conversely, reducing the training time steps to match the inference scale might improve focus, though significantly reducing $T$ carries the risk of overfitting or failing to capture the complex data distribution. To verify this, we perform an ablation study on the Paraphrase task, comparing our optimized FlowLM (trained with $T=20$) against a baseline Flow Matching model trained with $T=2000$ (denoted as fm2k).
    \end{minipage}
    % --- 绘图配置 ---
    \pgfplotsset{
        every axis/.style={
            width=\linewidth, height=5cm, grid=major, grid style={dashed, gray!30},
            xlabel={MBR (n)}, xmin=1, xmax=10, xtick={1,3,5,10},
            tick label style={font=\tiny}, scaled y ticks=false,
            yticklabel style={/pgf/number format/.cd, fixed, fixed zerofill, precision=3},
            label style={font=\scriptsize},
        }
    }

    % --- 图例 (Legend) ---
    \begin{tikzpicture}
        \begin{customlegend}[
            legend columns=4, legend style={draw=none, column sep=0.15cm, font=\tiny},
            legend entries={DiffuSeq(2k), DPM10, Flow(Ours-S5), Flow(Ours-S3), Flow(Ours-S1), Flow(fm2k-S5), Flow(fm2k-S3), Flow(fm2k-S1)}
        ]
        \addlegendimage{color=blue!70!black, mark=*, thick}         % DiffuSeq
        \addlegendimage{color=black, mark=*, thick}                 % DPM
        \addlegendimage{color=orange!90!black, mark=square*, thick} % Ours S5
        \addlegendimage{color=green!70!black, mark=square*, thick}  % Ours S3
        \addlegendimage{color=red!70!black, mark=square*, thick}    % Ours S1
        \addlegendimage{color=yellow, mark=diamond*, thick}         % fm2k S5
        \addlegendimage{color=brown, mark=diamond*, thick}          % fm2k S3
        \addlegendimage{color=cyan, mark=diamond*, thick}           % fm2k S1
        \end{customlegend}
    \end{tikzpicture}
    
    \vspace{0.1cm}

    % --- 子图 1: BLEU ---
    \begin{subfigure}{0.48\linewidth}
        \centering
        \begin{tikzpicture}
            \begin{axis}[ylabel={BLEU}]
            \addplot[color=blue!70!black, mark=*, thick] coordinates {(1,0.1868)(3,0.2087)(5,0.2229)(10,0.2377)};
            \addplot[color=black, mark=*, thick] coordinates {(1,0.1952)(3,0.2091)(5,0.2145)(10,0.2204)};
            \addplot[color=orange!90!black, mark=square*, thick] coordinates {(1,0.1916)(3,0.2114)(5,0.2204)(10,0.2319)};
            \addplot[color=green!70!black, mark=square*, thick] coordinates {(1,0.1987)(3,0.2114)(5,0.2188)(10,0.2278)};
            \addplot[color=red!70!black, mark=square*, thick] coordinates {(1,0.1910)(3,0.1914)(5,0.1908)(10,0.1919)};
            \addplot[color=yellow, mark=diamond*, thick] coordinates {(1,0.1826)(3,0.2033)(5,0.2127)(10,0.2255)};
            \addplot[color=brown, mark=diamond*, thick] coordinates {(1,0.1891)(3,0.2054)(5,0.2119)(10,0.2225)};
            \addplot[color=cyan, mark=diamond*, thick] coordinates {(1,0.1909)(3,0.1941)(5,0.1934)(10,0.1950)};
            \end{axis}
        \end{tikzpicture}
        \caption{BLEU}
    \end{subfigure}
    \hfill
    % --- 子图 2: ROUGE-L ---
    \begin{subfigure}{0.48\linewidth}
        \centering
        \begin{tikzpicture}
            \begin{axis}[ylabel={ROUGE-L}]
            \addplot[color=blue!70!black, mark=*, thick] coordinates {(1,0.5316)(3,0.5561)(5,0.5721)(10,0.5870)};
            \addplot[color=black, mark=*, thick] coordinates {(1,0.5583)(3,0.5632)(5,0.5713)(10,0.5761)};
            \addplot[color=orange!90!black, mark=square*, thick] coordinates {(1,0.5289)(3,0.5515)(5,0.5633)(10,0.5784)};
            \addplot[color=green!70!black, mark=square*, thick] coordinates {(1,0.5357)(3,0.5523)(5,0.5612)(10,0.5715)};
            \addplot[color=red!70!black, mark=square*, thick] coordinates {(1,0.5394)(3,0.5407)(5,0.5420)(10,0.5432)};
            \addplot[color=yellow, mark=diamond*, thick] coordinates {(1,0.5162)(3,0.5401)(5,0.5518)(10,0.5670)};
            \addplot[color=brown, mark=diamond*, thick] coordinates {(1,0.5227)(3,0.5418)(5,0.5518)(10,0.5649)};
            \addplot[color=cyan, mark=diamond*, thick] coordinates {(1,0.5431)(3,0.5440)(5,0.5458)(10,0.5491)};
            \end{axis}
        \end{tikzpicture}
        \caption{ROUGE-L}
    \end{subfigure}

    \vspace{0.2cm}

    % --- 子图 3: BERTScore (数据已补全) ---
    \begin{subfigure}{0.48\linewidth}
        \centering
        \begin{tikzpicture}
            \begin{axis}[ylabel={BERTScore}]
            % DiffuSeq(2k)
            \addplot[color=blue!70!black, mark=*, thick] coordinates {(1,0.7920)(3,0.8065)(5,0.8217)(10,0.8333)};
            % DPM
            \addplot[color=black, mark=*, thick] coordinates {(1,0.7932)(3,0.7982)(5,0.8055)(10,0.8105)};
            % Ours S5
            \addplot[color=orange!90!black, mark=square*, thick] coordinates {(1,0.7827)(3,0.7972)(5,0.8079)(10,0.8188)};
            % Ours S3
            \addplot[color=green!70!black, mark=square*, thick] coordinates {(1,0.7784)(3,0.7909)(5,0.7995)(10,0.8103)};
            % Ours S1
            \addplot[color=red!70!black, mark=square*, thick] coordinates {(1,0.7560)(3,0.7561)(5,0.7585)(10,0.7601)};
            % fm2k S5
            \addplot[color=yellow, mark=diamond*, thick] coordinates {(1,0.7744)(3,0.7899)(5,0.8004)(10,0.8120)};
            % fm2k S3
            \addplot[color=brown, mark=diamond*, thick] coordinates {(1,0.7702)(3,0.7837)(5,0.7933)(10,0.8053)};
            % fm2k S1
            \addplot[color=cyan, mark=diamond*, thick] coordinates {(1,0.7612)(3,0.7632)(5,0.7659)(10,0.7682)};
            \end{axis}
        \end{tikzpicture}
        \caption{BERTScore}
    \end{subfigure}
    \hfill
    % --- 子图 4: Dist-1 (数据已补全) ---
    \begin{subfigure}{0.48\linewidth}
        \centering
        \begin{tikzpicture}
            \begin{axis}[ylabel={Dist-1}]
            % DiffuSeq(2k)
            \addplot[color=blue!70!black, mark=*, thick] coordinates {(1,0.9737)(3,0.9755)(5,0.9787)(10,0.9813)};
            % DPM
            \addplot[color=black, mark=*, thick] coordinates {(1,0.9566)(3,0.9615)(5,0.9635)(10,0.9661)};
            % Ours S5
            \addplot[color=orange!90!black, mark=square*, thick] coordinates {(1,0.9785)(3,0.9787)(5,0.9795)(10,0.9805)};
            % Ours S3
            \addplot[color=green!70!black, mark=square*, thick] coordinates {(1,0.9757)(3,0.9772)(5,0.9779)(10,0.9784)};
            % Ours S1
            \addplot[color=red!70!black, mark=square*, thick] coordinates {(1,0.9446)(3,0.9452)(5,0.9462)(10,0.9463)};
            % fm2k S5
            \addplot[color=yellow, mark=diamond*, thick] coordinates {(1,0.9734)(3,0.9737)(5,0.9748)(10,0.9755)};
            % fm2k S3
            \addplot[color=brown, mark=diamond*, thick] coordinates {(1,0.9720)(3,0.9719)(5,0.9733)(10,0.9732)};
            % fm2k S1
            \addplot[color=cyan, mark=diamond*, thick] coordinates {(1,0.9359)(3,0.9370)(5,0.9357)(10,0.9357)};
            \end{axis}
        \end{tikzpicture}
        \caption{Dist-1}
    \end{subfigure}
    \caption{Paraphrase experimental results. Our optimized Flow Matching (Ours) compared with fm\_num\_steps=2000 version (fm2k) and DiffuSeq baselines.}
    \label{fig:time_steps}
\end{figure*}
\FloatBarrier
% =======================================================
% TABLE: 全量 MBR 1, 3, 5, 10 数据对比
% =======================================================
\begin{table*}[t]
\centering
\caption{Comprehensive comparison of Paraphrase results for all versions across multiple MBR candidate sizes ($n \in \{1, 3, 5, 10\}$).}
\begin{tabular}{cclcccc}
\toprule
MBR ($n$) & Category & Model & BLEU$\uparrow$ & R-L$\uparrow$ & BERTScore$\uparrow$ & dist-1$\uparrow$ \\ 
\midrule
% --- MBR 1 ---
\multirow{8}{*}{1} 
 & Multi-step & Diffuseq(2000) & 0.1868 & 0.5316 & 0.7920 & 0.9737 \\
 \cmidrule(lr){2-7}
 & \multirow{7}{*}{Few-step} & Diffuseq(DPM, 10) & 0.1952 & \textbf{0.5583} & \textbf{0.7932} & 0.9566 \\
 & & FlowLM (Ours, S5) & 0.1916 & 0.5289 & 0.7827 & \textbf{0.9785} \\
 & & FlowLM (Ours, S3) & \textbf{0.1987} & 0.5357 & 0.7784 & 0.9757 \\
 & & FlowLM (Ours, S1) & 0.1910 & 0.5394 & 0.7560 & 0.9446 \\
 & & FlowLM (fm2k, S5) & 0.1826 & 0.5162 & 0.7744 & 0.9734 \\
 & & FlowLM (fm2k, S3) & 0.1891 & 0.5227 & 0.7702 & 0.9720 \\
 & & FlowLM (fm2k, S1) & 0.1909 & 0.5431 & 0.7612 & 0.9359 \\
\midrule
% --- MBR 3 ---
\multirow{8}{*}{3} 
 & Multi-step & Diffuseq(2000) & 0.2087 & 0.5561 & \textbf{0.8065} & 0.9755 \\
 \cmidrule(lr){2-7}
 & \multirow{7}{*}{Few-step} & Diffuseq(DPM, 10) & 0.2091 & \textbf{0.5632} & 0.7982 & 0.9615 \\
 & & FlowLM (Ours, S5) & \textbf{0.2114} & 0.5515 & 0.7972 & \textbf{0.9787} \\
 & & FlowLM (Ours, S3) & \textbf{0.2114} & 0.5523 & 0.7909 & 0.9772 \\
 & & FlowLM (Ours, S1) & 0.1914 & 0.5407 & 0.7561 & 0.9452 \\
 & & FlowLM (fm2k, S5) & 0.2033 & 0.5401 & 0.7899 & 0.9737 \\
 & & FlowLM (fm2k, S3) & 0.2054 & 0.5418 & 0.7837 & 0.9719 \\
 & & FlowLM (fm2k, S1) & 0.1941 & 0.5440 & 0.7632 & 0.9370 \\
\midrule
% --- MBR 5 ---
\multirow{8}{*}{5} 
 & Multi-step & Diffuseq(2000) & \textbf{0.2229} & \textbf{0.5721} & \textbf{0.8217} & 0.9787 \\
 \cmidrule(lr){2-7}
 & \multirow{7}{*}{Few-step} & Diffuseq(DPM, 10) & 0.2145 & 0.5713 & 0.8055 & 0.9635 \\
 & & FlowLM (Ours, S5) & 0.2204 & 0.5633 & 0.8079 & \textbf{0.9795} \\
 & & FlowLM (Ours, S3) & 0.2188 & 0.5612 & 0.7995 & 0.9779 \\
 & & FlowLM (Ours, S1) & 0.1908 & 0.5420 & 0.7585 & 0.9462 \\
 & & FlowLM (fm2k, S5) & 0.2127 & 0.5518 & 0.8004 & 0.9748 \\
 & & FlowLM (fm2k, S3) & 0.2119 & 0.5518 & 0.7933 & 0.9733 \\
 & & FlowLM (fm2k, S1) & 0.1934 & 0.5458 & 0.7659 & 0.9357 \\
\midrule
% --- MBR 10 ---
\multirow{8}{*}{10} 
 & Multi-step & Diffuseq(2000) & \textbf{0.2377} & \textbf{0.5870} & \textbf{0.8333} & \textbf{0.9813} \\
 \cmidrule(lr){2-7}
 & \multirow{7}{*}{Few-step} & Diffuseq(DPM, 10) & 0.2204 & 0.5761 & 0.8105 & 0.9661 \\
 & & FlowLM (Ours, S5) & 0.2319 & 0.5784 & 0.8188 & 0.9805 \\
 & & FlowLM (Ours, S3) & 0.2278 & 0.5715 & 0.8103 & 0.9784 \\
 & & FlowLM (Ours, S1) & 0.1919 & 0.5432 & 0.7601 & 0.9463 \\
 & & FlowLM (fm2k, S5) & 0.2255 & 0.5670 & 0.8120 & 0.9755 \\
 & & FlowLM (fm2k, S3) & 0.2225 & 0.5649 & 0.8053 & 0.9732 \\
 & & FlowLM (fm2k, S1) & 0.1950 & 0.5491 & 0.7682 & 0.9357 \\
\bottomrule
\end{tabular}

\label{tab:table_time_steps}
\begin{minipage}{\textwidth}
\vspace{0.5em}
The quantitative results are presented in Table \ref{tab:table_time_steps} and figure \ref{fig:time_steps} above. It is evident that the model trained with a reduced number of time steps (Ours) significantly outperforms the fm2k baseline across all evaluated metrics (BLEU, ROUGE-L, BERTScore, and Dist-1), particularly under strict few-step inference constraints ($N=1$ to $5$).

We attribute this performance gap to two main factors:
\begin{itemize}
    \item \textbf{Training-Inference Alignment:} The fm2k model distributes its learning capacity across 2000 discrete timesteps. During few-step inference, the solver takes large stride sizes, jumping over many of these learned steps. This renders the fine-grained information learned during training redundant and potentially noisy for the straight flow trajectory required for fast sampling.
    \item \textbf{Capacity Concentration:} By training with fewer potential time steps (e.g., 20), FlowLM focuses on learning the global flow trajectory rather than local fluctuations. This allows the model to construct a straighter and more stable probability flow, which is crucial for maintaining generation quality when $N$ is small.
\end{itemize}

Therefore, we conclude that for few-step diffusion/flow models, aligning the training discretization granularity with the target inference budget is a more effective strategy than using standard high-resolution training.
\end{minipage}
\end{table*}

\FloatBarrier

\begin{table*}[t]
\begin{minipage}{\textwidth}
    We further provide more ablation experiments on different T numbers(5,20,100,2000), T is the number of steps t sample from during training.  We also provide comparison between different input time-step rescale(20,200,1000), which represents the maximum value to which the input time-steps are rescaled. Furthermore, we provide ablation results of the new regularization loss term.
\end{minipage}
\centering
\caption{Comprehensive comparison of Paraphrase results for different T values(5,20,100)}
\renewcommand{\arraystretch}{1.1} %稍微增加行高，让表格不那么拥挤
% 定义列格式：
% l: 左对齐
% c: 居中
% |: 竖线
\begin{tabular}{lll | ccc | cc }
\toprule
Tasks & Type &Methods & BLEU$\uparrow$ & R-L$\uparrow$ & BERTScore$\uparrow$ & dist-1$\uparrow$ &Training epoch\\
\midrule

\hline 
% ==================== Task 2: Paraphrase ====================
\multirow{12}{*}{\shortstack[l]{Paraphrase}} 
 & \multirow{12}{*}{Few-step}& FlowLM(T=2000, step=5) & 0.1826 & 0.5162 & 0.7744 & 0.9734 &10000\\
 & & FlowLM(T=2000, step=3) & 0.1891 & 0.5227 & 0.7702 & 0.9720 &10000\\
 & & FlowLM(T=2000, step=1) & 0.1909 & 0.5431 & 0.7612 & 0.9359 &10000\\
 
  && FlowLM(T=100, step=5) & 0.1839 & 0.5251 & 0.7771 & 0.9744 & 10000 \\
  && FlowLM(T=100, step=3) & 0.1892 & 0.5238 & 0.7161 & 0.9752 & 10000 \\
  && FlowLM(T=100, step=1) & 0.1879 & 0.5345 & 0.7701 & 0.9725 & 10000 \\
  
  && FlowLM(T=20, step=5) & 0.1942 & 0.5352 & \textbf{0.7830} & 0.9764 & 10000 \\
 && FlowLM(T=20, step=3) & \textbf{0.2001} & 0.5390 & 0.7809 & \textbf{0.9766} & 10000 \\
 && FlowLM(T=20, step=1) & 0.1896 & \textbf{0.5454} & 0.7570 & 0.9443 & 10000 \\
 
 && FlowLM(T=5, step=5) & 0.1816 & 0.5237 & 0.7739 & 0.9781 & 10000 \\
  && FlowLM(T=5, step=3) & 0.1453 & 0.4754 & 0.7320 & 0.9671 & 10000 \\
  && FlowLM(T=5, step=1) & 0.0874 & 0.3342 & 0.5493 & 0.8284 & 10000 \\
\hline 
 % \cmidrule(lr){2-8}
 % & \multirow{3}{*}{Few-step}& \textsc{FlowLM2}(step=5) & 0.2294 & 0.4676 & 0.6886 & 0.8779 & 11.15 \\
 %  && \textsc{FlowLM2}(step=10) & 0.2539 & 0.4827 & 0.7179 & 0.8834 & 16.29 \\
 % && \textsc{FlowLM2}(step=20) & 0.2316 & 0.4513 & 0.6420 & 0.7622 & 13.40 \\
\hline 
\end{tabular}

\vspace{1em}
\begin{minipage}{\textwidth}
    Our results show that choosing T moderately larger than your target generation step is optimal. If T is too small, it will cause overfitting. If T is too large, it is not beneficial to few-step generation because it allocates excessive computational resources to intermediate steps that do not contribute significantly to generation quality.
\end{minipage}
\label{tab: Ablation of T}

\caption{Comprehensive comparison of Question generation results for input time-step rescale(20,200,1000). 1000 is original diffusion rescale value.}
\begin{tabular}{lll | ccc | cc }
\toprule
Tasks & Type &Methods & BLEU$\uparrow$ & R-L$\uparrow$ & BERTScore$\uparrow$ & dist-1$\uparrow$ &Training epoch\\
\midrule

\hline 
% ==================== Task 2: Paraphrase ====================
\multirow{9}{*}{\shortstack[l]{Paraphrase}} 
&\multirow{9}{*}{Few-step} & FlowLM(Rescale to 1000, step=5) & \textbf{0.1596} & 0.3484 & 0.5898 & \textbf{0.9206}&6000  \\
 & & FlowLM(Rescale to 1000, step=3) & 0.1595 & 0.3489 & 0.5878 & 0.9169 &6000\\
 & & FlowLM(Rescale to 1000, step=1) & 0.1524 & \textbf{0.3550} & 0.5713 & 0.8411 &6000\\
 
  && FlowLM(Rescale to 200, step=5) & 0.1584 & 0.3471 & \textbf{0.5901} & 0.9184 & 6000 \\
  && FlowLM(Rescale to 200, step=3) & 0.1592 & 0.3476 & 0.5865 & 0.9145 & 6000 \\
  && FlowLM(Rescale to 200, step=1) & 0.1518 & 0.3571 & 0.5703 & 0.8413 & 6000 \\
  
  && FlowLM(Rescale to 20, step=5) & 0.1574 & 0.3478 & 0.5864 & 0.9153 & 6000 \\
 && FlowLM(Rescale to 20, step=3) & 0.1572 & 0.3466 & 0.5721 & 0.9105 & 6000 \\
 && FlowLM(Rescale to 20, step=1) & 0.1511 & 0.3542 & 0.5692 & 0.8338 & 6000 \\
 
\hline 
 % \cmidrule(lr){2-8}
 % & \multirow{3}{*}{Few-step}& \textsc{FlowLM2}(step=5) & 0.2294 & 0.4676 & 0.6886 & 0.8779 & 11.15 \\
 %  && \textsc{FlowLM2}(step=10) & 0.2539 & 0.4827 & 0.7179 & 0.8834 & 16.29 \\
 % && \textsc{FlowLM2}(step=20) & 0.2316 & 0.4513 & 0.6420 & 0.7622 & 13.40 \\
\hline 
\end{tabular}

\label{tab: Ablation of rescale}
\vspace{0.5em}

\begin{minipage}{\textwidth}
    These results show that using the rescale value of original diffusion model when finetuning is the best choice, though the difference is marginal.
\end{minipage}
\end{table*}

\FloatBarrier
\begin{table*}[t]

\vspace{1em}
\centering
\caption{Ablation result of the regularization loss term on Question generation }
\renewcommand{\arraystretch}{1.1} %稍微增加行高，让表格不那么拥挤
% 定义列格式：
% l: 左对齐
% c: 居中
% |: 竖线
\begin{tabular}{lll | ccc | cc }
\toprule
Tasks & Type &Methods & BLEU$\uparrow$ & R-L$\uparrow$ & BERTScore$\uparrow$ & dist-1$\uparrow$ &Training epoch\\
\midrule

\hline 
% ==================== Task 2: Paraphrase ====================
\multirow{9}{*}{\shortstack[l]{Paraphrase}} 

 & \multirow{9}{*}{Few-step} & FlowLM($Reg_{rate}$=0, step=5) & 0.1596 & 0.3484 & 0.5898 & 0.9206&6000  \\
 & & FlowLM($Reg_{rate}$=0, step=3) & 0.1595 & 0.3489 & 0.5878 & 0.9169 &6000\\
 & & FlowLM($Reg_{rate}$=0, step=1) & 0.1524 & 0.3550 & 0.5713 & 0.8411 &6000\\
 
  && FlowLM($Reg_{rate}$=0.01, step=5) & \textbf{0.1607} & 0.3492 & \textbf{0.5906} & \textbf{0.9217} & 6000 \\
  && FlowLM($Reg_{rate}$=0.01, step=3) &  0.1605 & 0.3496 & 0.5876 & 0.9174 & 6000 \\
  && FlowLM($Reg_{rate}$=0.01, step=1) & 0.1532 &\textbf{ 0.3562} & 0.5736 & 0.8573 & 6000 \\
  
  && FlowLM($Reg_{rate}$=1, step=5) & 0.1511 & 0.3279 & 0.5757 & 0.9026 & 6000 \\
 && FlowLM($Reg_{rate}$=1, step=3) & 0.1509 & 0.3302 & 0.5763 & 0.9013 & 6000 \\
 && FlowLM($Reg_{rate}$=1, step=1) & 0.1402 & 0.3331 & 0.5534 & 0.8248 & 6000 \\
\hline 
\end{tabular}

\label{tab: Ablation of regulation}

\vspace{1em}
\begin{minipage}{\textwidth}
    According to \citet{fan2025online}, to mitigate the risk of policy collapse and preserve the diversity of the generated samples during online fine-tuning, they introduce a regularization term based on the Wasserstein-2 ($W_2$) distance. Since directly computing the $W_2$ distance between the distributions induced by continuous flow models is computationally intractable, they adopt a tractable upper bound derived from the transport dynamics. Specifically, they constrain the fine-tuned model $v_{\theta}$ to stay close to the pre-trained reference model $v_{\theta_{\text{ref}}}$ by minimizing the expected squared difference between their vector fields:

\begin{equation}
    \mathcal{L}_{\text{reg}}(\theta) = \mathbb{E}_{t \sim \mathcal{U}(0,1), z \sim p_t(z)} \left[ \| v_{\theta}(t, z) - v_{\theta_{\text{ref}}}(t, z) \|^2 \right].
    \label{eq:w2_reg}
\end{equation}

To apply this to our flow language model, we make some modifications as we predict $z_0$ instead of v. 
\begin{equation}
    \mathcal{L}_{\text{reg}}(\theta) =  reg_{rate}*\| v_{\theta}(t, z) - v_{\theta_{\text{ref}}}(t, z) \|^2 =reg_{rate}* \| \frac{z_t-z_{\theta}(t, z)}{t} - \frac{z_t-z_{\theta_{\text{ref}}}(t, z)}{t} \|^2=reg_{rate}*\frac{\|z_{\theta}(t, z) - z_{\theta_{\text{ref}}}(t, z) \|^2}{t^2}.
    \label{eq:w2_reg}
\end{equation}

This regularization term effectively bounds the $W_2$ distance between the learned distribution and the reference distribution. By incorporating $\mathcal{L}_{\text{reg}}$ into the training objective, we ensure that the model explores high-reward regions without deviating excessively from the data manifold captured by the reference model, thereby balancing the trade-off between reward maximization and generative diversity.
\end{minipage}

\begin{minipage}{\textwidth}
   We observe that training with $Reg_{rate}=0.01$ shows better performance, but the performance will drop if the $Reg_{rate}$ is too huge, which indicating that adding appropriate regularization loss term helps to improve the performance of flow matching language model.
\end{minipage}
\end{table*}

\FloatBarrier
\section{Some generated examples}
\label{sec:examples}
To further investigate the behavior of our model compared to the baseline under different sampling budgets, we present a case study on the Question Generation task. Table \ref{tab:case_study} displays the generated outputs given the source context related to \textit{"Karl Landsteiner"} and his \textit{"1930 Nobel Prize"}.

% =======================================================
% TABLE: Qualitative Comparison (Case Study)
% =======================================================
\begin{table*}[ht]
\centering
\caption{Case study on the Question Generation task. Comparison of generated samples between FlowLM (Ours) at few-step inference and DiffuSeq baselines. Semantic inconsistencies and lexical errors are highlighted in bold.}
\vspace{0.5em}
\small
\renewcommand{\arraystretch}{1.2}
\begin{tabular}{l|p{0.8\linewidth}}
\toprule
\textbf{Model (Steps)} & \textbf{Generated Question} \\
\midrule
\multicolumn{2}{c}{\textit{\textbf{Reference:} Karl Landsteiner won the Nobel Prize for medicine in 1930 for his discovery of what?}} \\
\midrule

% --- FlowLM N=1 ---
\multirow{4}{*}{\shortstack[l]{FlowLM\\(Ours, $N=1$)}} 
 & karl landsteiner won a nobel prize in \textbf{villains} for which \textbf{which} discovery \\
 & karl landsteiner won a nobel prize in \textbf{harley} for which in discovery \\
 & karl landsteiner won a nobel prize in 1930 for which \textbf{blanca} \\
 & the \textbf{starvation theer} won a nobel prize in 1930 for which medical discovery \\
\cmidrule{1-2}
% Analysis Note: Good syntax structure (S-V-O), but severe lexical hallucinations (villains, harley, blanca).
 
% --- FlowLM N=3 ---
\multirow{4}{*}{\shortstack[l]{FlowLM\\(Ours, $N=3$)}} 
 & karl landsteiner won a nobel prize in \textbf{intimidation} for which medical discovery \\
 & karl landsteiner won a nobel prize in \textbf{inquisition} for which medical discovery \\
 & karl landsteiner won a nobel\textbf{\textbackslash u53e4} in 1930 for which medical knees \\
 & karl landsteiner won a nobel prize in 1930 for which \textbf{know} \\
\cmidrule{1-2}
% Analysis Note: Semantics improving, "medical discovery" appears, but still some noise/artifacts.

% --- FlowLM N=5 ---
\multirow{4}{*}{\shortstack[l]{FlowLM\\(Ours, $N=5$)}} 
 & karl landsteiner won a nobel prize in 1930 for which \textbf{medical discovery} \\
 & karl landsteiner won a nobel prize in 1930 for which \textbf{medical famous} \\
 & karl landsteiner won a 1930 prize in 1930 for which medical discovery \\
 & karl landsteiner won a nobel prize in 1930 for which medical \textbf{montagu} \\
\cmidrule{1-2}
% Analysis Note: Best performance. Accurate semantics, consistent structure.

% --- DiffuSeq N=2000 ---
\multirow{4}{*}{\shortstack[l]{DiffuSeq\\(Baseline, $N=2000$)}} 
 & karl landsteiner won the \textbf{film} in 1930 for which medical flew which condition \\
 & karl landsteiner was a stand scientific in \textbf{music} for which medical medical discovery \\
 & karl landsteiner won a nobel prize in 1930 for which \textbf{two} \\
 & \textbf{else whose actresser} won a nobel prize in 1930 for field 1930... \\
\cmidrule{1-2}
% Analysis Note: Fluent grammar, but severe semantic drift (Hallucinations: film, music, actresser).

% --- DiffuSeq N=10 ---
\multirow{4}{*}{\shortstack[l]{DiffuSeq\\(Baseline, $N=10$)}} 
 & karl \textbf{landsteinerly} a 1930 leaves in 1930 for which in his \\
 & karl landsteinerly a nobel prize in 1930 for which medical medical \\
 & the \textbf{average theer} won a nobel prize in 1930 for which medical discovery \\
 & karl landsteiner won a nobel in in 1930 for which \\
\bottomrule
\end{tabular}

\label{tab:case_study}
\end{table*}

\subsection{Qualitative Analysis}

The examples in Table \ref{tab:case_study} provide significant insights into the behavior of flow matching versus standard diffusion in few-step scenarios.

\textbf{Structural Stability in Few-Step Generation:}
Our \textbf{FlowLM} model demonstrates remarkable structural stability even at the extreme setting of $N=1$. The generated sentences maintain a correct syntactic skeleton (Subject-Verb-Object), although they suffer from lexical hallucinations (e.g., replacing "1930" with random nouns like "villains" or "harley"). As the number of steps increases to $N=3$ and $N=5$, these lexical errors are rapidly corrected, and the model converges to semantically accurate outputs (e.g., correctly identifying "medical discovery"). This suggests that FlowLM learns a straight and stable probability flow that preserves syntax early in the generation process.

\textbf{Failure of Baselines in Fast Inference:}
In contrast, the Diffuseq baseline exhibits catastrophic failure when forced to generate in few steps ($N=10$). It suffers from morphological breakdown (e.g., "landsteinerly") and repetition, indicating that the complex noise schedule learned during 2000-step training cannot be approximated by a 10-step stride.

\textbf{Semantic Drift in High-Step Baselines:}
Surprisingly, even with the full budget of $N=2000$, Diffuseq shows signs of semantic drift. While the sentences are grammatically fluent, they frequently hallucinate incorrect topics, such as associating the Nobel Prize with "film" or "music" instead of medicine. This implies that the prolonged denoising process in standard diffusion might accumulate errors or lose condition specificity, whereas FlowLM's trajectory is more direct and condition-faithful.

% 这里再放个TS的详细结果 2页

% 比较一下不同时间步数量的结果 20步比2000步好吗 2页

% 也可以比较一下meanflow那种时间步采样 1页

% 每种放一下一些采样的例子 2页

\end{document}